\documentclass[twocolumn,9pt,compsoc]{IEEEtran}
\usepackage{preamble_ieee}
\usepackage{textcomp}
\usepackage{flushend}

\renewcommand{\captionN}[1]{\caption{\color{darkgray} \sffamily \fontsize{8}{10}\selectfont #1  }}
\newsavebox\psyncom
\sbox{\psyncom}{%
\raisebox{-4pt}{\begin{tikzpicture}[]\node[anchor=south] (A) {$\sync$};
\draw [->,>=stealth, thick] ([yshift=-1.5pt, xshift=2pt]A.north west) -- ([yshift=-1.5pt]A.north east); \end{tikzpicture}}}
\renewcommand{\psync}{\usebox{\psyncom}}
\title{Causality Networks} 
\author{
\begin{tabular}{cc}
Ishanu Chattopadhyay \\ \texttt{ic99@cornell.edu} 
\end{tabular}
\thanks{
Author is affiliated with the Computation Institute \& The Institute For Genomics and Systems Biology, University of Chicago. He holds a 5visiting appointment with the 
 Department of Computer Science, and the  Department of Mechanical \& Aerospace Engineering, Cornell University}
}
\newif\ifdetails
\detailstrue
\ifdetails
\prooftrue
\else
\prooffalse 
\renewcommand{\DETAILS}[1]{}
\fi
\def\NOPROOF{\begin{IEEEproof} Omitted. \end{IEEEproof}}

\begin{document}  
\maketitle 
\begin{abstract} 
While  correlation measures are  used to discern statistical relationships between 
observed variables in almost all branches of data-driven scientific inquiry, what we are really interested in is the existence of causal dependence. Statistical tests for causality, it turns out, are significantly
harder to construct; the difficulty stemming from both philosophical hurdles in making precise the notion of causality, and  the practical issue  of obtaining an operational procedure from a philosophically sound definition.  In particular, designing an efficient causality test,  that may be carried out in the absence of  restrictive pre-suppositions on the underlying dynamical structure of the data at hand, is non-trivial. Nevertheless, ability to computationally infer  statistical prima facie evidence of causal dependence may yield a far more discriminative tool for data analysis compared to the calculation of  simple correlations. In the present work, we present a new non-parametric test of Granger causality for quantized or symbolic  data streams generated by ergodic stationary sources. 
In contrast to state-of-art binary tests, our approach makes precise and computes the degree of causal dependence between data streams, without making any restrictive assumptions, linearity or otherwise. Additionally, without any a priori imposition of specific dynamical structure, we  infer explicit generative  models of causal cross-dependence, which may be then used for prediction. 
These explicit models are represented as  generalized probabilistic automata, referred to crossed  automata, and are shown to be sufficient to capture a fairly general class of causal dependence.
The proposed algorithms are computationally efficient in the PAC sense; $i.e.$, we find good models of  cross-dependence with high probability, with polynomial run-times and sample complexities.
The theoretical results are applied to weekly search-frequency data from Google Trends API for a chosen set of socially ``charged'' keywords. The causality network inferred from this dataset reveals, quite expectedly,  the causal importance of certain keywords. It is also illustrated that correlation analysis fails to gather such insight.
\end{abstract}

\ifdetails
{
\small
\tableofcontents
}
\else
\fi

\allowdisplaybreaks{
\section{Motivation}\label{sec1}
\begin{figure}[t]

\centering
\includegraphics[width=3.25in]{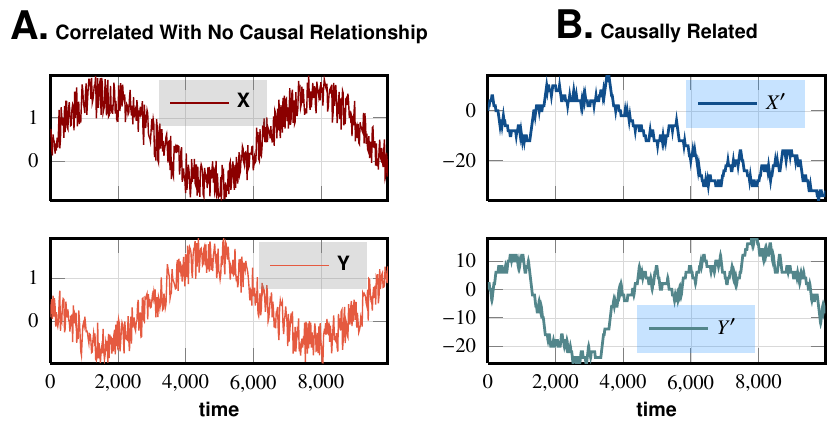}
\captionN{Correlation vs Causal Dependence. The signals in plate A are negatively correlated; but there is no statistically discernible causal dependence. This is because  future predictions of the  variable $X$  cannot be improved by considering  past values of the variable $Y$; the past values of  $X$ itself is sufficient to provide maximally correct prediction (Same argument applies on interchanging $X$ with $Y$). In contrast, the signals in plate B are causally related; while the past values of $X'$ are not useful to predict its own future values (it is an unbiased random walk), detailed analysis would reveal that the past values of $Y'$ do indeed carry unique information that improves future prediction in $X'$.
Thus, in addition to the negative correlation between $X'$ and $Y'$, there is prima facie statistical evidence for  causal dependency (in the sense of Granger causality) from $Y'$ to $X'$.
}\label{figCORR}
\end{figure}
``Correlation does not imply causation'' is a lesson  taught early and often in statistics. The obvious next question is almost always left untouched by the preliminary texts: how do we then test for causality?
This is an old  question debated in philosophy~\cite{hume,kant,Krikorian,Moyal,GOOD1961}, law~\cite{HH59}, statistics~\cite{Granger1963,Blalock1964,Granger1969,Suppes1970}, and more recently, in learning theory; with experts largely failing to agree on a philosophically sound operational approach. 
Causality, as an intuitive notion, is not hard to grasp. The lack of consensus on how to  infer  causal relationships from data  is perhaps ascribable  to  the difficulty in making this intuitive notion mathematically  precise.

\begin{quote}\itshape
``Unlike art, causality is a concept (for) whose definition people
know what they do not like, but few know what they do like.''
\end{quote}
\hfill - C.W.J. Granger~\cite{granger2}

Granger's attempt at obtaining a precise definition of causal influence  proceeds with the setting up of a  framework sufficiently general for statistical discourse: 
Consider a universe  in which  variables are
measured  at pre-specified time points  $t=1,2,\cdots$.
Denote all available knowledge in the universe upto time $n$ as  $\Omega_{\rightarrow n}$, and let $\Omega_{\rightarrow n} \setminus Y_{\rightarrow n}$ denote this complete information except the values taken
by a variable $Y_t$ up to time $n$, where $Y_{\rightarrow n} \in \Omega_{\rightarrow n}$. $\Omega_{\rightarrow n}$ includes no variates measured
at time points $t > n$, although it may well contain expectations or forecasts of
such values. However, these expectations will simply be functions of $\Omega_{\rightarrow n}$. We need  additional structure before we define causality, namely:

\begin{itemize}
\item \textbf{Axiom A:}  The past and present may cause the future, but the future cannot
cause the past.
\item \textbf{Axiom B:} $\Omega_{\rightarrow n}$ contains no redundant information, so that if some variable
$Z_t$ is functionally related to one or more other variables, in a deterministic
fashion, then $Z_{\rightarrow n}$ should be excluded from $\Omega_{\rightarrow n}$.

\end{itemize}
Within this framework, Granger suggests the following  definition, noting that it is not  effective~\cite{HMU01},  $i.e.$, not directly applicable to data:
\begin{defn}[Granger Causality]\label{defG}
$Y_{\rightarrow n}$ is said to cause $X_{n+1}$ if given a set $A$ in which the variable $X_{n+1}$ takes values in, we have: 
\cgather{
Pr(X_{n+1} \in A \vert \Omega_{\rightarrow n} ) \neq Pr(X_{n+1} \in A \vert \Omega_{\rightarrow n} \setminus Y_{\rightarrow n}  )
}
\end{defn}

Granger's notion  is intuitively simple: $Y$ is a cause of $X$, if it has unique information that alters the probabilistic  estimate of the immediate future of $X$.  Not all notions of causal influence are expressible in this manner, neither can all philosophical subtleties be adequately  addressed. Granger's  motivation was more pragmatic - he was primarily interested in  obtaining a mathematically precise framework that leads to an effective or algorithmic solution - a concrete statistical test for causality. 

\subsection{Granger's Operational Definitions of Causality}
Short of encoding ``all knowledge'' in the universe upto a given time point, Definition~\ref{defG} is not directly useful. Suppose that one is interested in the possibility that a
vector series $Y_t$ causes another vector $X_t$. Let $J_n$ be an information set
available at time $n$, consisting of terms of the vector series $Z_t$, i.e.,
\cgather{
J_n = \{Z_t: t \leqq n \}
}
$J_n$ is said to be a proper information set with respect to $X_t$, if $X_t$ is included
within $Z_t$. Further, suppose that $Z_t$ does not include any component of $Y_t$, and define
\cgather{
J_n' = \{ (Z_t, Y_t) : t \leqq n \}
}
Denote by $F(X_{n+1} \vert J_n )$ the conditional distribution function of $X_{n+1}$ given
$J_n$ with  mean $\mathbf{E}(X_{n+1} \vert J_n)$. Then, we may define:

\begin{defn}\label{defG2}
\begin{itemize}
\item $Y_n$ does not cause $X_{n+1}$  with respect to $J_n'$ if:
\cgather{
 F(X_{n+1} \vert J_n) = F(X_{n+1} \vert J_n')
}
$i.e.$, the extra information in $J_n'$, has not affected the conditional
distribution. A necessary condition is that:
\cgather{
\mathbf{E}(X_{n+1} \vert J_n ) = \mathbf{E}(X_{n+1} \vert J_n')
}
\item If $J_n' = \Omega_n$, the universal information set, and if 
\cgather{
F(X_{n+1} \vert J_n) \neq F(X_{n+1}  \vert J_n' )
}
then, $Y_n$ is said to cause $X_{n+1}$.
\item $Y_n$ is  a prima facie cause of $X_{n+1}$ with respect to $J_n'$ if:
\cgather{
 F(X_{n+1} \vert J_n) \neq F(X_{n+1} \vert J_n')
}
\item $Y_n$ is  said not to cause $X_{n+1}$ in the mean with respect to $J_n'$ if:
\cgather{\label{EQCm}
\Delta (J_n') \triangleq \mathbf{E}(X_{n+1} \vert J_n' ) - \mathbf{E}(X_{n+1} \vert J_n) = 0
}
\item If $\Delta(J_n')$ is not identically zero, then $Y_n$ is a prima facie of cause $X_{n+1}$ in the mean with respect to $J_n'$.
\end{itemize}
\end{defn}

Definition~\ref{defG2} is far more useful;  with a little more structure we may obtain an effective causality test. We will shortly discuss  these additional assumptions that are commonly employed. But first, we elucidate some key implications of Granger's definition of causality.
\subsection{Properties of Granger Causality}
\subsubsection{Deterministic Causation}It is impossible to find a cause for a series that is self-deterministic.
Thus, if $X_n$ is expressible as  a deterministic function of its previous values, then no additional information can alter this ``prediction'', and hence no other cause is necessary.
 In the light of Taken's embedding theorem~\cite{Takens80}, this has an important implication. For certain classes of dynamical systems specified by systems of ordinary differential equations, a single variable may be able to perfectly reconstruct the dynamics through Taken's delay-coordinate construction, implying that  other variables may be found to be causally superfluous as far as Granger's notion is concerned.

\subsubsection{Refexivity, Symmetry, \& Transitivity}Clearly, causality is not required to be symmetric; $X_t$ may cause $Y_t$ but  not the other way around.
Additionally, $X_t, X_{t'} $ could be independent for all $t\neq t'$, and yet $Y_t$ could be a cause for $X_t$. Thus, $X_t$ is not required to be  a cause for itself, $i.e.$, causality is not necessarily reflexive.
It  is also not required to be  transitive (See Example 1 in \cite{granger2}), $i.e.$, $X_t$ causes $Y_t$ and $Y_t$ causes $Z_t$ does not necessarily imply $X_t$ causes $Z_t$ in the sense of Definition~\ref{defG2}.

\subsubsection{Missing Variables \& Unobserved  Causes}Missing variables can induce spurious causality. Unobserved common causes are particularly important. For example~\cite{granger2}, suppose:
\cgather{
Z_t = a_t \\
X_t = a_{t-1} + b_t\\
Y_t = a_{t-2} + c_t
}
where $a_t,b_t,c_t$ are independent white noise processes. Here $Z_t$ is a common cause. However, if we only observe $X_t$ and $Y_t$, then $X_t$ seems to be causing $Y_t$. There is no general fix for this; but it has been shown that such one-way spurious causation is unlikely in physical systems, and a two-way or feedback relationship is a more likely outcome with unobserved common causes~\cite{sims77}.

\subsection{Additional Assumptions  in Standard  Approaches}
Inferring causality in the mean (See Eq.~\eqref{EQCm}) is easier, and  if one is satisfied with using minimum mean square prediction error as the criterion to
evaluate incremental predictive power, then one may use linear one-step-ahead least squares predictors to obtain an operational procedure from Eq.~\eqref{EQCm}: if $VAR(X \vert J_n)$ is the variance of one-step forecast error of $X_{n+1}$ given $J_n$, then $Y$ is a prima facie cause of $X$ with respect to $J_n'$ if:
\cgather{
VAR(X\vert J_n')  < VAR(X \vert J_n)
}
Testing for
bivariate Granger causality \textit{in the mean} involves estimating a linear reduced-form vector autoregression:
\cgather{
X_t = A(L) X_t + B(L) Y_t + U_{X,t}\\
Y_t = C(L) X_t + D(L) Y_t + V_{Y,t}
}
where $A(L)$, $B(L)$, $C(L)$, and $D(L)$ are one-sided lag polynomials in the lag operator $L$ with roots all-distinct, and  outside the unit circle. The regression errors $U_{X,t},V_{Y,t}$ are assumed to be mutually independent and individually i.i.d. with zero mean and constant variance. A standard joint test ($F$ or $\chi^2$-test)  is used to determine whether lagged $Y$ has significant linear predictive power for current $X$.
The null hypothesis that $Y$ does not strictly Granger cause $X$ is rejected if
the coefficients on the elements in $B(L)$ are jointly
significantly different from zero. 
\begin{figure*}[t]
\centering
\includegraphics[width=6in]{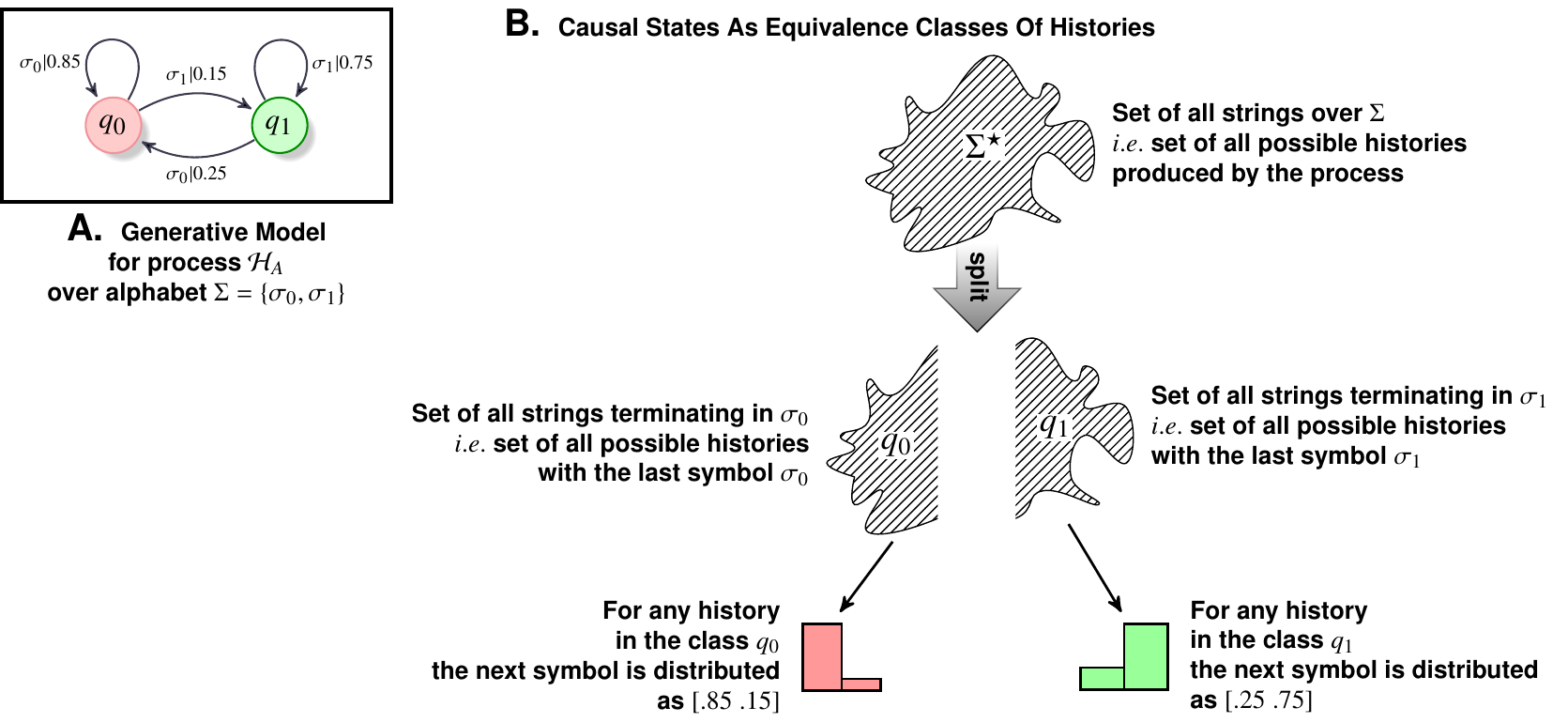}

\captionN{Illustration of the notion of causal states. Plate A shows a probabilistic finite state automata which generates a stationary ergodic quantized stochastic process $\mathcal{H}_A$, taking values over the alphabet $\Sigma=\{\sigma_0, \sigma_1\}$. The meaning of the states is illustrated in plate B: the set of all possible histories or sequences produced by the process $\mathcal{H}_A$ may be split into two classes, which contain respectively all sequences ending in $\sigma_0$ and $\sigma_1$. This classes represent the states $q_0$ and $q_1$ respectively. For any string in the class mapping to $q_0$,  the next symbol is distributed as $[0.85 \ 0.15]$ and for any string in the class mapping to $q_1$, the next symbol is distributed as $[.25 \ 0.75]$ over the alphabet $\Sigma$. Note that this is dictated by the structure of the generative model shown in plate A, $e.g.$, it is easy to follow the edges, and conclude that any string terminating in $\sigma_0$ ends in state $q_0$ irrespective of where it starts from. Thus, $\mathcal{H}_A$ has two causal states, since there is only two such classes of histories that lead to different future evolution of any generated sequence.}\label{figcausalstate}

\end{figure*}

Linear tests pre-suppose restrictive and often unrealistic~\cite{darnell90,epstein87}  structure on  data. Brock~\cite{Brock91} presents a simple bivariate  model to analytically demonstrate  the limitations of  linear  tests in uncovering nonlinear influence. To address this issue, a number of nonlinear  tests  have been reported, $e.g.$, with generalized autoregressive conditional heteroskedasticity (GARCH) models~\cite{asi00}, using wavelet transforms~\cite{papa03}, or heuristic additive relationships~\cite{hr05}. However, these  approaches often assume the class of allowed  non-linearities; thus not quite alleviating the problem of pre-supposed structure. This is not just an academic issue; Granger causality  has been shown to be  significantly  sensitive to  non-linear transformations~\cite{ref53}.

Non-parametric approaches, $e.g.$ the Hiemstra-Jones (HJ) test~\cite{Hiemstra1994} on the other hand,  attempt to completely  dispense with pre-suppositions on the causality structure.   Given two series $X_t$ and $Y_t$, the HJ test (which is  a modification of the Baek-Brock test~\cite{Baek1992}) uses  correlation integrals to test if the probability of similar  futures  for  $X_t$ given similar pasts, change significantly if we  condition instead on similar pasts for both $X_t$ and $Y_t$ simultaneously. 
Nevertheless, the data series are required  to be  ergodic, stationary, and  absolutely regular $i.e.$ $\beta$-mixing, with an upper bound on the rate at which the $\beta$-coefficients approach zero~\cite{DK83}, in order to achieve consistent estimation of the correlation integrals. The additional assumptions beyond ergodicity and stationarity serve to guarantee that sufficiently separated fragments of the data series are  nearly independent. The HJ test and its variants~\cite{diks06,sp10} have been quite successful in econometrics;  uncovering nonlinear causal relations
between money \& income~\cite{Baek1992}, aggregate stock returns \& macroeconomic factors~\cite{hk93}, currency future returns~\cite{asima05} and stock price \& trading volume~\cite{Hiemstra1994}. Surprisingly, despite clear evidence that linear tests typically have low power in uncovering nonlinear causation~\cite{Hiemstra1994,asima05}, application of non-parametric tests  has been limited in  areas beyond financial, or macroeconomic interests.
\section{Contribution Of The Present Work}

The HJ test and its variants are specifically designed to detect  presence of   Granger causality at a pre-specified significance level; there is  no obvious extension by which a generative nonlinear model of this  cross-dependence may be distilled from the data at hand. We are left with an oracle in a black box - it answers questions without any insight on the dynamical structure of the system under inquiry.
On  the other hand, linear regression-based, as well as  parametric nonlinear approaches,  have one discernible advantage; they produce  generative models of causal influence between the observed variables.  Hiemstra's suggestion was to view non-parametric tests purely as a  tool for uncovering existence of  non-linearities in system dynamics; leaving the task of detailed  investigation of dynamical structure to parametric model-based approaches:
\begin{quote}
\itshape
Although the nonlinear (non-parametric) approach to causality testing presented here can
detect nonlinear causal dependence with high power, it provides no guidance
regarding the source of the nonlinear dependence. 
Such guidance must be left to theory, which may suggest specific parameterized structural models. 
\end{quote}
\hfill - Hiemstra $\etal$~\cite{Hiemstra1994}

This is perhaps the motivation behind applying the HJ test in \cite{Hiemstra1994} to error residuals from an estimated linear autoregressive model; by removing linear structure using regression the authors conclude that any additional causal influence must be nonlinear in origin.

However is it completely unreasonable to ask for a generative model of causal cross-dependence, where we are unwilling to a priori specify any dynamical structure? 
The central objective  of the present work is to show that such an undertaking  is indeed fruitful; beginning with sequential observations on two variables, we may  infer non-heuristic generative models of causal influence with no pre-supposition on the nature of the hidden dynamics, linear or nonlinear.

This however is a non-trivial exercise; if we are to allow for the appearance of dynamical models with unspecified and unrestricted structure, we need to rethink the framework within which such inference is carried out. It is well understood that this task is unattainable in the absence of at least some broad  assumptions on the statistical nature of the sources~\cite{granger2}, particularly on the nature of temporal or sequential variation in the underlying statistical parameters. We restrict ourselves to ergodic and stationary sources, and additionally assume that the data streams take values within finite sets; $i.e.$, we only consider ergodic, stationary quantized stochastic processes (explicit definition given later).

We briefly recapitulate an earlier result that  underlying generators for individual data streams from ergodic stationary quantized processes  may be represented as probabilistic automata.  And then we show that  for two streams, generative models of causal influence may be represented as generalized probabilistic automata,  referred to as \textit{crossed automata}. Our task then reduces to inferring these crossed machines from data, in the absence of a priori knowledge of structure and parameters involved. True to the possible asymmetric nature of causality, we show that such inferred logical  machines are direction-specific; the crossed machine capturing the influence from stream $s_A$ to stream $s_B$ is not required to be identical to the one from $s_B$ to $s_A$. 
Additionally, we show that absence of causal influence between data streams manifests as a trivial crossed machine, and the existence of such trivial representations in both directions is necessary and sufficient for statistical independence between the data streams under consideration. 

Our ability to find generative models of causal dependence allows us to carry out out-of-sample prediction. In contrast, the HJ test is vulnerable to Granger's objection~\cite{granger2}, that in absence of an inferred model, one is not strictly adhering to the original definition of Granger causality, which requires improved predictive ability, and not simply analysis of past data. Model-based approaches can indeed make and test predictions, but at the cost of the pre-imposed model structure (See proposed recipe in \cite{granger2}). The current approach, in contrast, produces generative models without pre-supposed structure; and is therefore able to carry out and test predictions without the aforementioned cost.

In addition to obtaining explicit models of causal dependence between observed data streams, the present work identifies a new test for Granger causality, for quantized processes ($i.e.$ processes which take values within a finite set). Our approach involves computing the \textit{coefficient for causal dependence $\gamma^A_B$}, from the process generating a stream $s_A$ to the process generating a stream $s_B$. It  is defined as the ratio of the expected change in the entropy of the next-symbol distribution in stream $s_B$ conditioned over observations in the stream $s_A$ to the entropy of the next-symbol distribution in stream $s_B$, conditioned on the fact that no observations are made on stream $s_A$. We show that $\gamma^A_B$ takes values on the closed unit interval, and higher values indicate stronger predictability of $s_B$ from $s_A$, $i.e.$, a higher degree of causal influence. Thus, true to Granger's notion of causality,  $\gamma^A_B$ quantifies the amount of additional information that observation on the stream $s_A$ provide about the immediate future in stream $s_B$. We show that streams $s_A,s_B$ are statistically independent if and only if $\gamma^A_B = \gamma^B_A = 0$. Importantly, it is also easy to give examples where $\gamma^A_B = 0 $ and $\gamma^B_A > 0$, thus illustrating the existence of directional influence (See Figure~\ref{figEX1}).

It is important to note that  the state of the art  techniques, including the HJ test, merely ``test'' for the existence of a causal relationship; setting up the problem in the framework of a classical binary hypothesis testing.  No attempt is made to infer the  degree of the  causal connection, once the existence of such a relationship is statistically established. Perhaps one may point to the significance value at which the test is passed (or failed); but statistical significance of the tests is not, at least in any obvious manner, related to the \textit{degree of causality}. In contrast, our definition of  $\gamma^A_B$ has this notion clearly built in. As we stated earlier, higher values of the coefficient indicate a stronger causal connection; and $\gamma^A_B=1$ indicates a situation in which the  symbol in the immediate future of $s_B$ is deterministically fixed given the past values of $s_A$, but looks completely random if only the past values of $s_B$ are available.

While the HJ test and the computational inference of the coefficient of causality imposes similar assumptions on the data, the assumptions in the latter case are perhaps more physically transparent. Both approaches require ergodicity and stationarity; the HJ test further requires the processes to be absolutely regular ($\beta$-mixing), with a certain minimum asymptotic decay-rate of the $\beta$ coefficients (See~\cite{miz95}, footnote on pg. 4, and \cite{DK83}). Absolute regularity is one of the several ways one can have weak dependence; essentially implying that two sufficiently separated  fragments of a  data stream are nearly independent. Our algorithms also require weak dependence in addition to stationarity and ergodicity; however instead of invoking mixing coefficients, we require that the processes have a finite number of \textit{causal states} (See Figure~\ref{figcausalstate}). Causal states are equivalence classes of histories that produce similar futures; and hence a finite number of causal states dictates that we need  a finite number of classes of histories for future predictions.

As for the computational cost of the algorithms, we show that the inference of the crossed automata is PAC-efficient~\cite{valiant84}, $i.e.$, we can infer good models with high probability, in asymptotically polynomial time and sample complexity. 
The HJ test may well have good computational properties; but the literature lacks a detailed investigation.

In summary, the key  contributions of the present work may be enumerated as:
\begin{enumerate}
\item A new non-parametric test for Granger causality for quantized processes is introduced. Going beyond binary hypothesis testing, we quantify the  notion of the degree of causal influence between observed data streams, without pre-supposing any particular model structure.
\item Generative models  of causal influence  are shown to be inferrable with no a priori imposed dynamical structure beyond ergodicity, stationarity, and a form of weak dependence. The explicit generative models  may be used for prediction.
\item The proposed  algorithms are shown  to be PAC-efficient.
\end{enumerate}

\subsection{Organization}
The rest of the paper is organized as follows: Section~\ref{sec3}
makes precise the notion of quantized stochastic processes, and the connection to probabilistic automata. Some of the material in this section has appeared elsewhere~\cite{CL12g}, but is included for the sake of completeness, and due to some key technical differences, and extensions to the exposition. Section~\ref{sec4} presents the framework for representing generative models for cross-dependence; introducing  crossed probabilistic automata. The coefficient of causal dependence is defined, and the directional nature of the causality is investigated in this context.  Section~\ref{sec5} presents algorithm \algo for inferring generative self-models from individual data streams. Again, \algo has been reported earlier in~\cite{CL12g}, but is included here for the sake of completeness. Section~\ref{sec6} presents algorithm \xalgo, which infers crossed automata from pairs of data streams, as generative models of direction-specific causal dependence. The complexity and PAC-efficiency of \xalgo is investigated.  Section~\ref{sec7} deals with the inference of  causality networks between multiple data streams, and the fusion of future predictions from  inferred crossed models. A simple application of the developed theory is illustrated in Section~\ref{sec8}, where the causality network between weekly search-frequency data (data source: Google Trends) for a chosen list of keywords is computed.
The paper is concluded in Section~\ref{sec9}.

\section{Quantized Stochastic Processes \& Probabilistic Automata}\label{sec3}
Our approach hinges upon effectively using probabilistic automata to model stationary, ergodic processes. Our automata models are distinct to those reported in the literature~\cite{P71,VTCC05}.  The details of this  formalism can be found in \cite{CL12g}; we include a brief overview here for  completeness.

\begin{notn} 
 $\Sigma$  is a finite alphabet  of symbols. The set of all finite but possibly unbounded strings over $\Sigma$ is denoted by $\Sigma^\star$~\cite{HMU01}. The set of finite strings over $\Sigma$ form a concatenative monoid, with the empty word $\lambda$ as identity. 
The set of strictly infinite strings on $\Sigma$ is denoted as $\Sigma^\omega$, where $\omega$ denotes the first transfinite cardinal. 
For a string $x$,  $\vert x \vert$ denotes its length, and for a set $A$,    $\vert A \vert$ denotes its cardinality. Also, $\Sigma^d_+ = \{x \in \Sigma^\star \textrm{ s.t. } \vert x \vert  \leqq d\}$.
\end{notn}

\begin{defn}[QSP]\label{defQSP}
A QSP $\mathcal{H}$ is a discrete time $\Sigma$-valued strictly stationary, ergodic stochastic process, $i.e.$ 
\cgather{
\mathcal{H} = \left \{ X_t: X_t \textrm{ is a $\Sigma$-valued random variable}, t \in \mathbb{N}\cup \{0\} \right \}
} 
A  process is ergodic if  moments may be calculated from a sufficiently long realization, and strictly stationary if moments are time-invariant.
\end{defn}
We next formalize the connection of QSPs to PFSA generators.
  We develop the theory assuming multiple realizations of the QSP $\mathcal{H}$, and  fixed initial conditions. Using ergodicity, we will be then able to apply our construction to a single sufficiently long realization, where  initial conditions cease to matter.
\begin{defn}[$\sigma$-Algebra On Infinite Strings]\label{defsigmaalgebra}
 For the set of infinite strings on  $\Sigma$, we define $\mathfrak{B}$ to be the smallest $\sigma$-algebra generated by the family of sets $\{  x \Sigma^\omega : x \in \Sigma^\star\}$.
\end{defn}
\begin{lem}\label{QSPtoProb}
Every QSP  induces a  probability space $(\Sigma^\omega,\mathfrak{B},\mu)$.
\end{lem} 
\Proof{
Assuming stationarity, we can construct a probability measure $\mu: \mathfrak{B} \rightarrow [0,1]$  by defining
for any sequence $x\in \Sigma^\star\setminus \{\lambda\}$, and a sufficiently large number of realizations $N_R$ (assuming ergodicity):
\cgathers{
 \mu( x \Sigma^\omega) = \lim_{N_R \rightarrow \infty}\frac{ \textrm{\small \#  of initial occurrences of $x$}}{\begin{array}{c} \textrm{\small \#  of initial occurrences} \\\textrm{\small of all sequences of length  $\vert x \vert$}\end{array}}
}
and extending the measure to elements of $\mathfrak{B} \setminus B$ via at most countable sums. Thus $\mu(\Sigma^\omega) = \sum_{x \in \Sigma^\star} \mu( x \Sigma^\omega) = 1$, and for the null word  $\mu(\lambda \Sigma^\omega) =  \mu(\Sigma^\omega) = 1$.
}

 \begin{notn}
 For notational brevity, we denote $\mu( x \Sigma^\omega)$ as $Pr(x)$.
 \end{notn}

Classically,  automaton states are equivalence classes for the  Nerode relation;  two strings are  equivalent if and only if any finite extension of the strings is either both in the language under consideration, or neither are~\cite{HMU01}. We use a probabilistic extension~\cite{CR08}.

\begin{defn}[Probabilistic Nerode Equivalence Relation]\label{defnerode} $(\Sigma^\omega,\mathfrak{B},\mu)$ induces an equivalence relation $\sim_{N}$ on the set of finite strings $\Sigma^\star$ as:
\cgather{
\forall x,y \in \Sigma^\star, 
\smash{x \sim_{N} y \iff \forall z \in \Sigma^\star \bigg (\big (} Pr(xz)=Pr(yz)=0 \big )  \notag \\  \bigvee \big \vert  Pr(xz)/Pr(x)- Pr(yz)/Pr(y)\big \vert =0\bigg )
}
\end{defn}

\begin{notn}
For $x \in \Sigma^\star$,  the equivalence class of $x$ is  $[x]$.
\end{notn}

It is easy to see   that $\sim_{N}$ is right invariant, $i.e.$ 
\cgather{
x \sim_{N} y \Rightarrow \forall z \in \Sigma^\star, xz \sim_{N} yz 
}
A right-invariant equivalence on $\Sigma^\star$ always induces an automaton structure; and hence the probabilistic Nerode relation induces  a probabilistic automaton: states are equivalence classes of $\sim_{N}$, and the transition structure arises as follows: For states $q_i,q_j$, and  $x \in \Sigma^\star$,
\begin{gather}
([x]=q ) \wedge ([x \sigma ] = q' 
 )\Rightarrow q \xrightarrow{\sigma} q'
\end{gather}
Before formalizing the above construction, we introduce the notion of probabilistic automata with initial, but no final, states.


\begin{defn}[Initial-Marked PFSA]\label{defpfsa} An initial marked probabilistic finite state automaton (a Initial-Marked PFSA)   is a quintuple $(Q,\Sigma,\delta,\pitilde,q_0)$, where $Q$ is a finite  state set, $\Sigma$ is the alphabet,  $\delta:Q \times \Sigma \rightarrow Q$ is the  state transition function,  $\pitilde : Q \times \Sigma \rightarrow [0,1]$  specifies the conditional symbol-generation probabilities, and $q_0\in Q$ is the initial state.  $\delta$ and $\pitilde$ are recursively extended to arbitrary  $y=\sigma x \in \Sigma^\star$ as follows:
\begin{subequations}
\cgather{
\forall q \in Q, \delta(q,\lambda) = q\\
\delta(q,\sigma x) = \delta(\delta(q,\sigma),x)\\
\forall q \in Q, \pitilde(q,\lambda) = 1\\
\pitilde(q,\sigma x) = \pitilde(q,\sigma)\pitilde(\delta(q,\sigma),x)
}
\end{subequations}
Additionally, we impose  that for  distinct states $q_i,q_j \in Q$, there exists a string $x \in \Sigma^\star$, such that $\delta(q_i,x) = q_j$, and $\pitilde(q_i,x) > 0$.
\end{defn}

Note that the probability of the null word is unity from each state.

If the current state and  the next symbol is  specified, our next  state is fixed; similar to Probabilistic Deterministic Automata~\cite{Gavalda06}. However, unlike the latter, we lack final states in the model. Additionally, we assume our graphs to be strongly connected. Later we will remove initial state dependence using ergodicity. 
Next we formalize how  a PFSA arises  from  a  QSP.

\begin{lem}[PFSA Generator]\label{lemPFSAgen}
Every Initial-Marked PFSA $G=(Q,\Sigma,\delta,\pitilde,q_0)$ induces a unique probability measure $\mu_G$ on the measurable space  $(\Sigma^\omega,\mathfrak{B})$.
\end{lem}
\Proof{
Define   set function $\mu_G$ on the measurable space  $(\Sigma^\omega,\mathfrak{B})$:
\begin{subequations}
\calign{
\mu_G(\varnothing) &\triangleq 0\\
\forall x \in \Sigma^\star, \mu_G(x\Sigma^\omega) &\triangleq \pitilde(q_0,x)\\
\forall x,y \in \Sigma^\star, \mu_G(\{x,y\}\Sigma^\omega) &\triangleq \mu_G(x\Sigma^\omega)+ \mu_G(y\Sigma^\omega)
}
\end{subequations}
Countable additivity of  $\mu_G$ is immediate, and we have (See Definition~\ref{defpfsa}):
\cgather{
\mu_G(\Sigma^\omega) = \mu_G(\lambda\Sigma^\omega) = \pitilde(q_0, \lambda) = 1
}
implying that $(\Sigma^\omega,\mathfrak{B}, \mu_G)$ is a probability space.
}

We refer to $(\Sigma^\omega,\mathfrak{B}, \mu_G)$ as the probability space generated by the Initial-Marked PFSA $G$. 

\begin{lem}[Probability Space To PFSA]\label{lemPROB2PFSA}
If the probabilistic Nerode relation corresponding to a  probability space $(\Sigma^\omega,\mathfrak{B}, \mu)$ has a finite index, then the latter has an  initial-marked PFSA generator.
\end{lem}
\Proof{
Let  $Q$ be the set of equivalence classes of the probabilistic Nerode relation (Definition~\ref{defnerode}),  and define functions $\delta:Q \times \Sigma \rightarrow Q$, $\pitilde:Q \times \Sigma \rightarrow [0,1]$ as:
\begin{subequations}
\calign{
& \delta([x],\sigma) = [x\sigma]\\
& \pitilde([x],\sigma) = \frac{Pr(x'\sigma)}{Pr(x')} \textrm{ for any choice of } x' \in [x] \label{eqpit}
}
\end{subequations}
where we  extend $\delta,\pitilde$  recursively  to   $y=\sigma x \in \Sigma^\star$ as 
\begin{subequations}
\cgather{\delta(q,\sigma x) = \delta(\delta(q,\sigma),x)\\\pitilde(q,\sigma x) = \pitilde(q,\sigma)\pitilde(\delta(q,\sigma),x)}\end{subequations}
For verifying the null-word probability, choose a $x \in \Sigma^\star$ such that $[x] = q$ for some $q \in Q$. Then, from  Eq.~\eqref{eqpit}, we have:
\cgather{
 \displaystyle \pitilde(q,\lambda)= \frac{Pr(x'\lambda)}{Pr(x')} \textrm{ for any  } x' \in [x] \Rightarrow \pitilde(q,\lambda)
=  \frac{Pr(x')}{Pr(x')} = 1
}
Finite index of $\sim_{N}$ implies  $\vert Q\vert < \infty$, and hence denoting  $[\lambda]$ as $q_0$, we conclude:  $G=(Q,\Sigma,\delta,\pitilde,q_0)$ is an Initial-Marked PFSA.  Lemma~\ref{lemPFSAgen} implies that $G$ generates $(\Sigma^\omega,\mathfrak{B}, \mu)$,  which completes the proof.
}

The above construction yields a \textit{minimal realization} for the Initial-Marked PFSA, unique up to state renaming. 


\begin{lem}[QSP to PFSA]\label{lemQSP2PFSA}
Any QSP  with a finite index Nerode equivalence is generated by an Initial-Marked PFSA.
\end{lem}
\Proof{
Follows immediately from Lemma~\ref{QSPtoProb} (QSP to Probability Space) and Lemma~\ref{lemPROB2PFSA} (Probability Space to PFSA generator).
}

\subsection{Canonical Representations}
 We have defined a QSP as both ergodic and stationary, whereas the Initial-Marked PFSAs have a designated initial state. Next we introduce  canonical representations to remove initial-state dependence. We use $\Pitilde$ to denote the matrix representation of  $\pitilde$, $i.e.$, $\Pitilde_{ij} = \pitilde(q_i,\sigma_j)$,  $q_i \in Q, \sigma_j \in \Sigma$. We need the notion of transformation matrices $\Gamma_\sigma$.

\begin{defn}[Transformation Matrices]\label{defGamma}
 For an initial-marked PFSA $G=(Q,\Sigma,\delta,\pitilde,q_0)$, the symbol-specific transformation matrices $\Gamma_\sigma \in \{0,1\}^{\vert Q \vert \times \vert Q \vert}$ are:
\cgather{
\Gamma_\sigma \big \vert_{ij} = \begin{cases}
                                 \pitilde(q_i,\sigma), & \textrm{if } \delta(q_i,\sigma) = q_j \\
				 0, & \textrm{otherwise}
                                \end{cases}
}
\end{defn}
Transformation matrices have a single non-zero entry per row, reflecting our generation rule that given a state and a generated symbol, the next state is fixed. 

First, we note that, given an initial-marked PFSA $G$,  we can associate a probability distribution $\wp_x$ over the states of $G$  for each $x \in \Sigma^\star$ in the following sense:
if $x=\sigma_{r_1}\cdots \sigma_{r_m} \in \Sigma^\star$, then we have:
\cgather{
\wp_x = \wp_{\sigma_{r_1}\cdots \sigma_{r_m}} = \underbrace{\frac{1}{\vert \vert \wp_\lambda \prod_{j=1}^m \Gamma_{\sigma_{r_j}}  \vert \vert_1 }}_{\textrm{Normalizing factor}}\wp_\lambda \prod_{j=1}^m \Gamma_{\sigma_{r_j}}
\label{eqLEP}}
where $\wp_\lambda$ is the stationary distribution over the states of $G$.  Note that there may exist more than one string that leads to a distribution $\wp_x$, beginning from the stationary distribution $\wp_\lambda$. Thus, $\wp_x$  corresponds to an equivalence class of strings, $i.e.$, $x$ is  not unique. 
\begin{defn}[Canonical Representation]\label{defcanon}

 An  initial-marked PFSA $G=(Q,\Sigma,\delta,\pitilde,q_0)$ uniquely induces a canonical representation $(Q^C,\Sigma,\delta^C,\pitilde^C)$, where $Q^C$ is  a subset of the  set of probability distributions over  $Q$, and  $\delta^C: Q^C \times \Sigma \rightarrow Q^C$,  $\pitilde^C: Q^C \times \Sigma \rightarrow [0,1]$ are constructed  as follows:
\begin{enumerate}
 \item Construct the stationary distribution on $Q$ using the transition probabilities of the Markov Chain induced by $G$, and include this as the first element $\wp_\lambda$ of $Q^C$. Note that the transition matrix for $G$ is the row-stochastic matrix $M \in [0,1]^{\vert Q \vert \times \vert Q \vert}$, with
$
M_{ij} = \sum_{\sigma: \delta(q_i,\sigma)=q_j}\pitilde(q_i,\sigma)
$, and hence $\wp_\lambda$ satisfies:
\cgather{
\wp_\lambda M = \wp_\lambda
}
\item Define  $\delta^C$ and $\pitilde^C$ recursively:
\calign{
&\delta^C(\wp_x, \sigma) = \frac{1}{\vert \vert \wp_x \Gamma_\sigma \vert \vert_1}\wp_x \Gamma_\sigma \triangleq \wp_{x\sigma}\\
&\pitilde^C(\wp_x,\sigma) = \wp_x \Pitilde
}
\end{enumerate}
\end{defn}
For a QSP $\mathcal{H}$, the canonical representation is denoted as $\mathcal{C}_\mathcal{H}$.
\begin{lem}[Properties of Canonical Representation]\label{lemcanonstruc}
 Given an initial-marked PFSA $G=(Q,\Sigma,\delta,\pitilde,q_0)$:
\begin{enumerate}
\item The canonical representation is independent of the initial state.
\item The canonical representation $(Q^C,\Sigma,\delta^C,\pitilde^C)$ contains a copy of $G$ in the sense that there exists a set of states $Q' \subset Q^C$, such that there exists a one-to-one map $\zeta:Q \rightarrow Q'$, with:
\cgather[5pt]{\forall q \in Q, \forall \sigma \in \Sigma, \left \{ \begin{array}{l}
\pitilde(q,\sigma) = \pitilde^C(\zeta(q),\sigma) \\
\delta(q,\sigma) = \delta^C(\zeta(q),\sigma)\end{array}\right.
}
\item If during the  construction (beginning with $\wp_\lambda$) we  encounter $\wp_x = \zeta(q)$ for some $x \in \Sigma^\star$,  $q \in Q$ and any map $\zeta$ as defined in (2), then we stay within the graph of the copy of the  initial-marked PFSA  for all right extensions of $x$.
\end{enumerate}

\end{lem} 
\Proof{
(1) follows  the ergodicity of QSPs, which makes $\wp_\lambda$ independent of the initial state in the initial-marked PFSA.

(2) The canonical representation  subsumes the initial-marked representation in the  sense that the states of the latter may themselves be seen as degenerate distributions over $Q$, $i.e.$, by letting 
\cgather{\label{eqE}
\mathcal{E}=\big \{ e^i \in [0 \ 1]^{\vert Q \vert } , i= 1,\cdots, \vert Q \vert \big \}
}
  denote  the set of distributions satisfying:
\cgather{
e^i\vert_j = \begin{cases}
              1, & \textrm{if } i=j\\
              0, & \textrm{otherwise}
             \end{cases}
}
(3) follows from the strong connectivity of $G$.
}



Lemma~\ref{lemcanonstruc} implies that initial states are unimportant;   we may denote the initial-marked PFSA induced by a QSP $\mathcal{H}$, with the initial marking removed, as $\mathcal{P}_\mathcal{H}$, and refer to it simply as a ``PFSA''.  States in $\mathcal{P}_\mathcal{H}$ are representable as  states in $\mathcal{C}_\mathcal{H}$ as elements of $\mathcal{E}$. Note that  we always  encounter a state arbitrarily close to some element in  $\mathcal{E}$ in the canonical construction starting from the stationary distribution $\wp_\lambda$ on the states of $\mathcal{P}_\mathcal{H}$. However, before we go further, we establish the existence of unique minimal realizations. Note that even if initial-marked PFSAs are strongly connected, the canonical representations might not be. 
%
\DETAILS{
\begin{defn}[Structural Isomorphism Between PFSA]\label{defstructiso}
PFSAs $G=(Q,\Sigma,\delta,\pitilde)$ and $ G'=(Q',\Sigma,\delta',\pitilde')$, defined on the same alphabet $\Sigma$,  are  structurally isomorphic if there exists a bijective mapping $\xi\colon Q \to Q'$ such that:
\cgather{
\forall q \in Q, \sigma \in \Sigma, \left \{ \begin{array}{l} \xi(\delta(q, \sigma)) = \delta'(\xi(q),\sigma)\\
\pitilde(q,\sigma) = \pitilde'(\xi(q),\sigma)\end{array}\right.
}
Note, bijectivity of $\xi$ requires $\vert Q\vert = \vert Q'\vert$.
\end{defn}
Structural isomorphism between two PFSA implies  that there exists a permutation of the states such that one is transformed to the other. Thus, structurally isomorphic PFSAs encode the same QSP. 

\begin{thm}[Existence Of Unique Strongly Connected Minimal Realization]\label{thmminrez}
 If the probabilistic Nerode relation corresponding to a  probability space $(\Sigma^\omega,\mathfrak{B}, \mu)$, representing a stationary ergodic QSP,  has a finite index, then it has a  strongly connected PFSA generator unique upto structural isomorphism.
\end{thm}
\Proof{
First we use the construction described in Lemma~\ref{lemPROB2PFSA} to obtain a PFSA generator $G=(Q,\Sigma,\delta,\pitilde)$ for the  finite index Nerode relation $\sim_N$  corresponding to a  probability space $(\Sigma^\omega,\mathfrak{B}, \mu)$. Note that since the QSP that the probability space represents is ergodic, we can drop the initial state from the construction of Lemma~\ref{lemPROB2PFSA}.

Let $G'=(Q',\Sigma,\delta\vert_{Q'},\pitilde\vert_{Q'})$ be a strongly connected component of $G$, $i.e.$, we have  $Q' \subseteqq Q$, and $\delta\vert_{Q'}, \pitilde\vert_{Q'}$ are the restriction of the corresponding functions  to the possibly smaller set of states, and $(Q',\delta\vert_{Q'})$ defines a strongly connected graph with $Q'$ as the set of nodes, and there is a labeled  edge $q_i \xrightarrow{\sigma} q_j$ iff $\delta\vert_{Q'}(q_,\sigma) = q_j$.

Let $q_0' \in Q'$, such that $\exists x_0 \in \Sigma^\star$,  $[x_0] = q_0'$, such that $\mu(x_0\Sigma^\omega) > 0$. Let $H$ be an initial marked PFSA obtained by augmenting $G'$ with $q_0'$ as the initial state, $i.e.$ $H=(Q',\Sigma,\delta\vert_{Q'},\pitilde\vert_{Q'},q'_0)$. 
Let us denote:
\cgather{
\mathcal{E}_H = \{[x_0 y]: y \in \Sigma^\star  \}
}
Let $\mathcal{E}$ be the set of equivalence classes of $\sim_N$. It is immediate that:
\cgather{
\mathcal{E}_H \subseteqq \mathcal{E}
}
Also, since $H$ is strongly connected, and any right extension of $x_0$ terminates on some state $q \in Q'$, it is immediate that there exists bijective map $\mathds{H}: Q' \to \mathcal{E}_H$.

Let if possible there exist $E \in \mathcal{E}$ such that $E \notin \mathcal{E}_H$. And let $z \in \Sigma^\star$, such that $z \in E$. Then, it follows that 
\cgather{
\forall z' \in \Sigma^\star, x_0z' \nsim_N z
}
which contradicts our assumption that the QSP is ergodic. Hence, we conclude that $\mathcal{E} = \mathcal{E}_H$, $i.e.$, $H$ is a generator for $\sim_N$. 

We claim that the map $\mathds{H}: Q' \to \mathcal{E}$ is injective. To see this, assume if possible that  for some distinct  $q_1,q_2 \in Q'$:
\cgather{
\mathds{H}(q_1)  = \mathds{H}(q_2) = E \in \mathcal{E} \label{eqASS}
}
 Since $q_1,q_2$ are distinct, there exist strings $x_1,x_2   \in \Sigma^\star$ such that $[x_i] \neq [x_2]$, which contradicts Eq.~\eqref{eqASS}. Hence, we conclude that $H$ is a minimal realization. Since $G'$ is an arbitrary strongly connected component of $G$, and the above argument is valid for any permutation of the state labels, we conclude that $H$ is unique upto structural equivalence. This completes the proof.
}

To summarize,  every PFSA $G=(Q,\Sigma,\delta,\pitilde)$ represents a probability space $(\Sigma^\omega, \mathfrak{B}, \mu)$ for any fixed initial state, and  there is always a minimal realization that encodes the  latter; however $G$ could possibly be a non-minimal realization of the underlying probability space. Thus, given a PFSA $G=(Q,\Sigma,\delta,\pitilde)$, and a choice of an initial state $q_0$,  we have two associated equivalence relations on $\Sigma^\star$:

\begin{enumerate}
\item The \textit{transition equivalence} $\sim_G$ defined by the graph of the PFSA, $i.e.$ its transition structure and its states:
\cgather{
x \sim_G y \textrm{ if } \delta(q_0, x) = \delta(q_0,y)
}
\item The probabilistic Nerode equivalence $\sim_N$  given by:
\cgather{
x \sim_N y \textrm{ if } \forall z \in \Sigma^\star, \mu(xz \Sigma^\omega) = \mu(yz \Sigma^\omega)
}

\end{enumerate}
We have the following immediate result:
\begin{lem}[Transition Equivalence]\label{lemtranseq}
Given a PFSA $G=(Q,\Sigma,\delta,\pitilde)$, and a choice of an initial state $q_0 \in Q$, the transition equivalence is necessarily a refinement of the corresponding Nerode equivalence. The two equivalences are identical if $G$ is a minimal realization
\end{lem}
\Proof{
Follows immediately by noting that:
\mltlne{
x \sim_G y \Rightarrow \delta(q_0, x) = \delta(q_0,y)  \Rightarrow \forall z \in \Sigma^\star, \delta(q_0, xz) = \delta(q_0,yz)\\ \Rightarrow \forall z \in \Sigma^\star,\mu(xz \Sigma^\omega) = \mu(yz \Sigma^\omega)
}
}

}

Next we introduce the notion of $\epsilon$-synchronization of probabilistic automata (See Figure~\ref{figsync}). Synchronization of automata is fixing or determining the current state; thus it is analogous to contexts in Rissanen's ``context algorithm''~\cite{rissanen83}. We show that while not all PFSAs are synchronizable,  all are $\epsilon$-synchronizable.
\begin{figure}[t]
\centering 
\includegraphics[width=3in]{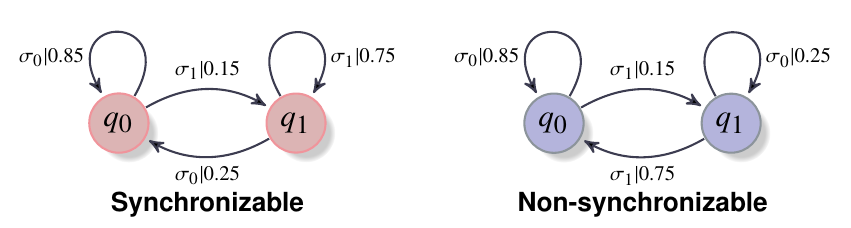}
\vspace{-5pt}

\captionN{{\bf Synchronizable and non-synchronizable machines.} Identifying contexts is a key step in estimating the entropy rate of stochastic signals sources; and for PFSA generators, this translates to a state-synchronization problem. However, not all PFSAs are synchronizable, $e.g.$, while the top machine is synchronizable, the bottom one is not. Note that  a history of just one symbol suffices to determine the current state in the synchronizable machine (top), while no finite history can do the same in the non-synchronizable machine (bottom). However, we show that a $\epsilon$-synchronizable string always exists (Theorem~\ref{thmepssynchro}).
}\label{figsync} 
\vspace{-15pt}

\end{figure}

\begin{thm}[$\epsilon$-Synchronization of Probabilistic Automata]\label{thmepssynchro}
 For any QSP $\mathcal{H}$ over  $\Sigma$, the PFSA  $\mathcal{P}_\mathcal{H}$ satisfies: 
\cgather{
\forall \epsilon' > 0, \exists x \in \Sigma^\star, \exists\bvec  \in \mathcal{E},  \vert \vert \wp_x -\bvec \vert \vert_\infty \leqq \epsilon'\label{eqsync}
}
\end{thm}
%

\Proof{
 We show that all PFSA are at least approximately synchronizable~\cite{BICP99,Ito84}, which is not true for deterministic automata. If the graph of $\mathcal{P}_\mathcal{H}$ ($i.e.$, the deterministic automaton obtained by removing the arc probabilities) is synchronizable, then Eq.~\eqref{eqsync} trivially holds true for $\epsilon' = 0$ for any synchronizing string $x$. Thus, we assume   the graph of $\mathcal{P}_\mathcal{H}$ to be non-synchronizable. From definition of non-synchronizability, it follows:
\cgather{
\forall q_i,q_j \in Q, \textrm{with } q_i \neq q_j,  \forall x \in \Sigma^\star, \delta(q_i,x) \neq \delta(q_j,x)
}
If the PFSA has a single state, then every string satisfies the condition in Eq.~\eqref{eqsync}. Hence, we assume that the PFSA has more than one state.
Now if we have:
\cgather{
\forall x \in \Sigma^\star, \frac{Pr(x'x)}{Pr(x')} = \frac{Pr(x''x)}{Pr(x'')} \textrm{ where } [x']= q_i, [x'']=q_j
}
then, by the Definition~\ref{defnerode} , we have a contradiction $q_i =q_j$. Hence  $\exists x_0$ such that 
\cgather{
\frac{Pr(x'x_0)}{Pr(x')} \neq \frac{Pr(x''x_0)}{Pr(x'')} \textrm{ where } [x']= q_i, [x'']=q_j \\
\mathrm{Since: } 
\sum_{x \in \Sigma^\star} \frac{Pr(x'x)}{Pr(x')}=1, \textrm{ for any } x' \textrm{ where }  [x']= q_i
}
we conclude without loss of generality  $\forall q_i, q_j \in Q$, with $q_i \neq q_j$:
\cgathers{
\exists x^{ij} \in \Sigma^\star, \frac{Pr(x'x^{ij})}{Pr(x')} > \frac{Pr(x''x^{ij})}{Pr(x'')} \textrm{ where } [x']= q_i, [x'']=q_j
}
It follows from induction that if we start with a distribution $\wp$ on $Q$ such that $\wp_i = \wp_j = 0.5$, then for any $\epsilon' > 0$ we can construct a finite string $x^{ij}_0$ such that if $\delta(q_i,x^{ij}_0) = q_r, \delta(q_j,x^{ij}_0) = q_s$, then for the new distribution $\wp'$ after execution of $x^{ij}_0$ will satisfy
$\wp_s' > 1-\epsilon'$. Recalling that $\mathcal{P}_\mathcal{H}$ is strongly connected, we note that,  for any $q_t \in Q$, there exists a string $y \in \Sigma^\star$, such that $\delta(q_s,y) = q_t$. Setting $x^{i,j\rightarrow t}_\star = x^{ij}_0 y$, we can ensure that the distribution $\wp''$ obtained after execution of $x^{ij}_\star$ satisfies $\wp_t'' > 1-\epsilon'$ for any $q_t$ of our choice. For arbitrary initial distributions $\wp^A$ on $Q$, we must consider contributions arising from simultaneously executing $x^{i,j\rightarrow t}_\star$ from states other than just $q_i$ and $q_j$. Nevertheless, it is easy to see that executing $x^{i,j\rightarrow t}_\star$ implies that in the new distribution $\wp^{A'}$, we have
$ \wp^{A'}_t > \wp^A_i +\wp^A_j -\epsilon'$. It  follows that executing  the string $x^{1,2\rightarrow \vert Q\vert}x^{3,4\rightarrow \vert Q\vert} \cdots x^{n-1,n\rightarrow \vert Q\vert}$, where
\cgather{
n = \begin{cases}
  \vert Q\vert & \textrm{if $\vert Q\vert$ is even}\\
\vert Q \vert -1 & \textrm{otherwise}   
    \end{cases}
}
would result in a final distribution $\wp^{A''}$ which satisfies $\wp^{A''}_{\vert Q \vert } > 1 - \frac{1}{2}n \epsilon'$.
Appropriate scaling of $\epsilon'$ then completes the proof.
}
Theorem~\ref{thmepssynchro} induces the notion of  $\epsilon$-synchronizing strings, and guarantees their existence for arbitrary PFSA.
\begin{defn}[$\epsilon$-synchronizing Strings]\label{defepsilonsynchro}
  A  string $x\in \Sigma^\star$ is $\epsilon$-synchronizing for a PFSA if:
\cgather{
\exists\bvec  \in \mathcal{E}, \vert \vert \wp_x -\bvec  \vert \vert_\infty \leqq \epsilon
}
\end{defn}

Theorem~\ref{thmepssynchro} is an existential result, and  does not yield an algorithm for computing synchronizing strings (See Theorem~\ref{thmderivheap}). 
We may  estimate an asymptotic upper bound on such a search.

\begin{cor}[To Theorem~\ref{thmepssynchro}]\label{corsynchrodepth}
At most $O(1/\epsilon)$ strings from the lexicographically ordered set of all strings over the given alphabet need to be analyzed to find an $\epsilon$-synchronizing string.
\end{cor}
\Proof{
Theorem~\ref{thmepssynchro}  multiplies entries from the $\Pitilde$ matrix, which cannot be all identical (otherwise the states would collapse). Let the minimum difference between two unequal entries be $\eta$. Then, following the construction in Theorem~\ref{thmepssynchro},  the length $\ell$ of the synchronizing string, up to linear scaling, satisfies: $\eta^\ell = O(\epsilon)$, implying $\ell = O(log(1/\epsilon)$. Hence, the number of strings to be analyzed  is at most all strings of length $\ell$,  where
$ \vert \Sigma\vert^\ell = \vert \Sigma\vert^{O(log(1/\epsilon)} = O(1/\epsilon)$.
}
%
\subsection{Symbolic Derivatives}
Computation of $\epsilon$-synchronizing strings requires the notion of symbolic derivatives. PFSA states  are not  observable; we observe  symbols generated from hidden states. A symbolic derivative at a given string  specifies the distribution of the next symbol over the alphabet.

\begin{notn}
We denote the set of probability distributions 
over a finite  set of cardinality $k$ as $\mathscr{D}(k)$. 
\end{notn}

\begin{defn}[Symbolic Count Function]\label{defcount}
 For a string $s$ over  $\Sigma$, the count function $\#^s: \Sigma^\star \rightarrow \mathbb{N}\cup \{0\}$,  counts the number of times a particular substring occurs in $s$. The count is overlapping, $i.e.$, in a string $s=0001$, we count the number of occurrences of $00$s as $\underline{00}01$ and $0\underline{00}1$, implying $\#^s 00 =2$.
\end{defn}

\begin{defn}[Symbolic Derivative]\label{defsymderivative}
 For a string $s$  generated by a QSP over $\Sigma$, the symbolic derivative  $\phi^s:\Sigma^\star \rightarrow \mathscr{D}(\vert \Sigma\vert -1)$ is defined:
\vspace{-5pt}
\cgather{
\phi^s(x) \big \vert_i = \frac{\#^s x\sigma_i}{\sum_{\sigma_i \in \Sigma }\#^s x\sigma_i}
}
Thus,  $\forall x \in \Sigma^\star, \phi^s(x)$ is a probability distribution over $\Sigma$. $\phi^s(x)$ is referred to as the symbolic derivative at $x$.
\end{defn}

Note that  $\forall q_i \in Q$, $\pitilde$  induces a  probability distribution over $\Sigma$ as  $[\pitilde(q_i,\sigma_1), \cdots , \pitilde(q_i,\sigma_{\vert \Sigma \vert})]$. We denote this as $\pitilde(q_i,\cdot)$.

We next show that the symbolic derivative at $x$ can be used to estimate this distribution for $q_i = [x]$, provided $x$ is $\epsilon$-synchronizing.
\begin{thm}[$\epsilon$-Convergence]\label{thmsymderiv} If $x \in \Sigma^\star$ is $\epsilon$-synchronizing, then:
 \cgather{
\forall \epsilon > 0,  \lim_{\vert s \vert \rightarrow \infty}\vert \vert \phi^s(x) -\pitilde([x],\cdot)\vert \vert_\infty \leqq_{a.s} \epsilon\label{eqsync2} 
}
\end{thm}
\Proof{
 We use the Glivenko-Cantelli theorem~\cite{Fl70} on  uniform convergence of empirical distributions. Since $x$ is $\epsilon$-synchronizing:
\cgather{
\forall \epsilon > 0, \exists\bvec  \in \mathcal{E}, \vert \vert \wp_x -\bvec  \vert \vert_\infty \leqq \epsilon
}
Recall that $\mathcal{E}=\big \{ e^i \in [0 \ 1]^{\vert Q \vert } , i= 1,\cdots, \vert Q \vert \big \}$  denotes  the set of distributions over $Q$ satisfying:
\cgather{
e^i\vert_j = \begin{cases}
              1, & \textrm{if } i=j\\
              0, & \textrm{otherwise}
             \end{cases}
}
Let  $x$ $\epsilon$-synchronize to $q \in Q$. Thus, when we encounter $x$ while 
reading  $s$, we are guaranteed to be distributed over  $Q$ as $\wp_x$, where:
\cgather{
 \vert \vert \wp_x -\bvec  \vert \vert_\infty \leqq \epsilon
\Rightarrow \wp_x = \alpha \bvec +(1-\alpha) u
}
where $ \alpha \in [0,1]$, $\alpha \geqq 1 - \epsilon$, and $u$ is an unknown distribution over $Q$. Defining $A_\alpha = \alpha \pitilde(q,\cdot) + (1-\alpha) \sum_{j=1}^{\vert Q\vert}u_j \pitilde(q_j,\cdot)
$, we note that $\phi^s(x)$ is an empirical distribution for $A_\alpha$, implying:
 \caligns{
 &\lim_{\vert s \vert \rightarrow \infty}\vert \vert \phi^s(x) - \pitilde(q,\cdot) \vert \vert_\infty 
 =\lim_{\vert s \vert \rightarrow \infty}
\vert \vert \phi^s(x) - A_\alpha + A_\alpha 
- \pitilde(q,\cdot) \vert \vert_\infty \\
 & \leqq \overbrace{\lim_{\vert s \vert \rightarrow \infty}   \vert \vert \phi^s(x) -A_\alpha \vert\vert_\infty }^{\textrm{\scriptsize  a.s. $0$ by Glivenko-Cantelli}}  +  \lim_{\vert s \vert \rightarrow \infty}\vert \vert  A_\alpha - \pitilde(q,\cdot) \vert \vert_\infty \\
&\leqq_{a.s}  (1-\alpha) \left ( \vert \vert\pitilde(q,\cdot) - u \vert\vert_\infty  \right ) \leqq_{a.s}  \epsilon
 }
This completes the proof.
}
\begin{cor}[To Theorem~\ref{thmsymderiv}: Right Extension of $\epsilon$-Synchronizing Strings]\label{corrightinveps}
If $x \in \Sigma^\star$ is $\epsilon$-synchronizing, then $\exists \sigma \in \Sigma$, such that 
$x\sigma$ is $\epsilon'$-synchronizing with $\epsilon' = C_0 \epsilon$, and $C_0$ is the finite constant:
\cgather{\label{eqC0}
C_0 \triangleq \max_{\begin{subarray}{c}q_i,q_j \in Q, \sigma \in \Sigma \\ \text{s.t. } \pitilde(q_j,\sigma) > 0\end{subarray}} \frac{\pitilde(q_i,\sigma)}{\pitilde(q_j,\sigma)} < \infty
}
\end{cor}
\Proof{
Let $x \in \Sigma^\star$ be $\epsilon$-synchronizing.
 Definition~\ref{defepsilonsynchro} implies that:
\cgather{
\exists\bvec  \in \mathcal{E}, \vert \vert \wp_x -\bvec  \vert \vert_\infty \leqq \epsilon
}
We note that if the Nerode relation has a single equivalence class ($i.e.$ the underlying minimal PFSA has a single state), then the result holds true for every $\sigma \in \Sigma$. Hence, we assume that the minimal realization of the underlying PFSA has more than one state.

Without loss of generality, let $\bvec\vert_{i^\star} = 1$, implying (Definition~\ref{defepsilonsynchro}):
\cgather{
\wp_x\vert_{i^\star} > 1-\epsilon
}
Since we assume the underlying PFSAs to be strongly connected, there exists $\sigma' \in \Sigma$ such that $\delta(q_{i^\star},\sigma') \neq q_{i^\star}$, and $\pitilde(q_{i^\star} , \sigma') > 0$. We compute $\wp_{x\sigma'}$ explicitly, using Eq.~\eqref{eqLEP}, and note that if $\delta(q_{i^\star},\sigma') = q_{j^\star} \neq q_{i^\star}$, then we have:
\cgather {
\wp_{x\sigma'} \vert_{j^\star} = \frac{\pitilde(q_{i^\star},\sigma')(1-\epsilon)}{\pitilde(q_{i^\star},\sigma')(1-\epsilon)+ \sum_{q_k \in Q \setminus \{ q_{i^\star} \}  } \pitilde(q_k,\sigma') \epsilon_k  } \\
\textrm{where } \forall k, \epsilon_k \geqq 0, \textrm{and }\sum_{\mathclap{q_k \in Q \setminus \{ q_{i^\star} \} } } \epsilon_k = \epsilon \notag \\
\Rightarrow \wp_{x\sigma'} \vert_{j^\star} \geqq \frac{1-\epsilon}{1 + \epsilon (C-1)} \triangleq 1 -\epsilon' \\
\textrm{where } C = \max_{q_k \in Q \setminus \{ q_{i^\star} \}} \frac{\pitilde(q_k,\sigma'))}{\pitilde(q_{i^\star},\sigma')}  < \infty \notag \\
\Rightarrow \epsilon' = \epsilon \frac{C}{1 + \epsilon (C-1)} \leqq \epsilon C \leqq \epsilon C_0
}
This completes the proof.}
%
%
\begin{rem}[$C_0$ As A System Property]\label{remrightext}
It is crucial to note that the above argument remains valid for any right extension of an $\epsilon$-synchronizing string, as long as the probability of that extension being generated from the synchronized state is non-zero. Specifically, note that the argument in Corollary~\ref{corrightinveps}  is valid for any $\sigma'$ as long as the probability of generating $\sigma'$ from state $q_{i^\star}$ is non-zero (to ensure finiteness of $c$).
Also, note that $C_0$ is a property of the underlying QSP, and is independent of $x$.
This would be important in establishing the efficient PAC-learnability of QSPs with finite number of causal states using  PFSA as the hypothesis class.
\end{rem}
\subsection{Computation of  $\epsilon$-synchronizing Strings}
Next we describe identification of  $\epsilon$-synchronizing strings given a sufficiently long observed string ($i.e.$ a sample path) $s$. Theorem~\ref{thmepssynchro}  guarantees existence, and Corollary~\ref{corsynchrodepth} establishes that $O(1/\epsilon)$ substrings need to be analyzed till we encounter an $\epsilon$-synchronizing string.
%
These  do not provide an executable algorithm, which arises from  
an inspection of the geometric structure of the set of probability vectors over $\Sigma$, obtained by constructing $\phi^s(x)$ for different choices of the candidate string $x$.


\begin{defn}[Derivative Heap]\label{defderivheap}
  Given a string $s$ generated by a QSP, a derivative heap $\mathcal{D}^s: 2^{\Sigma^\star} \rightarrow \mathscr{D}(\vert \Sigma \vert -1)$ is the set of probability distributions over $\Sigma$ calculated for a  subset of strings  $L \subset \Sigma^\star$ as:
\cgather{
\mathcal{D}^s(L) = \big \{ \phi^s(x): x \in L \subset \Sigma^\star\big \}
}
\end{defn}
\begin{lem}[Limiting Geometry]\label{lemlimderiv}
 Let us define:
\cgather{
\mathcal{D}_\infty = \lim_{\vert s \vert \rightarrow \infty }\lim_{L \rightarrow \Sigma^\star} \mathcal{D}^s(L)
}
If $\mathscr{U}_\infty$ is the convex hull of $\mathcal{D}_\infty$, and $u$ is a vertex of $\mathscr{U}_\infty$, then 
\cgather{
\exists q \in Q, \textrm{such that } u=\pitilde(q,\cdot)
}
\end{lem}

\Proof{
 Recalling Theorem~\ref{thmsymderiv}, the result follows from noting that any element of $\mathcal{D}_\infty$ is a convex combination of elements from the set $\{\pitilde(q_1,\cdot), \cdots , \pitilde(q_{\vert Q\vert},\cdot)  \}$.
}

Lemma~\ref{lemlimderiv} does not claim that the number of vertices of the convex hull of $\mathds{D}_\infty$ equals the number of states, but that every vertex  corresponds to a state.  We cannot generate $\mathcal{D}_\infty$  since we 
have a finite observed string $s$, and we can  calculate $\phi^s(x)$ for a finite number of  $x$. Instead, we show that  choosing a string corresponding to the vertex of the convex hull of the  heap, constructed by considering $O(1/\epsilon)$ strings, gives us an $\epsilon$-synchronizing string with high probability.

\begin{thm}[Derivative Heap Approx.]\label{thmderivheap}
 For  $s$ generated by a QSP, let $\mathcal{D}^s(L)$ be computed with $L=\Sigma^{O(log(1/\epsilon))}$. If for  $x_0 \in \Sigma^{O(log(1/\epsilon))}$,  $\phi^s(x_0)$  is a vertex of the convex hull of $\mathcal{D}^s(L)$, then 
\cgather{
Pr(\textrm{$x_0$ is not $\epsilon$-synchronizing}) \leqq e^{-\tfrac{\vert s \vert p_0}{4}\epsilon }
}
 where $p_0$ is the probability of encountering  $x_0$ in $s$.
\end{thm}

%
\Proof{
The result follows from Sanov's Theorem~\cite{Cs84} for convex set of probability distributions. If $\vert s\vert \rightarrow \infty$, then $x_0$ is guaranteed to be $\epsilon$-synchronizing (Theorem~\ref{thmepssynchro}, and Corollary~\ref{corsynchrodepth}). 
Denoting the number of times we encounter $x_0$ in $s$ as $n(\vert s \vert)$, and since $\mathcal{D}_\infty$ is a convex set of distributions (allowing us to drop the polynomial factor in Sanov's bound), we apply Sanov's Theorem to the case of finite $s$:
\cgather{
Pr\left (KL\big (\phi^s(x_0) \parallel \wp_{x_0}\Pitilde\big ) > \epsilon \right ) \leqq e^{-n(\vert s \vert) \epsilon} 
}
where $KL(\cdot \vert \vert \cdot)$ is the Kullback-Leibler divergence~\cite{leh2005}.
Using~\cite{tsy09}:
\cgather{
\frac{1}{4}\norm{ \phi^s(x_0) - \wp_{x_0}\Pitilde }_\infty^2 \leqq  KL\left (\phi^s(x_0) \parallel \wp_{x_0}\Pitilde\right )
}
and  $n(\vert s \vert) \rightarrow \vert s \vert p_0$, where $p_0 > 0$ is the stationary probability of encountering $x_0$ in $s$, we conclude:
\calign{
&Pr\left (\frac{1}{4}\norm{ \phi^s(x_0) - \wp_{x_0}\Pitilde }_\infty^2 \leqq KL\left (\phi^s(x_0) \parallel \wp_{x_0}\Pitilde\right ) \leqq \epsilon  \right ) 
> 1-  e^{-\vert s \vert p_0 \epsilon} \notag \\
\Rightarrow &Pr\left (\frac{1}{4}\norm{ \phi^s(x_0) - \wp_{x_0}\Pitilde }_\infty^2 > \epsilon  \right )\leqq e^{-\vert s \vert p_0 \epsilon} \\ \Rightarrow &Pr\left (\norm{ \phi^s(x_0) - \wp_{x_0}\Pitilde }_\infty > 4\epsilon  \right )\leqq e^{-\vert s \vert p_0 \epsilon} \\
\Rightarrow &Pr\left (\norm{ \phi^s(x_0) - \wp_{x_0}\Pitilde }_\infty > \epsilon  \right )\leqq e^{-\tfrac{\vert s \vert p_0}{4}\epsilon } \label{eqconfi}
}
 which completes the proof.
}

\section{Probabilistic Models For  Cross-talk}\label{sec4}

Consider two ergodic stationary QSPs $\mathcal{H_A},\mathcal{H_B}$ evolving over two finite alphabets $\Sigma_A,\Sigma_B$ respectively. We make no additional assumptions   as to the properties of the two  alphabets, other than requiring that they be finite, $i.e.$,  $\Sigma_A,\Sigma_B$ may be identical, disjoint or have different cardinalities. 


The dynamical dependence of the   process $\mathcal{H_B}$ on the first process  $\mathcal{H_A}$ is assumed to be dictated by the cross-talk   map $\mathds{F}$ (defined next), which specifies the probability with which a   string might  transpire in the second process, given some specific string in the first.
\begin{notn}
Given a finite alphabet $\Sigma$, and the corresponding set of strictly infinite strings  $\Sigma^\omega$, and the $\sigma$-algebra $\mathfrak{B}$ as constructed in Definition~\ref{defsigmaalgebra}, we denote the set of all probability spaces of the form $(\Sigma^\omega, \mathfrak{B}, \mu)$ as $\mathds{P}_\Sigma$.
\end{notn}

\begin{defn}[Cross-talk Map $\mathds{F}$]\label{defdep} Given  stationary ergodic QSPs $\mathcal{H_A},\mathcal{H_B}$  over  finite alphabets $\Sigma_A,\Sigma_B$ respectively, the dependency of $\mathcal{H_B}$ on  $\mathcal{H_A}$ is determined by  the cross-talk  map $\mathds{F} \colon \{x\Sigma_A^\omega : x \in \Sigma^\star_A \} \to \mathds{P}_{\Sigma_B}$ defined as:
\cgather{
\forall x \in \Sigma_A^\star,  \mathds{F} (x \Sigma^\omega_A) = (\Sigma_B^\omega, \mathfrak{B}_B, \mu_x^{\mathds{F}})
}
where $\mathfrak{B}_B$ is the $\sigma$-algebra over $\Sigma_B$ constructed following Definition~\ref{defsigmaalgebra}. If the probability space (See Lemma~\ref{QSPtoProb}) induced by $\mathcal{H_B}$ is $(\Sigma^\omega_B, \mathfrak{B}_B, \mu^B)$, then the cross-talk map is required to satisfy the following consistency criterion:
\cgather{
\mathds{F}(\Sigma^\omega) = (\Sigma_B^\omega, \mathfrak{B}_B, \mu^{B}) \tag{Consistency}
}
Additionally, we assume that the cross-talk map is ergodic in the sense, that if $\mathds{F} (y \Sigma^\omega_A) = (\Sigma_B^\omega, \mathfrak{B}_B, \mu_y^{\mathds{F}}), \mathds{F} (xy \Sigma^\omega_A) = (\Sigma_B^\omega, \mathfrak{B}_B, \mu_{xy}^{\mathds{F}})$, then:
\cgather{
\forall x,y,z \in \Sigma^\star_A, \lim_{\vert y \vert \rightarrow \infty} \lVert\mu_y^{\mathds{F}}(z\Sigma_A^\omega) - \mu_{xy}^{\mathds{F}}(z\Sigma_A^\omega) \rVert = 0
\tag{Ergodicity}}
$i.e.$, the effect of some initial segment $x$ vanishes in the limit.
\end{defn}

The cross-talk map induces the notion of the cross-derivative. 
\begin{defn}[Cross-derivative]\label{defcrossderiv} Given two stationary ergodic QSPs $\mathcal{H_A},\mathcal{H_B}$ over  finite alphabets $\Sigma_A,\Sigma_B$ respectively, and the cross-talk map $\mathds{F}$, the cross-derivative $\phi^{\mathcal{H_A},\mathcal{H_B}}_x$ at  $x \in \Sigma^\star_A$ is a probability distribution over $\Sigma_B$ such that  if   $ \phi^{\mathcal{H_A},\mathcal{H_B}}_x = \begin{pmatrix} p_1 & \cdots & p_i & \cdots \end{pmatrix}^T$, then   the next symbol in  $\mathcal{H_B}$ is  $\sigma_i$ with probability $p_i$, given that the string transpired in  $\mathcal{H_A}$ is $x$.
\end{defn}

\begin{lem}[Explicit Expression For Cross-derivative]\label{lemExpXd}
 Given  stationary ergodic QSPs $\mathcal{H_A},\mathcal{H_B}$ over  finite alphabets $\Sigma_A,\Sigma_B$ respectively,  the cross-talk map $\mathds{F}$, and assuming  $\mathcal{H_B}$ has a  PFSA representation $(Q_B,\Sigma_B,\delta_B,\pitilde_B)$,  we have:
\cgather{
\forall \sigma_i' \in \Sigma_B, 
\phi^{\mathcal{H_A},\mathcal{H_B}}_x \vert_i = \sum_{\mathclap{\tau \in \Sigma^\star_B}} \mu^\mathds{F}\left (\tau\Sigma^\omega_B \right ) \pitilde_B\left ([\tau],\sigma_i'\right )
}
\end{lem}
\Proof{
For any $\tau \in \Sigma^\star_B$, $\mu^\mathds{F}\left (\tau\Sigma^\omega_B \right )$ is the probability that it transpired in $\mathcal{H_B}$ given a string $x \in \Sigma^\star_A$. Recalling that the terminal state or the equivalence class corresponding to  $\tau$ is denoted by $[\tau]$, we note that the probability of generating $\sigma' \in \Sigma_B$ after $\tau$ is given by  $\pitilde_B\left ([\tau],\sigma_i'\right )$. The result  follows from noting that the $i^\textrm{th}$ entry of the cross-derivative at $x$ is the expected probability of generating $\sigma_i'$ next  in  $\mathcal{H_B}$.
}
\begin{figure*}[t]
\centering
\includegraphics[width=6in]{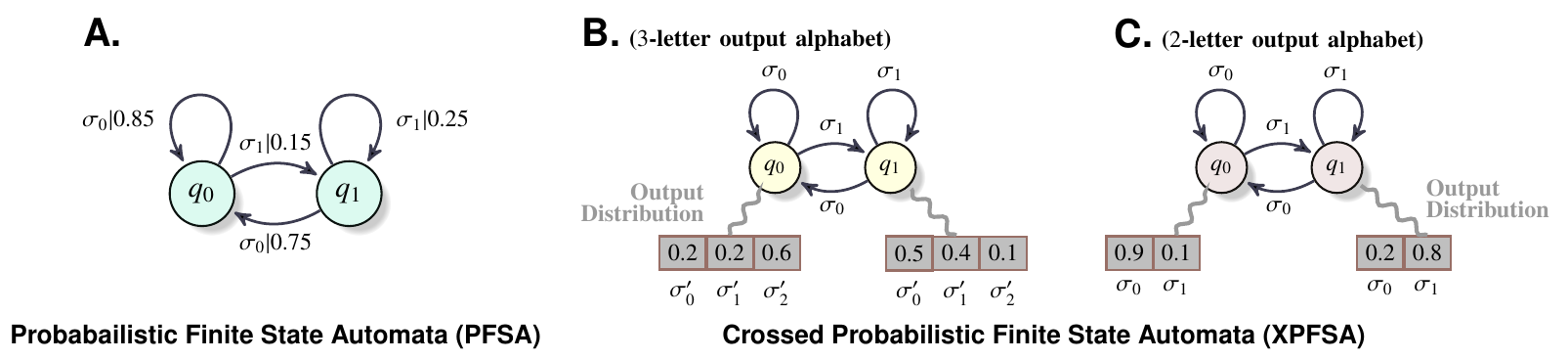}
\vskip 1em
\includegraphics[width=4.5in]{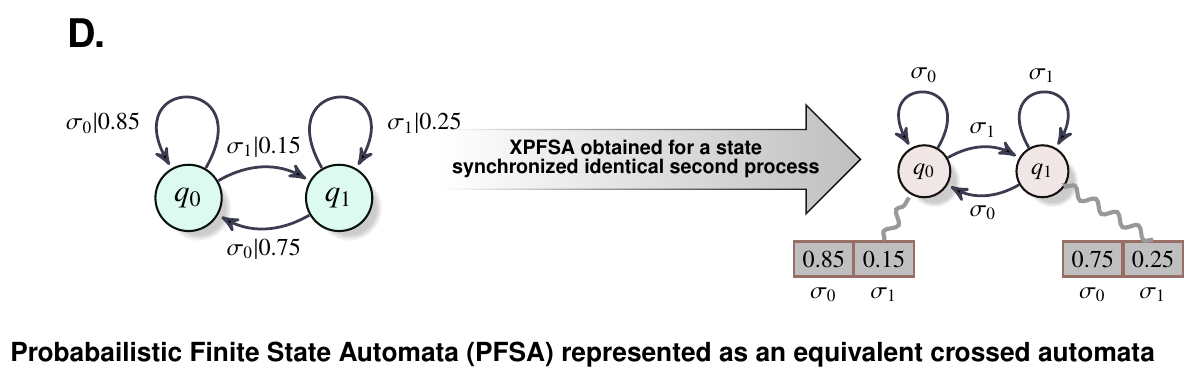}
\captionN{\textbf{Illustration of Crossed probabilistic Finite State Automata.} Plate A illustrates a PFSA, while plates B and C illustrate crossed automata capturing dependency between two ergodic stationary processes. The machine in plate B captures dependency from  a process evolving over a $2$-letter alphabet $\{\sigma_0,\sigma_1\}$ to a process evolving over a $3$-letter alphabet $\{\sigma_0',\sigma_1'\sigma_2'\}$. Plate C illustrates a XPFSA capturing dependency between processes over the same $2$-letter alphabets $\{\sigma_0,\sigma_1\}$. Note that XPFSAs differ structurally from PFSAs; while the latter have probabilities associated with the transitions between states, the former have no such specification. XPFSAs on the other hand have an output distribution at each state, and hence each time a new state is reached, XPFSAs may be thought to generate an output symbol drawn from the state-specific output distribution. It follows that one could represent a PFSA as a XPFSA as shown in plate D. The output distribution in this case must be on the same alphabet, and the probability of outputting a particular symbol at a particular state would be the generation probability of that symbol from that state. }\label{figcrossa}
\end{figure*}
\begin{cor}[To Lemma~\ref{lemExpXd}]\label{corExpXd} The cross-derivative at the empty string may be expressed as:
\cgather{
\phi^{\mathcal{H_A},\mathcal{H_B}}_\lambda = \wp^B_\lambda \Pitilde_B
}
where $\wp^B_\lambda$ is the unique stationary distribution on the states of a PFSA representation for $\mathcal{H_B}$.
\end{cor}
\Proof{
Using the expression given by Lemma~\ref{lemExpXd}, we have:
\calign{
\forall \sigma_i' \in \Sigma_B, \phi^{\mathcal{H_A},\mathcal{H_B}}_\lambda \vert_i & =  \sum_{\mathclap{\tau \in \Sigma^\star_B}} \mu^\mathds{F}\left (\tau\Sigma^\omega_B \right ) \pitilde_B\left ([\tau],\sigma_i'\right ) \\
& = \sum_{\mathclap{\tau \in \Sigma^\star_B}} \mu^B\left (\tau\Sigma^\omega_B \right ) \pitilde_B\left ([\tau],\sigma_i'\right )  \tag{Using consistency of $\mathds{F}$, see Definition~\ref{defdep}}\\
& = \sum_{\mathclap{\textrm{Equivalence Classes of $\mathcal{H_B}$}}} Pr([\tau]) \pitilde_B\left ([\tau],\sigma_i'\right )=  \left ( \wp^B_\lambda \Pitilde_B \right ) \bigg \vert_i
}
}

We intend to model cross-dependency between QSPs with objects similar to probabilistic automata; to that effect we need to formalize a notion of states in this context. As before, we realize this by defining an appropriate equivalence relation, which we call the probabilistic cross-Nerode relation.
\begin{defn}[Probabilistic Cross-Nerode Equivalence Relation]\label{defxnerode}
Given  stationary ergodic QSPs $\mathcal{H_A},\mathcal{H_B}$ over  finite alphabets $\Sigma_A,\Sigma_B$ respectively, and  the cross-talk map $\mathds{F}$, the cross-Nerode equivalence on $\Sigma_A^\star$, denoted as $\displaystyle\sim_\mathcal{H_B}^\mathcal{H_A}$, is defined as:
\cgather{
\forall x, y \in \Sigma^\star_A, x \sim_\mathcal{H_B}^\mathcal{H_A} y \notag \\ \textrm{ if } \forall z \in \Sigma^\star_A, \phi^{\mathcal{H_A},\mathcal{H_B}}_{xz} = \phi^{\mathcal{H_A},\mathcal{H_B}}_{yz}
}
\end{defn}

Clearly, the cross-Nerode equivalence is right-invariant, $i.e$,
\cgather{
\forall x, y \in \Sigma^\star_A, x \sim_\mathcal{H_B}^\mathcal{H_A} y \Rightarrow \forall z \in \Sigma^\star_A,  xz \sim_\mathcal{H_B}^\mathcal{H_A} yz
}
which induces a notion of states, in the sense that that if two strings are equivalent, we can forget which one was the actual history. This leads us to the notion of crossed probabilistic finite state automata as logical machines to represent cross-dependencies.

\subsection{Crossed Probabilistic Finite State Automata (XPFSA)}
A crossed automata has an input alphabet and an output alphabet, and the idea is to model a finite state probabilistic transducer, that maps strings over the input alphabet  to a distribution over set of finite strings over the output alphabet. Recall that these  alphabets  need not be identical with respect to their elements or  their cardinalities. Formally, we define:

\begin{defn}[Crossed Probabilistic Finite State Automata]\label{defxpfsa}
A crossed probabilistic finite state automaton (XPFSA) is a $4$-tuple $G\equiv (Q,\Sigma,\delta,\pitilde_{\Sigma'})$, where $Q$ is a finite set of states, $\Sigma$ is a finite alphabet of symbols (known as the input alphabet), $\delta: Q \times \Sigma^\star \rightarrow Q$ is the recursively extended transition function, $\Sigma'$ is a finite output alphabet with possibly $\Sigma \neq \Sigma'$, and $\pitilde_{\Sigma'}:Q \times \Sigma' \rightarrow [0,1]$ is the output morph function parameterized by the output alphabet $\Sigma'$. In particular,  $\pitilde_{\Sigma'}(q,\sigma')$ is  the probability of generating $\sigma' \in \Sigma'$ from a state $q \in Q$, and hence: $\forall q \in Q, \sum_{\sigma' \in \Sigma'}\pitilde_{\Sigma'}(q,\sigma') = 1$.

A XPFSA with a marked initial state $q_0 \in Q$, is an initial-marked XPFSA, which is described by the augmented quintuple $(Q,\Sigma,\delta,\pitilde_{\Sigma'}, q_0)$.
\end{defn}

\begin{lem}[Cross-Nerode Equivalence to initial-marked XPFSA]\label{lemnerodetoxpfsa}
A cross-Nerode equivalence relation with finite index may be encoded by  a XPFSA. 
\end{lem}
\Proof{
For  stationary ergodic QSPs $\mathcal{H_A},\mathcal{H_B}$ over  finite alphabets $\Sigma_A,\Sigma_B$ respectively, let  $Q$ be the set of equivalence classes of the finite index cross-Nerode relation  $\displaystyle\sim_\mathcal{H_B}^\mathcal{H_A}$ (Definition~\ref{defxnerode}),  and noting that $\vert Q\vert < \infty$, define functions $\delta:Q \times \Sigma_A \rightarrow Q$, $\pitilde:Q \times \Sigma_B \rightarrow [0,1]$ as:
\begin{subequations}
\calign{
\forall x \in \Sigma_A^\star, 
& \delta([x],\sigma) = [x\sigma]\\
\forall x \in \Sigma_A^\star, \sigma_i' \in \Sigma_B, & \pitilde_{\Sigma_B}([x],\sigma_i') = \phi^{\mathcal{H_A},\mathcal{H_B}}_{x'} \big \vert_i \textrm{ for any choice of } x' \in [x] \label{eqpit}
}
\end{subequations}
where we  extend $\delta$  recursively  to   $y=\sigma x \in \Sigma^\star$ as 
\cgather{\delta(q,\sigma x) = \delta(\delta(q,\sigma),x)}
Denoting  $[\lambda]$ as $q_0$, we have an  encoding for  $\displaystyle\sim_\mathcal{H_B}^\mathcal{H_A}$ as a initial-marked XPFSA $(Q,\Sigma_A,\delta,\pitilde_{\Sigma_B},q_o)$.
}

Using the same argument from ergodicity as used in eliminating the initial making for PFSAs, we note that $q_0$ can be dropped without any loss of information. Later we argue that 
XPFSA have unique (upto state renaming) minimal realizations, with strongly connected graphs.

Figure~\ref{figcrossa} illustrates the differences between  PFSAs and  XPFSAs. Note that inter-state transitions have generation probabilities in addition to symbol labels in the PFSA, while in the XPFSA the transitions are labeled with only  symbols from the input alphabet. The XPFSAs on the other hand have an output distribution over at each state. This output distribution is over the output alphabet, and as shown in Figure~\ref{figcrossa} plates B and C, the size of the output alphabet (or its elements) may be different from that of the input alphabet.

\subsubsection{Specific cases: No Dependence \& Identical Sample Paths}
Next we investigate some specific  dependencies that may arise  between stationary ergodic  processes. The first case is when there is  no dependence, $e.g.$, when the evolution of  $\mathcal{H_B}$ cannot be predicted to any degree from a knowledge of $\mathcal{H_A}$ evolution.
\begin{thm}[XPFSA Structure: First Result]\label{thmindXPFSA}
For  stationary ergodic QSPs $\mathcal{H_A},\mathcal{H_B}$ over  finite alphabets $\Sigma_A,\Sigma_B$ respectively, the following statements are equivalent:
\caligns{
& 1)  && \forall x, y \in \Sigma^\star_A, x \sim_\mathcal{H_B}^\mathcal{H_A} y \\
&2)  && \forall x \in \Sigma_A^\star, \phi^{\mathcal{H_A},\mathcal{H_B}}_{x} = v \textrm{ where $v$ is a constant vector}\\
&3)  &&\forall x \in \Sigma_A^\star, \phi^{\mathcal{H_A},\mathcal{H_B}}_{x} = \wp^B_\lambda \Pitilde_B
}
\end{thm}
\Proof{
$1) \rightarrow 2)$: Let  $x,y \in \Sigma^\star_A$. From Definition~\ref{defxnerode}, we have:
\cgather{
 x \sim_\mathcal{H_B}^\mathcal{H_A} y \notag \Rightarrow  \forall z \in \Sigma^\star_A, \phi^{\mathcal{H_A},\mathcal{H_B}}_{xz} = \phi^{\mathcal{H_A},\mathcal{H_B}}_{yz}
}
Setting $z = \lambda$, and using $1)$ we conclude that $\phi^{\mathcal{H_A},\mathcal{H_B}}_{x}$ must be a constant vector for all $x \in \Sigma^\star_A$.

$1) \rightarrow 2)$: Follows from Definition~\ref{defxnerode}.

$3) \rightarrow 2)$: Trivial.

$2) \rightarrow 3)$: Setting $x = \lambda$, and using Corollary~\ref{corExpXd} to  Lemma~\ref{lemExpXd}:
\cgather{
\phi^{\mathcal{H_A},\mathcal{H_B}}_\lambda = v = \wp^B_\lambda \Pitilde_B
}
This completes the proof.
}

Theorem~\ref{thmindXPFSA} essentially identifies the structure of the XPFSA when $\mathcal{H_B}$ evolves independently from the first process $\mathcal{H_A}$, $i.e.$, knowledge of the transpired string in the latter does not affect the next-symbol distribution in $\mathcal{H_B}$. This is precisely what is stated in $2)$: the cross-derivative is a constant vector for any string $x \Sigma^\star_A$. Theorem~\ref{thmindXPFSA} establishes that in this situation, the minimal XPFSA is a single-state machine, and the output distribution from this single state is given by $\wp^B_\lambda \Pitilde_B$.

The simplest non-trivial dependency arises for the case of state synchronized  copies of the same process. More specifically, given a sample path from any quantized stochastic ergodic stationary  process, we can 
consider of the single-step right-shifted  sample path as a second process.   The XPFSA obtained in this case (See Figure~\ref{figcrossa}, plate D) is trivial to calculate:

\subsubsection{The Notion of Directional (In)Dependence}

The standard definition of independence between stochastic processes is as follows:

\begin{defn}[Independence of Two Stochastic Processes]\label{defIndStochProc}
Stochastic processes $\{ X(t)\}, t \in T $  and $ \{Y(t)\}, t \in T $ are independent if for any $n \in \mathbb{N}$, and $t_i, \cdots, t_n \in T$, the random vectors $X \triangleq \{X(t_1), \cdots, X(t_n)\}$ and $Y \triangleq \{Y(t_1), \cdots, Y(t_n)\}$ are independent.
\end{defn}
\begin{lem}[Independence Implies Trivial XPFSA]\label{lemInd2XPFSA}
For  stationary ergodic QSPs $\mathcal{H_A},\mathcal{H_B}$ over  finite alphabets $\Sigma_A,\Sigma_B$ respectively, if $\mathcal{H_A}$ and $\mathcal{H_B}$ are independent (in the sense of Definition~\ref{defIndStochProc}), then we have:
\cgather{
\left ( \forall x, y \in \Sigma^\star_A, x \sim_\mathcal{H_B}^\mathcal{H_A} y \right )  \bigwedge \left (\forall x', y' \in \Sigma^\star_B, x' \sim_\mathcal{H_A}^\mathcal{H_B} y' \right ) 
}
$i.e.$, for independent processes, the minimal XPFSA in both directions has a single state.
\end{lem}
\Proof{
Consider the sample paths generated by the processes $\mathcal{H_A},\mathcal{H_B}$  represented by the symbol sequences $s^A,s^B$ over the respective alphabets, and let the $k^{th}$ symbol in the streams be represented as $s^A_k,s^B_k$.
Fix $n \in N$, and consider the random vectors $V\triangleq \{s^A_0, \cdots,s^A_{n-1}\}, W\triangleq \{s^B_0, \cdots,s^B_{n-1}\}$. Also, let $Z_{n}$ denote the random variable for the $n^{\textrm{th}}$ symbol in $s^B$. Then, independence implies:
\mltlne{
\forall x \in \Sigma^\star_A, z \in \Sigma^\star_B, \sigma' \in \Sigma^B,  \\   Pr(V = x, W= z, Z_n = \sigma') =\\ Pr(V=x)Pr(W=z, Z_n = \sigma') 
}
Let a PFSA description of $\mathcal{H_B}$ be $(Q,\Sigma_B, \delta, \pitilde^B)$.
Assuming the canonical description for the ergodic process $\mathcal{H_B}$, without loss of generality, we fix the initial state as the stationary distribution $\wp_\lambda^B$ (See Definition~\ref{defcanon}). Then, by marginalizing out $W$, we get:
\mltlne{
\forall x \in \Sigma^\star_A, \sigma' \in \Sigma^B,    Pr(V = x,  Z_n = \sigma') =\\ Pr(V=x)\sum_{q_i \in Q}\wp^B_\lambda \big \vert_i \pitilde^B(q_i, \sigma')
}
which, then implies that:
\cgather{
\forall x \in \Sigma_A^\star, \phi^{\mathcal{H_A},\mathcal{H_B}}_{x} = \wp^B_\lambda \Pitilde_B
}
Similarly, using the argument from $\mathcal{H_B}$ to $\mathcal{H_A}$, we get 
\cgather{
\forall y \in \Sigma_B^\star, \phi^{\mathcal{H_B},\mathcal{H_A}}_{y} = \wp^A_\lambda \Pitilde_A
}
which then completes the proof using Lemma~\ref{thmindXPFSA}.
}
\begin{lem}[Unidirectional Dependence $\neq$ Independence]\label{lemunidep}
There exist dependent stationary ergodic QSPs $\mathcal{H_A},\mathcal{H_B}$ over  finite alphabets $\Sigma_A,\Sigma_B$ respectively, such that:
$
 \forall x, y \in \Sigma^\star_A, x \sim_\mathcal{H_B}^\mathcal{H_A} y 
$, $i.e.$, the minimal XPFSA from $\mathcal{H_A}$ to $\mathcal{H_B}$ has a single state.
\end{lem}
\Proof{ We give an explicit example.
Consider the processes $\mathcal{H_A},\mathcal{H_B}$ over a binary alphabet $\{0,1\}$ generating sample paths $s^A,s^B$ respectively via the following recursive rules (where  $s^A_k,s^B_k$ are the symbols at step $k$) :
\cgather{
s^B_{0}=0, s^A_{0}=0\\
Pr(s^A_{k+1}=0 \vert s^B_{k}=0) = 0.8,
Pr(s^A_{k+1}=1 \vert s^B_{k}=0) = 0.2\\
Pr(s^A_{k+1}=0 \vert s^B_{k}=1) = 0.2,
Pr(s^A_{k+1}=1 \vert s^B_{k}=1) = 0.8\\
Pr(s^B_{k+1}=0 \vert s^A_{k} \in\{0,1\} ) = 0.5,
Pr(s^B_{k+1}=1 \vert s^A_{k} \in\{0,1\} ) = 0.5 \label{eq54}
}
It is immediate from Eq.~\eqref{eq54} that $
 \forall x, y \in \Sigma^\star_A, x \sim_\mathcal{H_B}^\mathcal{H_A} y 
$, and hence it follows from Lemma~\ref{thmindXPFSA} that the minimal XPFSA from $\mathcal{H_A}$ to $\mathcal{H_B}$ has a single state. And since the current  symbol in $s^B$ dictates the distribution of the next symbol in $s^A$, it is also immediate that the processes are not independent. 
}

\DETAILS{
\begin{figure*}
\centering

\includegraphics[width=5in]{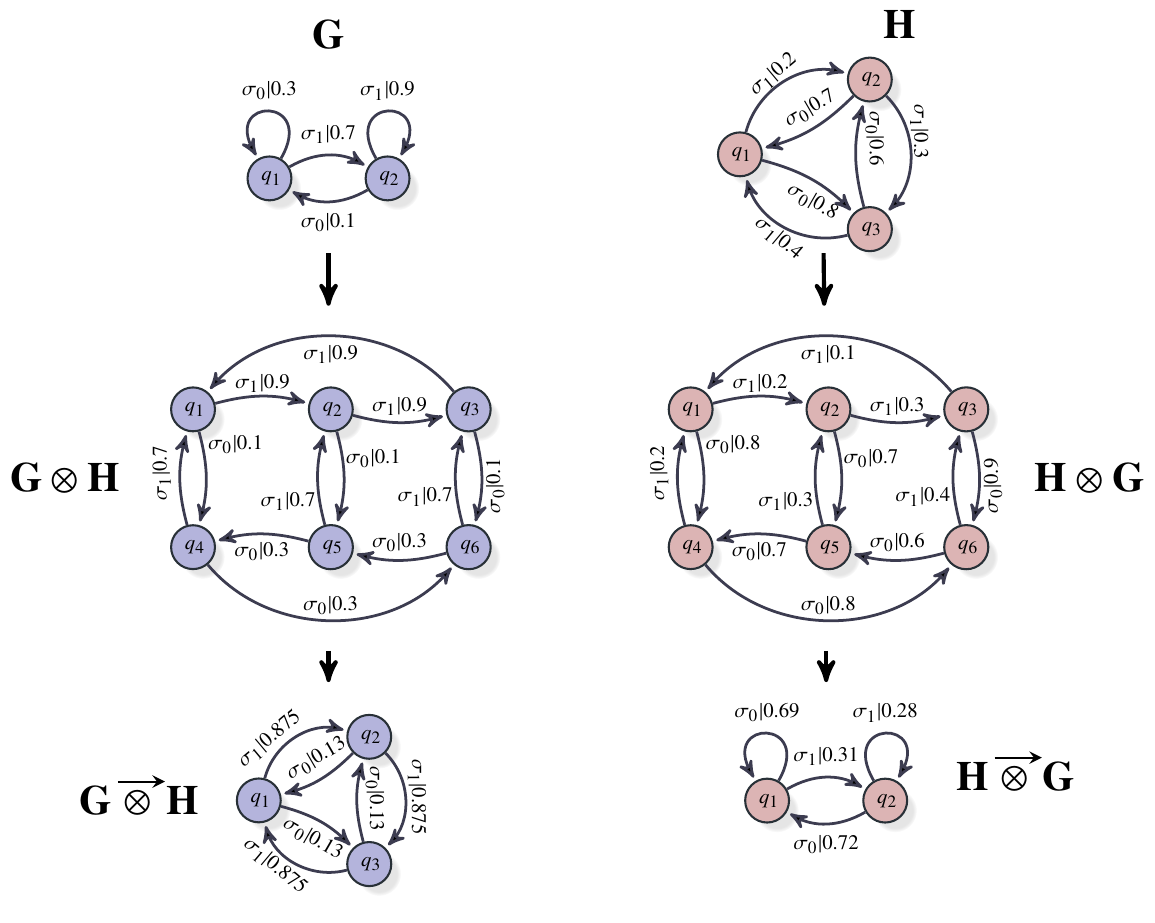}
\vspace{-10pt}

\captionN{\textbf{Illustration of synchronous and projective compositions.} 
Two PFSAs $G,H$ on the same alphabet can always be represented on the same, possibly larger, graph by constructing the synchronous compositions. Thus, $G \sync H$ and $H \sync G$ have the same graphs, but different probability specifications on the edges. There is no loss of information, since the PFSA obtained by synchronous composition is a non-minimal realization of the same underlying measure. We can force the PFSAs to a smaller structure using projective composition (as shown, we can compute the representation of $G$ in the graph of $H$ and vice verse), but this involves a loss of information. The projected distributions (See Definition~\ref{defprofdist}) still remain invariant.
 }\label{figCOMP}
\vspace{-10pt}

\end{figure*}
}

\begin{lem}[Transition From Stationary Distribution]\label{lemstat}
Let $G=(Q,\Sigma,\delta, \pitilde) $ be a PFSA representation for  a stationary ergodic QSP. If $G$ is distributed over its states as its stationary distribution $\wp_\lambda$, and the next symbol is generated according to the distribution $\wp_\lambda\Pitilde$, then the next expected state distribution remains unaltered. 
\end{lem}
\Proof{
Let $v = \wp_\lambda\Pitilde$. Then the next state distribution $\wp'$, may be computed using the $\vert Q \vert \times \vert Q \vert$ transformation matrices $\Gamma_\sigma, \sigma \in \Sigma$ (See Definition~\ref{defGamma}) as:
\cgather{
\wp' = \wp_\lambda \sum_{i=1}^{\vert \Sigma \vert} \Gamma_{\sigma_i} v_i = \wp_\lambda \sum_{j=1}^{\vert Q\vert} \wp_\lambda \big \vert_j \sum_{i=1}^{\vert \Sigma \vert} \Gamma_{\sigma_i} \Pitilde_{ji}
}
Since $\sum_\sigma \Gamma_\sigma = \Pi$, and $\forall j, \sum_i \Pitilde_{ji} = 1$, and $\sum_{j=1}^{\vert Q\vert} \wp_\lambda \big \vert_j=1$,  we conclude:
\cgather{
\wp' =  \wp_\lambda \sum_{j=1}^{\vert Q\vert} \wp_\lambda \big \vert_j \Pi =  \wp_\lambda \Pi =  \wp_\lambda
}
which completes the proof.
}
\begin{lem}[Condition For Trivial XPFSA To Imply Independence]\label{lembirectInd}
For  stationary ergodic QSPs $\mathcal{H_A},\mathcal{H_B}$ over  finite alphabets $\Sigma_A,\Sigma_B$ respectively, if we have:
\cgather{
\left ( \forall x, y \in \Sigma^\star_A, x \sim_\mathcal{H_B}^\mathcal{H_A} y \right )  \bigwedge \left (\forall x', y' \in \Sigma^\star_B, x' \sim_\mathcal{H_A}^\mathcal{H_B} y' \right ) \label{eq59}
}
then $\mathcal{H_A}$ and $\mathcal{H_B}$ are independent.
\end{lem}
\Proof{
To assign explicit labels to the sequences of random variables in the processes under consideration, let $\mathcal{H}_A= \{ W^A_k\}, \mathcal{H}_B= \{ W^B_k\}$, $k \in \mathbb{N}$.
For some arbitrary fixed $s,t \in \mathbb{N}$, satisfying $s \leqq t$, let  $\mathcal{H}_A^s= \{ W^A_{k+s}\}, \mathcal{H}_B^t= \{ W^B_{k+t}\}, k \in \mathbb{N}$, $i.e.$, $\mathcal{H}_A^s,\mathcal{H}_B^t$ are right-translated variants of the processes $\mathcal{H}_A,\mathcal{H}_B$ respectively. We assume without loss of generality that the initial state for the canonical representations of the processes $\mathcal{H}_A,\mathcal{H}_B$ are $\wp^A_\lambda, \wp^B_\lambda$ (stationary distributions over the respective causal states, see Definition~\ref{defcanon}) respectively.

We claim that:
\cgather{
\forall x \in \Sigma^s_A, \phi^{\mathcal{H_A}^s,\mathcal{H_B}^t}_x  = \wp^B_\lambda\Pitilde_B \tag{Claim A}
}

To establish this claim, we recall that the definition of  crossed-derivative from one process to another  has no reference to the transpired corresponding string in the second process. In other words, we marginalize out the transpired string in the second process. Since $\mathcal{H}_B$ is assumed to be initialized at the stationary distribution $\wp^B_\lambda$, we conclude that marginalizing over all strings in  $\mathcal{H}_B$ that can transpire for some $x \in \Sigma^s_A$, that the expected state is still $\wp^B_\lambda$ at $k=s$. Since,  the first conjunctive term in Eq.~\eqref{eq59} implies that:
\cgather{
\forall x \in \Sigma^s_A, \phi^{\mathcal{H_A},\mathcal{H_B}}_x  = \wp^B_\lambda\Pitilde_B
}
it follows that the expected state for $\mathcal{H}_B$ at $k=s+1$ is still $\wp^B_\lambda$ (Using Lemma~\ref{lemstat}). Continuing to marginalize of all sequences that may transpire between $k=s+1$ and $k=t$, we conclude that the state at $k=t$ is still $\wp^B_\lambda$, and hence the next symbol will be distributed as $\wp^B_\lambda\Pitilde_B$.
This establishes Claim A.

Next we claim that:
\cgather{
\forall x \in \Sigma^t_B, \phi^{\mathcal{H_B}^t,\mathcal{H_A}^s}_x  = \wp^A_\lambda\Pitilde_A \tag{Claim B}
}
which follows immediately by noting that if $x=x_1 \cdots x_s \cdots x_t=yx_{s+1}\cdots x_t$, then 
it follows from the second conjunctive term in Eq.~\eqref{eq59}:
\cgather{
\forall y=x_1\cdots x_s \in \Sigma^s_B, \phi^{\mathcal{H_B}^s,\mathcal{H_A}^s}_y  = \wp^A_\lambda\Pitilde_A
}
and that the future symbols $x_{s+1} \cdots x_t$ do not affect the next-symbol distribution at $k=s$.

Also, note that as before, when we are computing $\phi^{\mathcal{H_B}^t,\mathcal{H_A}^s}_x$, we can marginalize over all strings of length $s$ that occur in $\mathcal{H_A}$, implying that the expected state at $k=s$ is $\wp^A_\lambda$.

Using claims A and B, and the fact that the expected states at $k=s$ and $k=t$ for the 
 the processes $\mathcal{H_A}^s, \mathcal{H_B}^t$ are $\wp^B_\lambda$ and $\wp^A_\lambda$ respectively, we conclude that:
\mltlne{
\forall \sigma_i \in \Sigma_A, \sigma_j \in \Sigma_b,\\  Pr(W^A_{s+1}=\sigma_i,W^B_{t+1}=\sigma_j) = \wp^A_\lambda \Pitilde^A \big \vert_i \wp^B_\lambda \Pitilde^B \big \vert_j = \\Pr(W^A_{s+1}=\sigma_i) Pr(W^B_{t+1}=\sigma_j)
}
which establishes the following:
\cgather{
\forall s,t \in \mathbb{N}, W^A_{s}, W^B_{t} \textrm{ are pairwise independent} \tag{Claim C}
}
Finally, we use induction to complete the proof. We  consider a sequence $k_1, \cdots, k_m$ with $\forall i, k_i \in \mathbb{N}$. For our induction basis, we note that claim C implies that $W^A_{t_1}, W^B_{t_1} $  are pairwise independent.
As our induction hypothesis, let the random vectors $W^A_{t_1} \cdots W^A_{t_{m-1}}$ and $W^B_{t_1} \cdots W^B_{t_{m-1}}$ be independent. To conclude the proof, we argue:
\calign{
&Pr\left (W^A_{t_1} \cdots W^A_{t_{m}}, W^B_{t_1} \cdots W^B_{t_{m}} \right )  = \notag \\  &Pr\left (W^A_{t_1} \cdots W^A_{t_{m-1}}, W^B_{t_1} \cdots W^B_{t_{m-1}}\right ) \notag \\ &\mspace{40mu}\times  Pr\left (W^A_{t_{m}}, W^B_{t_{m}} \bigg \vert W^A_{t_1} \cdots W^A_{t_{m-1}}, W^B_{t_1} \cdots W^B_{t_{m-1}}\right )  \\
= & Pr\left (W^A_{t_1} \cdots W^A_{t_{m-1}}\right )   Pr \left ( W^B_{t_1} \cdots W^B_{t_{m-1}}\right ) \notag \\ &\mspace{40mu}\times Pr\left (W^A_{t_{m}}, W^B_{t_{m}} \bigg \vert W^A_{t_1} \cdots W^A_{t_{m-1}}, W^B_{t_1} \cdots W^B_{t_{m-1}}\right )  \\
= & Pr\left (W^A_{t_1} \cdots W^A_{t_{m-1}}\right ) Pr \left ( W^B_{t_1} \cdots W^B_{t_{m-1}}\right ) \notag \\ &\mspace{40mu}\times Pr\left (W^A_{t_{m}} \bigg \vert W^A_{t_1} \cdots W^A_{t_{m-1}} \right ) Pr \left ( W^B_{t_{m}} \bigg \vert W^B_{t_1} \cdots W^B_{t_{m-1}}  \right ) \notag  \\
= &Pr\left (W^A_{t_1} \cdots W^A_{t_{m}}\right ) Pr \left ( W^B_{t_1} \cdots W^B_{t_{m}}\right )
}
This completes the proof.
}
\begin{thm}[Directional (In)Dependence]\label{thmdirectind}
For stationary ergodic QSPs to be  independent, it is  necessary and sufficient for  the minimal XPFSAs in both directions to have a single state.
\end{thm}
\Proof{
Follows immediately from Lemmas~\ref{lemInd2XPFSA}, \ref{lemunidep}, and \ref{lembirectInd}.
}

Theorem~\ref{thmdirectind}, and the lemmas that build up to it, establish that XPFSAs capture a notion of directional dependence, and are well-suited to determine the directional causality flow between different processes. While the XPFSAs are generative models, it is useful to have a scalar quantification of the degree of directional dependence.

\subsubsection{Degree Of Directional Dependence}
The binary operation of synchronous composition on the space of probabilistic automata was introduced in the first author's earlier work\cite{CR08} for initial marked PFSA. We modify the definition to apply to PFSA representations for ergodic stationary QSPs, where the initial state is unimportant.

\begin{defn}[Synchronous Composition Of Probabilistic Automata]\label{defsynchro}
Let $G=(Q,\Sigma,\delta,\pitilde)$ be a PFSA representation of a stationary ergodic QSP. Let $H=(Q',\Sigma,\delta')$ represent a strongly connected directed  graph such that  $Q'$ is the set of nodes, and  $\forall q_i',q_j' \in Q'$, there is a directed edge $q_i' \xrightarrow{\sigma} q_j'$, labeled with $\sigma \in \Sigma$, if and only if $\delta'(q_i',\sigma) = q_j'$. Let $G^\otimes = (Q\times Q',\Sigma,\delta^\otimes,\pitilde^\otimes)$ be a PFSA where the relevant functions are defined as:
\cgather{
\forall q_i \in Q, q_j \in Q', \sigma \in \Sigma, \left \{ \begin{array}{l}\delta^\otimes((q_i,q_j'), \sigma) = (\delta(q_i, \sigma),\delta'(q_j',\sigma)) \\
\pitilde^\otimes((q_i,q_j'), \sigma) = \pitilde(q_i,\sigma)\end{array}\right.
}
Then the synchronous composition of $G,H$, denoted as $G \sync H$, is any strongly connected component of $G^\otimes$.
\end{defn}

We show that Definition~\ref{defsynchro} is consistent, in the sense that any two strongly connected components of $G^\otimes$ is structurally isomorphic.

\begin{lem}[Well-definedness Of Synchronous Composition]\label{lemsyncho}
Let $G^\otimes_1, G^\otimes_2$ be two strongly connected components of $G^\otimes$, as introduced in the course of Definition~\ref{defsynchro}. Then, we have:
\begin{enumerate}
\item $G^\otimes_1$ and $G^\otimes_2$ are structurally isomorphic.
\item $G^\otimes_i, i=1,2$ encodes a refinement of the probabilistic Nerode relation encoded by $G$.
\end{enumerate}
\end{lem}
\Proof{ 
To establish statement 2), we must show that $G^\otimes$ is a non-minimal realization of the Nerode relation encoded by $G$, for any choice of initial state. To see this, choose a state $q_0 \in Q$, and augment $G$ to an initial-marked PFSA $(Q,\Sigma,\delta,\pitilde,q_0)$. Also, choose a state $(q_0,q_0') \in Q \times Q'$, and augment $G^\otimes$ as $((Q\times Q',\Sigma,\delta^\otimes,\pitilde^\otimes, (q_0,q_0'))$.
Then, it is immediate that:
\mltlne{
\forall x,y \in  \Sigma^\star \bigg ( \delta((q_0,q_0'),x) = \delta((q_0,q_0'),y) \\ \Rightarrow 
 \delta(q_0,x) = \delta(q_0,y)
\bigg )
}
This establishes statement 2).

Next, consider the graph for the PFSA $G^\otimes$, and augment it with a new morph function $\pitilde''$, so as to get a PFSA $G''=(Q\times Q',\Sigma,\delta^\otimes,\pitilde'')$, such that each row of $\Pitilde''$ is distinct. It follows that no state in $G''$ may be merged with another, since for any two states $q_1,q_2 \in Q\times Q'$, and any $\sigma \in \Sigma$, we have by construction: $\pitilde''(q_1,\sigma) \neq \pitilde''(q_2,\sigma)$.

Since, $H$ is strongly connected, $G''$ represents some specific probabilistic Nerode equivalence on $\Sigma^\star$  (Note in contrast, if $H$ had multiple components, then the choice of the initial state might be important). Let the Nerode relation, corresponding to the stationary ergodic QSP encoded by $G''$, be denoted as $\sim_{G''}$.

Now, consider two strongly connected components $G_1,G_2$ for $G''$, with state sets $Q_1 \subseteqq Q\times Q', Q_2 \subseteqq Q\times Q'$. Since Theorem~\ref{thmminrez} establishes that strong components of PFSA are realizations of the same Nerode relation encoded by the full model, we conclude that $G_1$ and $G_2$ are both encodings of $\sim_{G''}$, $i.e.$ there exists a map $\mathds{H}_1:Q_1 \to \mathscr{E},\mathds{H}_2:Q_2 \to \mathscr{E}$, where $\mathscr{E}$ is the set of equivalence classes of $\sim_{G''}$. We note that it is immediate that $\mathds{H}_1,\mathds{H}_2$ are surjective. From definition of the Nerode equivalence, it also follows that no two states can map to the same equivalence class (to avoid merging), implying that $\mathds{H}_1,\mathds{H}_2$ are injective as well, implying that the inverse maps $\mathds{H}_1^{-1}\colon  \mathscr{E} \to Q_1,\mathds{H}_2^{-1}\colon  \mathscr{E} \to Q_2$ are well-defined. Now, we construct maps $\xi:Q_1 \to Q_2, \xi':Q_2 \to Q_1$ as follows:
\cgather{
\forall q \in Q_1, \xi(q) = \mathds{H}_2^{-1}\mathds{H}_1(q) \\
\forall q \in Q_2, \xi'(q) = \mathds{H}_1^{-1}\mathds{H}_2(q) 
}
It follows that:
\cgather{
\forall q \in Q_1, \xi'(\xi(q)) = \mathds{H}_1^{-1}\mathds{H}_2\mathds{H}_2^{-1}\mathds{H}_1(q) = q
}
implying  $\xi$ is bijective. 
Since $G_1,G_2$ are components of $G''$, we note:
\mltlne{
\forall \sigma \in \Sigma, q \in Q_1\subseteqq Q \times Q',  \\ \delta^\otimes(q, \sigma) = \mathds{H}_1^{-1}  ([x\sigma]), \forall x \in \mathds{H}_1(q)\\
\Rightarrow \xi(\delta^\otimes(q, \sigma)) = \mathds{H}_2^{-1}\mathds{H}_1  \mathds{H}_1^{-1}  ([x\sigma])  \\  =  \mathds{H}_2^{-1}([x\sigma]) = \delta^\otimes(\xi(q),\sigma)
}
Similarly, assuming the probability measure on $\Sigma^\omega$ encoded by $G''$ is $\mu$, we also have: 
\mltlne{
\forall \sigma \in \Sigma, q \in Q_1\subseteqq Q \times Q', \\ \pitilde^\otimes(q,\sigma) = \frac{\mu(x\sigma\Sigma^\omega)}{\mu(x\Sigma^\omega)}, \forall x \in \mathds{H}_1(q)
}
and, also:
\mltlne{\forall \sigma \in \Sigma, q \in Q_1\subseteqq Q \times Q',\\ \pitilde^\otimes(\xi(q),\sigma) = \frac{\mu(x'\sigma\Sigma^\omega)}{\mu(x'\Sigma^\omega)}, \forall x' \in \mathds{H}_2(\xi(q)) = \mathds{H}_1(q)
}
which implies:
\cgather{
\forall \sigma \in \Sigma, q \in Q_1\subseteqq Q \times Q',\pitilde^\otimes(q,\sigma) =\pitilde^\otimes(\xi(q),\sigma)
}
Hence, $G_1, G_2$ are structurally isomorphic.
Noting that the morph $\pitilde''$ is arbitrary completes the proof.
}

Next we introduce projective composition. Again, a slightly different version was introduced in  the authors' earlier work\cite{CR08}. 
\begin{defn}[Projective Composition]\label{defproj}
For a given PFSA $G=(Q,\Sigma,\delta,\pitilde)$, and a strongly connected directed  graph $H=(Q',\Sigma,\delta')$, such that  $Q'$ is the set of nodes, and  $\forall q_i',q_j' \in Q'$, there is a directed edge $q_i' \xrightarrow{\sigma} q_j'$, labeled with $\sigma \in \Sigma$, if and only if $\delta'(q_i',\sigma) = q_j'$, the projective composition $G \psync H = (Q',\Sigma,\delta',\pitilde')$ is a PFSA with:
\mltlne{
\forall q' \in Q',\sigma \in \Sigma, \\  \pitilde'(q',\sigma) = \left \{ \begin{array}{l}\frac{\displaystyle\sum_{(q,q')  \in Q''}  \pitilde''\big ((q,q'),\sigma \big )\wp_\lambda'' \big \vert_{(q,q')}}{\displaystyle\sum_{(q,q')  \in Q''}^{\phantom{.}}  \wp_\lambda'' \big \vert_{(q,q')}  }  \\ \\  \textrm{\hspace{30pt},if } \sum_{(q,q')  \in Q''}  \wp_\lambda'' \big \vert_{(q,q')} > 0 \\   0  \\  \textrm{\hspace{30pt},otherwise} \end{array}\right. 
}
where $G \sync H = (Q'',\Sigma,\delta'',\pitilde'')$ ,  and $  \wp_\lambda''$ is the corresponding  stationary distribution on $ Q'' $.
\end{defn}

Note that  synchronous  and projective compositions are defined to operate on a pair of arguments, the first of which is a PFSA and the second is strongly connected graph with edges labeled by symbols from  the same alphabet. However, we can extend them as binary operators on the space of strongly connected PFSA on a fixed alphabet, using the graph of the second PFSA as the second argument for the operators. Thus, it makes sense to talk about $G \sync G, G \sync H, G\psync H$ where $G,H$ are PFSA with strongly connected graphs. Indeed, one can show easily that, for any such PFSA $G,H$:
\cgather{
G \sync G = G \\
G \psync G = G \\
(G \psync H ) \psync H = G \psync H
}
Additionally, the projective composition preserves the projected distribution.
\begin{defn}[Projected Distribution]\label{defprofdist}
Given a PFSA $G$ encoding the probability space $(\Sigma^\omega,\mathfrak{B},\mu_G)$, and a  PFSA $H=(Q^H,\Sigma,\delta^H,\pitilde^H)$, the projected distribution $\base{G}_H$ of   $G$ with respect to   $H$  is a vector $\wp \in [0,1]^{\vert Q^H\vert}$, such that:
\cgather {
\forall j \in \{1,\cdots, \vert Q^H\vert \}, \wp_j =  \sum_{x \in \mathcal{E}_j} \mu_G(x\Sigma^\omega)
}
for any choice of initial state in $G$, and 
where $\mathcal{E}_j$ is the equivalence class for the transition equivalence (See Lemma~\ref{lemtranseq})  corresponding to state $q_j \in Q^H$, again for any choice of initial state in $H$.

We note $\base{G}_H$ is always a  ``probability vector'', i.e.,
\cgather{\forall j,  \base{G}_H \bigg \vert_j \geqq 0 
\textrm{ and }\sum_{j = 1}^{\vert Q^H \vert} \base{G}_H \bigg \vert_j =
\sum_{x \in \Sigma^\star}\mu_G(x\Sigma^\omega)   = \mu_G(\Sigma^\omega) = 1
}
\end{defn}
%
\begin{lem}[Projected Distribution Well-defined-ness \&  Invariance]\label{lemprojdis}
 For PFSA $P$ and $G$ encoding stationary ergodic QSPs over the same alphabet:
\begin{enumerate}
\item $\base{G}_H$ is independent of the choice of the initial states in Definition~\ref{defprofdist}
\item $\base{G}_H$ is the stationary distribution on the states of the projective composition of $G$ with $H$, $i.e.$, we have: 
\cgather{
\base{G}_H = \base{G \psync H}_H
}
\end{enumerate}
\end{lem}
\Proof{
We note that  statement 2) implies statement 1) from ergodicity.
To establish statement 2), we argue as follows: 
Let $G=(Q,\Sigma,\delta,\pitilde), H = (Q',\Sigma,\delta',\pitilde')$, and also let $G \psync H = (Q', \Sigma, \delta',\pitilde'')$. 

Let $\base{G}_H = \wp^\star$. Additionally, let the stationary distribution on the states of 
$G \sync H$ be denoted as $\wp^\otimes$. 

We claim $\wp^\star$ is also a stationary distribution for $G \psync H$.

Denoting the measure encoded by $G$ as $\mu_G$, and the equivalence class for the transitional equivalence corresponding to state $q_j \in Q'$ as $\mathcal{E}(q_j)$, we note that:
\cgather{
\wp^\star_j = \sum_{x \in \mathcal{E}(q_j), q_j \in Q'} \mu_G(x\Sigma^\omega) = \sum_{q \in Q} \wp^\otimes_{(q,q_j)}
}
where we have used the fact that states in $G \sync H$ are of the form $(q,q')$, with $q \in Q, q' \in Q'$. The transition probability matrix $\Pi''$ for $G \psync H$ (each entry being the probability of transitioning from one state to another via possibly different symbols, in a single step) is defined as:
\cgather{
\Pi''_{ij} = \sum_{\mathclap{\sigma \in \Sigma : \delta'(q_i,\sigma) = q_j}} \pitilde'(q_i, \sigma)
}
and we set (assuming $\wp^\star$ is a row-vector):
\cgather{
v = \wp^\star \Pi''
}
implying that we have:
\mltlne{\forall q_k \in Q',
v_k = \sum_{j=1}^{\vert Q' \vert} \wp^\star_j \Pi''_{jk} =  \sum_{j=1}^{\vert Q' \vert} \sum_{{\phantom{x}\sigma: q_j \xrightarrow[\sigma]{} q_k}} \wp^\star_j  \pitilde''(q_j,\sigma) \\ =  \sum_{j=1}^{\vert Q' \vert} \sum_{{\phantom{x}\sigma: q_j \xrightarrow[\sigma]{} q_k}} \sum_{q \in Q} \wp^\otimes_{(q,q_j)} \pitilde^\otimes((q,q_j),\sigma) \\
=  \sum_{q \in Q}\sum_{j=1}^{\vert Q' \vert}  \wp^\otimes_{(q,q_j)}\sum_{{\phantom{x}\sigma: q_j \xrightarrow[\sigma]{} q_k}} \pitilde^\otimes((q,q_j),\sigma) \\ = \sum_{q \in Q}\sum_{j=1}^{\vert Q' \vert}  \wp^\otimes_{(q,q_j)} \Pi^\otimes_{(q,q_j),(q,q_k)}
}
Since, $\wp^\otimes$ is a stationary distribution for $G \sync H$, it follows:
\cgather{\forall q_k \in Q',
v_k = \sum_{q \in Q} \wp^\otimes_{(q,q_k)} = \wp^\star_k
}
which establishes that $ v = \wp^\star$, $i.e.$, $\wp^\star$ is a stationary distribution for $G \psync H$. By ergodicity, it follows that stationary distributions for both $G$, and $G \psync H $ are unique, which completes  the proof.
}
\DETAILS{
\begin{figure*}
\centering

\includegraphics[width=6in]{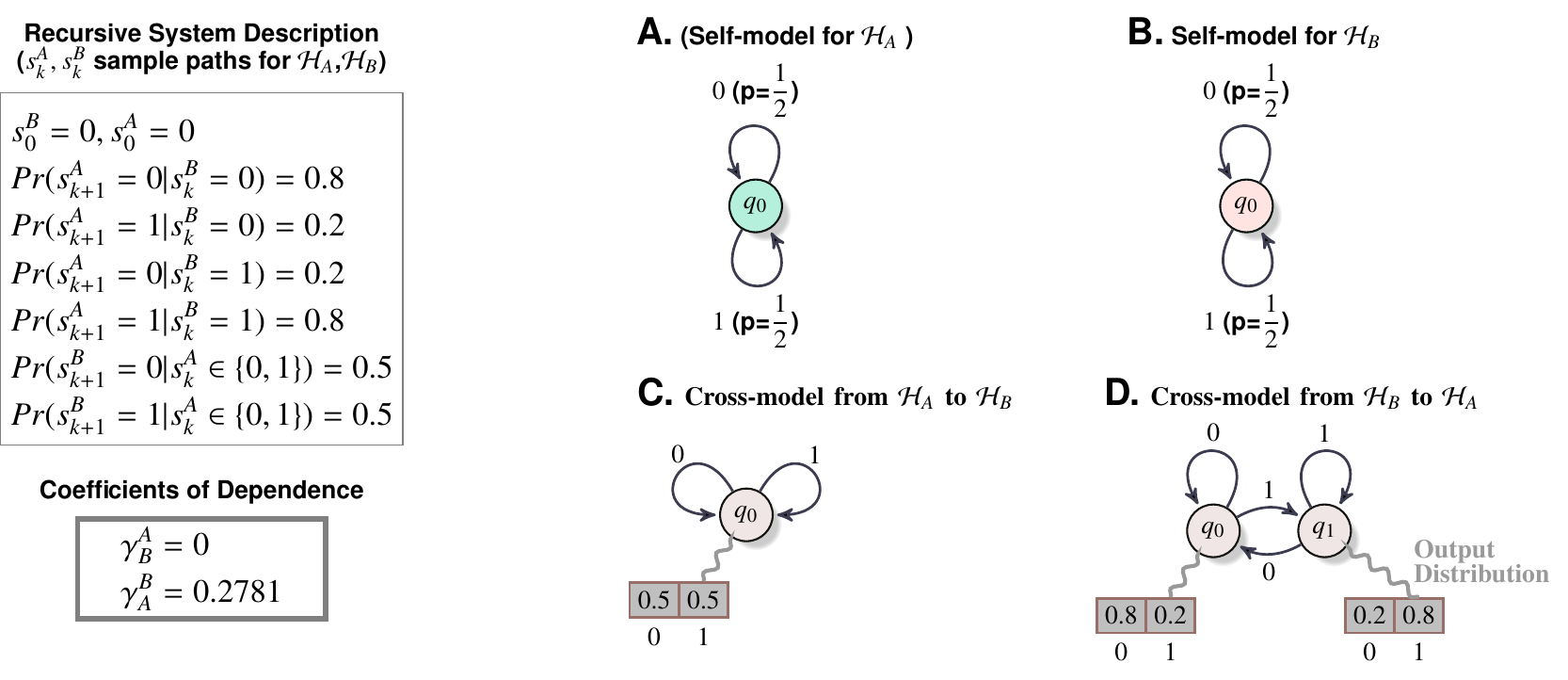}
\vspace{-10pt}

\captionN{\textbf{Example Processes With Uni-directional Dependence.} For the  system description tabulated above, we get the two self-models (plates A and B) which are single state PFSAs. It also follows that process $\mathcal{H}_A$ cannot predict any symbol in process $\mathcal{H}_B$, and we get the XPFSA from $A$ to $B$ as a single state machine as well (plate C). However, process $\mathcal{H}_A$ is somewhat predictable by looking at $\mathcal{H}_B$, and we have the XPFSA with two states in this direction (plate D). Note the tabulated coefficients of dependence in the two directions. Since for this example, $\mathcal{H}_A$ is the Bernoulli-$1/2$ process with entropy rate of $1$ {\sffamily  bit/letter}, it follows that making observations in the process $\mathcal{H}_B$ reduces the entropy of the next-symbol distribution in $\mathcal{H}_A$ by $0.2781$ {\sffamily bits}.  
 }\label{figEX1}
\vspace{-10pt}

\end{figure*}
}
We are now ready to define the coefficient of causal dependence. We recall from Definition~\ref{defcrossderiv}, that given two stationary ergodic QSPs $\mathcal{H}_A,\mathcal{H}_B$ over  finite alphabets $\Sigma_A,\Sigma_B$, the cross-derivative 
$\phi^{\mathcal{H}_A,\mathcal{H}_B}_x$ at  $x \in \Sigma^\star_A$ specifies the next-symbol distribution in  $\mathcal{H}_B$, given the knowledge that the string transpired in  $\mathcal{H}_A$ is $x$.
\begin{defn}[Coefficient Of Causal Dependence]\label{defcoeffcd}
Let $\mathcal{H}_A,\mathcal{H}_B$ be  stationary ergodic QSPs over  finite alphabets $\Sigma_A,\Sigma_B$ respectively. The coefficient of causal dependence of $\mathcal{H}_B$ on $\mathcal{H}_A$, denoted as $\gamma^A_B$, is defined as the ratio of the  expected change in entropy of the next symbol distribution in $\mathcal{H}_B$ due to observations in $\mathcal{H}_A$ to the entropy of the  next symbol distribution in $\mathcal{H}_B$ in the absence of observations in $\mathcal{H}_A$, $i.e.$, we have:
\cgather{
\gamma^A_B = 1 - \frac{\mathop{\mathbf{E}}_{x \in \Sigma^\star_A} \left (  \entropy{\phi^{\mathcal{H}_A,\mathcal{H}_B}_x} \right ) }{\entropy{\phi^{\mathcal{H}_A,\mathcal{H}_B}_\lambda }}
}
where the entropy $\entropy{u}$ of a discrete  probability distribution $u$ is given by $ \sum_i u_i \log_2 u_i$. We assume that $\mathcal{H}_B$ is not a trivial process, producing only a single alphabet symbol, thus precluding the possibility that the denominator is zero.
\end{defn}
\begin{lem}[XPFSA to Coefficient of Causal Dependence]\label{lemcoeffcd1}
For stationary ergodic QSPs $\mathcal{H}_A,\mathcal{H}_B$    over   alphabets $\Sigma_A,\Sigma_B$,    let the PFSAs encoding the processes be $A=(Q_A,\Sigma_A,\delta_A,\pitilde_A)$ and $B=(Q_B,\Sigma_B,\delta_B,\pitilde_B)$ respectively. Also, let the XPFSA from $\mathcal{H}_A$ to $\mathcal{H}_B$ be
$B^A = (Q^A_B, \Sigma_B, \delta^A_B, \pitilde^A_B)$. Then, if $Q^A_B = \{q_1, \cdots, q_m\}$, then we have:
\cgather{
\gamma^A_B = 1 - \frac{ \left \langle \base{A \psync B^A }_{B^A}, \begin{pmatrix}  \entropy{\pitilde^A_B(q_1,\cdot)} \\ \vdots \\  \entropy{\pitilde^A_B(q_m,\cdot)} \end{pmatrix} \ \right  \rangle  }{ \entropy{\wp_\lambda^B \pitilde_B}}
} 
where $\langle \cdot, \cdot \rangle$ is the standard inner product, and $\wp^B_\lambda$ is the stationary distribution on the states of $B$.
\end{lem}
\Proof{
The denominator follows from Corollary~\ref{corExpXd} to Lemma~\ref{lemExpXd}. 

Let  $A$ encode the probability space $(\Sigma_A^\omega, \mathfrak{B},\mu)$. Then, we have:
\cgather{
\mathop{\mathbf{E}}_{x \in \Sigma^\star_A} \left (  \entropy{\phi^{\mathcal{H}_A,\mathcal{H}_B}_x} \right ) = 
\sum_{x \in \Sigma^\star_A} \mu(x\Sigma_A^\omega) \entropy{\phi^{\mathcal{H}_A,\mathcal{H}_B}_x}
}
Noting that  the equivalence classes of the cross-Nerode equivalence $\sim^{\mathcal{H}_A}_{\mathcal{H}_B}$ correspond to elements in the set $Q^A_B$, we denote the equivalence class of strings in $\Sigma^\star_A$ corresponding to state $q \in Q^A_B$ as  $\mathscr{E}(q)$.
Then:
\cgather{
\sum_{x \in \Sigma^\star_A} \mu(x\Sigma_A^\omega) \entropy{\phi^{\mathcal{H}_A,\mathcal{H}_B}_x} = \sum_{q \in Q^A_B} \sum_{x \in \mathscr{E}(q)} \mu(x\Sigma_A^\omega) \entropy{\phi^{\mathcal{H}_A,\mathcal{H}_B}_x}
}
We note that:
\cgather{
\forall x  \in \mathscr{E}(q), \phi^{\mathcal{H}_A,\mathcal{H}_B}_x = \pitilde^A_B(q,\cdot)
}
and hence we have, using Definition~\ref{defprofdist}, and Lemma~\ref{lemprojdis}:
\calign{
&\sum_{q \in Q^A_B} \sum_{x \in \mathscr{E}(q)} \mu(x\Sigma_A^\omega) \entropy{\phi^{\mathcal{H}_A,\mathcal{H}_B}_x} \notag \\= &\sum_{q \in Q^A_B} \entropy{ \pitilde^A_B(q,\cdot) } \left ( \sum_{x \in \mathscr{E}(q)} \mu(x\Sigma_A^\omega)  \right )
 = \sum_{q \in Q^A_B} \entropy{ \pitilde^A_B(q,\cdot) } \base{A}_{B^A} \bigg \vert_q  \\= & \sum_{q \in Q^A_B} \entropy{ \pitilde^A_B(q,\cdot) } \base{A \psync B^A}_{B^A} \bigg \vert_q }
which completes the proof.
}
\begin{thm}[Properties of Coefficient Of Causal Dependence]\label{thmcoeffcd2}
For stationary ergodic QSPs $\mathcal{H}_A,\mathcal{H}_B$    over   alphabets $\Sigma_A,\Sigma_B$,  we have:
\begin{enumerate}
\item $ \gamma^A_B \in [0 ,1]$
\item $\mathcal{H}_A$ and $\mathcal{H}_B$ are independent if and only if $ \gamma^A_B =  \gamma^B_A =0$.
\end{enumerate}
\end{thm}
\Proof{
We note that non-negativity of entropy implies:
\cgather{
\frac{\mathop{\mathbf{E}}_{x \in \Sigma^\star_A} \left (  \entropy{\phi^{\mathcal{H}_A,\mathcal{H}_B}_x} \right ) }{\entropy{\phi^{\mathcal{H}_A,\mathcal{H}_B}_\lambda }} \geqq 0 \Rightarrow \gamma^A_B \geqq 0
}
For the upper bound, we note that by marginalizing out $x$ from $\phi^{\mathcal{H_A},\mathcal{H_B}}_x$, we get:
\cgather{
\phi^{\mathcal{H}_A,\mathcal{H}_B}_\lambda = \mathop{\mathbf{E}}_{x \in \Sigma^\star_A} \phi^{\mathcal{H}_A,\mathcal{H}_B}_x  \\ \Rightarrow
\frac{\mathop{\mathbf{E}}_{x \in \Sigma^\star_A} \left (  \entropy{\phi^{\mathcal{H}_A,\mathcal{H}_B}_x} \right ) }{\entropy{\phi^{\mathcal{H}_A,\mathcal{H}_B}_\lambda }} = \frac{\mathop{\mathbf{E}}_{x \in \Sigma^\star_A} \left (  \entropy{\phi^{\mathcal{H}_A,\mathcal{H}_B}_x} \right ) }{\entropy{    \mathop{\mathbf{E}}_{x \in \Sigma^\star_A} \phi^{\mathcal{H}_A,\mathcal{H}_B}_x }}
}
Since entropy is  concave, Jensen's inequality~\cite{cover} guarantees:
\cgather{
\mathop{\mathbf{E}}_{x \in \Sigma^\star_A} \left (  \entropy{\phi^{\mathcal{H}_A,\mathcal{H}_B}_x} \right ) \leqq   \entropy{    \mathop{\mathbf{E}}_{x \in \Sigma^\star_A} \phi^{\mathcal{H}_A,\mathcal{H}_B}_x } \Rightarrow \gamma^A_B \leqq 1
}
This establishes statement 1).

Next we note that:
\cgather{
\gamma^A_B =0 \Rightarrow \mathop{\mathbf{E}}_{x \in \Sigma^\star_A} \left (  \entropy{\phi^{\mathcal{H}_A,\mathcal{H}_B}_x} \right ) =  \entropy{    \mathop{\mathbf{E}}_{x \in \Sigma^\star_A} \phi^{\mathcal{H}_A,\mathcal{H}_B}_x } \\
\Rightarrow \forall x \in \Sigma^\star_A, \phi^{\mathcal{H}_A,\mathcal{H}_B}_x = v , \textrm{ where $v$ is independent of $x$} \\
\Rightarrow \forall x,y \in \Sigma^\star_A, x \sim^A_B y
}
which implies that  $B^A$ has a single state in its minimal realization. It then  follows from Theorem~\ref{thmdirectind}, that:
\cgather{
 \gamma^A_B =  \gamma^B_A =0 \Rightarrow \textrm{ $\mathcal{H}_A,\mathcal{H}_B$ are independent  }
}
To establish the converse, we simply note that if  $\mathcal{H}_A,\mathcal{H}_B$ are independent  then both $A^B,B^A$ have single state minimal realizations (Theorem~\ref{thmdirectind}), which implies that 
\mltlne{
\forall x,y \in \Sigma^\star_A,  x \sim^A_B y \Rightarrow \forall x \in \Sigma^\star_A, \phi^{\mathcal{H}_B,\mathcal{H}_B}_x = \phi^{\mathcal{H}_A,\mathcal{H}_B}_\lambda \Rightarrow \gamma^A_B = 0 }
And, also:
\mltlne{ \forall x,y \in \Sigma^\star_B,  x \sim^B_A y \Rightarrow \forall x \in \Sigma^\star_B, \phi^{\mathcal{H}_B,\mathcal{H}_A}_x = \phi^{\mathcal{H}_B,\mathcal{H}_A}_\lambda \Rightarrow \gamma^B_A = 0
}
This completes the proof.
}
\DETAILS{
We reuse the example constructed in Lemma~\ref{lemunidep} to illustrate the computation of the coefficients of dependence. The system is defined via a set of recursive specifications of the next symbol distribution (See Figure~\ref{figEX1}). We note that both self-models in this case are single state PFSAs, and in fact represent the Bernoulli-$1/2$ process. The XPFSA in one direction is also trivial, leading to a zero coefficient of dependence, while the coefficient in the other direction is positive illustrating a case of uni-directional dependence (See  Figure~\ref{figEX1}).
\section{Algorithm \algo: Self-model Inference}\label{sec5}
We  construct an effective procedure to infer $PFSA$ $\mathcal{P}_\mathcal{H}$ from a sufficiently long run from a QSP $\mathcal{H}$, and a pre-specified  $\epsilon > 0$.
This section (Section~\ref{sec5}) has largely appeared elsewhere~\cite{CL12g}, but is included for the sake of completeness.
\subsection{Implementation Steps}\label{sec5ips}
The inference algorithm for PFSA seeks  similar symbolic derivatives (similar in the sense that infinity norm of the difference is  within some pre-specified bound $\epsilon$), and ``merges''  string fragments at which the derivatives turn out to be similar, $i.e.$ define them to reach the same state in the underlying model. This is  more general to state splitting or state merging, since  both processes are going on simultaneously: when we  find a symbolic derivative that fails to match to any of the derivatives already encountered, we create a new state; while if we do find such a match, then we merge the two strings at which the derivatives are found to be similar. It is crucial that we first seek out an $\epsilon$-synchronizing string, and look at its right extensions to carry out the merge and split; which, due to the preceding theoretical development, ensures that we are  finding states of the underlying $\mathcal{P}_\mathcal{H}$ within $\epsilon$ error in the infinity norm. 

 We call our algorithm ``\underline{Gen}erator \underline{E}xtraction Using \underline{Se}lf-\underline{s}imilar \underline{S}emantics'', or \algo which for
 an observed sequence $s$,   consists of three  steps: 

\begin{enumerate}
\item { \textit{Identification of $\epsilon$-synchronizing string $x_0$:} }
Construct a derivative heap $\mathcal{D}^s(L)$ using the observed trace $s$ (Definition~\ref{defderivheap}), and set $L$ consisting of all strings up to a sufficiently large, but finite, depth. We suggest as initial choice of $L$ as $\log_{\vert \Sigma \vert} 1/\epsilon $. In $L$ is sufficiently large,  then the inferred  model structure will not change for larger values. We then identify a vertex of the convex hull for $\mathcal{D}_\infty$, via any standard algorithm for computing the hull~\cite{QHULL}.  
Choose $x_0$ as the string mapping to this vertex.
\item {\textit{Identification of the 
structure of $\mathcal{P}_\mathcal{H}$, $i.e.$,  transition function $\delta$:}} We   generate $\delta$ as follows: For each state $q$, we associate a string identifier $x^{id}_q \in x_0\Sigma^\star$, and a probability distribution $h_q$ on $\Sigma$ (which is an approximation of the  $\Pitilde$-row corresponding to state $q$). We extend the structure recursively:
\begin{enumerate} 
\item Initialize the set $Q$ as $Q=\{q_0\}$, and set $x^{id}_{q_0}=x_0$, $h_q = \phi^s(x_0)$. 
\item 
For each state $q \in Q$, compute for each symbol $\sigma \in \Sigma$, find symbolic derivative $\phi^s(x^{id}_q \sigma)$. If $\vert \vert \phi^s(x^{id}_q \sigma) - h_{q' \sigma}\vert \vert_\infty \leqq \epsilon$ for some $q' \in Q$, then define $\delta(q,\sigma) = q'$. If, on the other hand, no such $q'$ can be found in $Q$, then add a new state $q'$ to $Q$, and define  $x^{id}_{q'}=x^{id}_q\sigma$, $h_{q'} = \phi^s(x^{id}_q\sigma)$.

\end{enumerate}
The process terminates when every $q \in Q$ has  a target state, for each $\sigma \in \Sigma$. Then, if necessary, we ensure strong connectivity using~\cite{tar146}.
\item { \textit{Identification of  arc probabilities, $i.e.$,  function $\pitilde$:}} 
\begin{enumerate}
\item  Choose an arbitrary initial state $q\in Q$.
\item Run  sequence $s$ through the identified graph, as directed by $\delta$, $i.e.$, if current state is $q$, and the next symbol read from $s$ is $\sigma$, then move to $\delta(q,\sigma)$. Count arc traversals, $i.e$, generate  numbers $N^i_j$ where $q_i \xrightarrow[N^i_j]{\sigma_j} q_k$.
\item Generate $\Pitilde$ by row normalization, $i.e.$, $\Pitilde_{ij} = N^i_j/(\sum_{j}N^i_j)$
\end{enumerate}
\end{enumerate}
Reported  recursive structure  extension algorithms\cite{Gavalda06,CastroG08} lack the $\epsilon$-synchronization step, and  are restricted to inferring only synchronizable or short-memory models, or large approximations for long-memory ones. 
\subsection{Complexity Analysis \& PAC Learnability}
\algo has no upper bound on the number of states; which is  a function of the process complexity itself.

While  $h_q$ (in step 2) approximates $\Pitilde$ rows, we  find the arc probabilities via normalization of traversal count.  $h_q$ only uses sequences in $x_0\Sigma^\star$, while traversal counting uses the entire sequence $s$, and is  more accurate.


We assume that the $\Pitilde$ rows corresponding to distinct states are separated in the $\sup$ norm by at least $\epsilon$. A PFSA with distinct states may have identical rows corresponding to multiple states. However, not all rows can be identical, for then the states would collapse, and we would get a single state PFSA. The proposed algorithm can be easily modified to address this issue; if two states have identical corresponding $\Pitilde$ rows, then they can be disambiguated from the  multiplicity of outgoing transitions with identical labels; however we do not discuss this issue here.

Another issue is obtaining a strongly connected PFSA, which can be ensured if before Step 2, we extract a strong component from the structure inferred in Step 1. This can be carried out efficiently using Tarjan's algorithm~\cite{tar146}, which has  $O(\vert Q \vert + \vert \Sigma \vert)$ asymptotic space and time complexity. 
%
\begin{thm}[Time Complexity]\label{thmcomplexity}
Assuming $\vert s \vert > \vert \Sigma \vert$, the asymptotic time complexity of \algo is:
\cgather{
\mathscr{T}=O\left  (\frac{\vert s \vert \vert \Sigma\vert}{\epsilon} \right )
}
\end{thm}
\Proof{
Assuming  $\vert s \vert > \vert \Sigma \vert$, we note that
 \algo performs the following computations:
\begin{enumerate}
 \item[C1] Computation of a derivative heap by computing $\phi^s(x)$ for $O(1/\epsilon)$ strings (Corollary~\ref{corsynchrodepth}), each of which involves reading the input  $s$ and normalization to  distributions over $\Sigma$, thus contributing $O(1/\epsilon \times (\vert s \vert + \vert \Sigma\vert )) = O(1/\epsilon \times \vert s \vert ) $  .
 \item[C2] Finding a vertex of the convex hull of the   heap, which, at worst, involves inspecting $O(1/\epsilon)$ points (encoded by strings  generating the heap),  contributing $O(1/\epsilon \times \vert \Sigma \vert )$, where  each  inspection is done  in $O(\vert \Sigma \vert)$ time.
 \item[C3] Finding   $\delta$, involving computing  derivatives at string-identifiers (Step 2), thus contributing $O( \vert Q\vert \times \vert \Sigma \vert \times \vert s \vert )$. 
\item[C4] Identification of arc probabilities using traversal counts and normalization, done in time linear in the number of arcs, $i.e$ $O(\vert Q\vert \times \vert \Sigma\vert)$.
\end{enumerate}
Summing the contributions, we have:
\calign{
\mathscr{T} &= O( 1/\epsilon \times \vert s \vert + 1/\epsilon \times \vert \Sigma \vert + \vert Q\vert \times \vert s \vert \times \vert \Sigma \vert+  \vert Q\vert \times \vert \Sigma\vert ) \notag \\
& = O\bigg (\big ( 1/\epsilon + \vert Q\vert  \vert \Sigma \vert) \times    \vert s\vert \bigg )
}
Noting that $\vert Q \vert $ is bounded by the maximum number of symbolic derivatives that may be distinguished, and hence by $1 / \epsilon$, we conclude:
\cgather{
\mathscr{T}=O\left  (\frac{\vert s \vert \vert \Sigma\vert}{\epsilon} \right )
}
which completes the proof.
}

Finite probabilistic identification  is referred to as Probably Approximately Correct learning~\cite{valiant84,ang92b,k94b} (PAC-learning), which  accepts  a hypothesis that is not too different from the correct language with high probability. 
An identification method is said to identify a target language  $L_\star$ in the  Probably Approximately Correct (PAC) sense~\cite{valiant84,ang92b,k94b}, if it always halts and outputs $L$ such that:
\cgather{
\exists \epsilon, \delta > 0, P(d(L_\star, L) \leqq \epsilon) \geqq 1-\delta
}
where $d(\cdot,\cdot)$ is a metric on the space of target languages. A class  of languages is  efficiently PAC-learnable if there exists an algorithm that PAC-identifies every language in the class, and runs in time polynomial in $1/\epsilon$, $1/\delta$, length of sample input, and inferred model size.
We prove PAC-learnability of QSPs, by first establishing a metric on the space of probabilistic automata over  $\Sigma$.
\subsection{PAC Identifiability  Of QSPs}
 We first establish an appropriate  metric to establish PAC-learnability of $\algo$.
\begin{lem}[Metric For Probabilistic Automata]\label{lemmetric}
 For two strongly connected PFSAs $G_1,G_2$, let the symbolic derivative at $x\in \Sigma^\star$ be denoted as $\phi^s_{G_1}(x)$ and $\phi^s_{G_2}(x)$ respectively. Then, 
\cgather{
\Theta(G_1,G_2) = \sup_{x \in \Sigma^\star } \left \{ \lim_{\vert s_1\vert,\vert s_2\vert \rightarrow \infty}\big \vert \big \vert \phi^{s_1}_{G_1}(x) - \phi^{s_2}_{G_2}(x) \big \vert \big \vert_\infty \right \} 
}
defines  a metric on the space of probabilistic automata on $\Sigma$.
\end{lem}
\Proof{
 Non-negativity and symmetry follows immediately. Triangular inequality follows from noting that $\big \vert \big \vert \phi^{s_1}_{G_1}(x) - \phi^{s_2}_{G_2}(x) \big \vert \big \vert_\infty$ is upper bounded by $1$, and therefore for any chosen order of the strings in $\Sigma^\star$, we have two $\ell_\infty$ sequences, which would satisfy the triangular inequality under the $\sup$ norm. The metric is well-defined since for any sufficiently long $s_1,s_2$, the symbolic derivatives at arbitrary $x$ are uniformly convergent to some linear combination of the rows of the corresponding $\Pitilde$ matrices. 
}
Now, we can establish that the class of ergodic, stationary QSPs with a finite number of causal states is PAC-learnable.
\begin{thm}[PAC-Learnability of QSPs]\label{thmPAC}
Ergodic, stationary QSPs for which the probabilistic Nerode equivalence has a finite index satisfies the following property:

For $\epsilon,\eta > 0$, and for every sufficiently long sequence $s$ generated by QSP $\mathcal{H}$, \algo computes $\mathcal{P}_\mathcal{H}'$ as an estimate for $\mathcal{P}_\mathcal{H}$ with:
\cgather{
Pr \big (\Theta(\mathcal{P}_\mathcal{H},\mathcal{P}_\mathcal{H}') \leqq \epsilon \big ) \geqq 1-\eta \label{eqPAC}
}
Asymptotic runtime is polynomial in $1/\epsilon,1/\eta, \vert s \vert$.
\end{thm}
\Proof{
 \algo construction and Corollary~\ref{corrightinveps} to Theorem~\ref{thmsymderiv} implies that, once the initial $\epsilon$-synchronizing string $x_0$ is identified, right extensions of $x$ (with non-zero probability of occurrence from the synchronized state) are $\epsilon'$-synchronizing where $\epsilon' = \epsilon C_0$, with $C_0 < \infty $ is as defined in Eq.~\eqref{eqC0}.  

If the target QSP has $\vert Q\vert$ states, then $\vert Q\vert $ states need to be visited with right extensions of the computed $\epsilon$-synchronizing string $x$. 
Hence, for any $\epsilon'' >0$, 
\caligns{
& Pr \big (\Theta(\mathcal{P}_\mathcal{H},\mathcal{P}_\mathcal{H}') \leqq C_0^{\vert Q \vert}\epsilon'' \big ) = 1 - Pr\big (\vert \vert  \phi^s(x_0) - \wp_{x_0}\Pitilde \vert \vert_\infty > \epsilon'' \big )\\
\Rightarrow & Pr \big (\Theta(\mathcal{P}_\mathcal{H},\mathcal{P}_\mathcal{H}') \leqq \epsilon \big )\geqq 1- e^{-\epsilon\vert s \vert C_0^{-\vert Q\vert}   O(1)} \mspace{20mu} \textrm{(Using Eq.~\eqref{eqconfi})}
}
Thus, for any $\eta > 0$, if we have $\vert s \vert = O(C_0^{\vert Q \vert} \frac{1}{\epsilon} \log \frac{1}{\eta})$, then the required condition of Eq.~\eqref{eqPAC} is met. Polynomial runtimes is established in Theorem~\ref{thmcomplexity}.
}
\begin{cor}[To Theorem~\ref{thmPAC}: Sample Complexity]
The input length required for PAC-learning with  \algo is asymptotically linear in $\frac{1}{\epsilon},\log \frac{1}{\eta}$, but exponential in the number of causal states $\vert Q\vert$.
\end{cor}
\Proof{
Immediate from Theorem~\ref{thmPAC}.
}
\begin{rem}[Sample Complexity]
The exponential asymptotic dependence of the sample complexity on the number of causal states of the target QSP should not be interpreted as inefficiency. Unlike standard treatments of PAC learning, here we do not have a set of independent samples as training, but a single long input stream. Noting that an input $s$ is composed of an exponential number of subtrings (summed over all lengths), the exponential dependence on $\vert Q\vert$ vanishes, if we treat this set of subsequences as the sample set. Thus, it is a matter of how one chooses to define the notion of sample complexity for this setting.
\end{rem}
\begin{rem}[Remark On Kearns' Hardness Result]
 We are immune to Kearns' hardness result~\cite{Kearns1994}, since   $\epsilon > 0$ enforces state distinguishability~\cite{RST94}, and furthermore, the restriction of our systems of interest to ergodic stationary dynamical systems, which produce termination-free traces, makes Kearns' particular construction with parity function~\cite{Kearns1994}  inapplicable.
\end{rem}
}
\section{Algorithm \xalgo: Cross-model Inference}\label{sec6}
We note that the notion of structural isomorphism between PFSA (See Definition~\ref{defstructiso}) extends naturally to XPFSA, with the output morph function in the latter  playing the role of the morph function in the former. In particular, we note that the assumed ergodicity of the processes and of the cross-talk map (See Definition~\ref{defdep}), implies that we have a result similar to Theorem~\ref{thmminrez}; namely that XPFSA have unique minimal realizations, which are strongly connected.
\begin{lem}[Existence Of Unique Strongly Connected Minimal XPFSA Realization]\label{lemXminrez}
 For stationary ergodic QSPs $\mathcal{H}_A,\mathcal{H}_B$    over   alphabets $\Sigma_A,\Sigma_B$, if the probabilistic cross-Nerode relation $\sim^{\mathcal{H}_A}_{\mathcal{H}_B}$  on $\Sigma^\star_A$ (with respect to a consistent, ergodic cross-talk map, see Definition~\ref{defdep}) has a finite index, then it has a  strongly connected XPFSA generator unique upto structural isomorphism.
\end{lem}
\Proof{
The argument is  identical to that in Theorem~\ref{thmminrez}, using the construction described in Lemma~\ref{lemnerodetoxpfsa}.
}
\vspace{4pt}

$\epsilon$-synchronization plays an important role in \algo. A corresponding notion of synchronization is necessary for inferring  XPFSAs. However, since XPFSAs model dependence between processes, and not the processes themselves, the notion $\epsilon$-synchronization in this case cannot be based solely on the XPFSA structure or its output morph. Specifically, since the transitions in a XPFSA lack the generation probabilities, it does not make sense to talk about synchronization in the same sense as of a PFSA. However, synchronization is still necessary to ensure that we infer the XPFSA states, and not distributions on them. We need  induced cross-distributions, $i.e.$, distributions on the XPFSA states given a observed string in the first process,  to make the notion of synchronization well-defined. 
\begin{defn}[Induced Cross-Distribution]\label{defindXd}
Given stationary ergodic QSPs $\mathcal{H}_A,\mathcal{H}_B$    over   alphabets $\Sigma_A,\Sigma_B$, a PFSA generator $G_A$ for $\mathcal{H}_A$, and the minimal XPFSA $B^A = (Q^A_B,\Sigma_A,\delta',\pitilde_{\Sigma_B})$ from $\mathcal{H}_A$ to $\mathcal{H}_B$, each  $x \in \Sigma^\star_A$ induces a distribution $\wp^{\mathcal{H}_A,\mathcal{H}_B}_x$ over $Q^A_B$ defined recursively as:
\begin{subequations}
\cgather{
\wp^{\mathcal{H}_A,\mathcal{H}_B}_\lambda \mapsto \base{G_A \psync B^A}_{B^A} \\ 
\wp^{\mathcal{H}_A,\mathcal{H}_B}_{x\sigma} \mapsto  \frac{\wp^{\mathcal{H}_A,\mathcal{H}_B}_{x}\Gamma_\sigma^{B^A}}{\norm{\wp^{\mathcal{H}_A,\mathcal{H}_B}_{x}\Gamma_\sigma^{B^A}}_1}
}\end{subequations}
where we assume $\wp^{\mathcal{H}_A,\mathcal{H}_B}_{x}$ is  a row vector, and $\Gamma_\sigma^{B^A}$ is the symbol-specific transformation matrix for $B^A$ using the output morph as the morph function (See Definition~\ref{defGamma}).
\end{defn}
\begin{lem}[Interpretation of Induced Cross-Distribution]\label{lemindXd}
Given stationary ergodic QSPs $\mathcal{H}_A,\mathcal{H}_B$    over   alphabets $\Sigma_A,\Sigma_B$, a PFSA generator $G_A$ for $\mathcal{H}_A$, and the minimal XPFSA $B^A = (Q^A_B,\Sigma_A,\delta',\pitilde_{\Sigma_B})$ from $\mathcal{H}_A$ to $\mathcal{H}_B$, the induced cross-distribution $ \wp^{\mathcal{H}_A,\mathcal{H}_B}_{x}$ satisfies:
\cgather{
\forall x \in \Sigma^\star_A, \wp^{\mathcal{H}_A,\mathcal{H}_B}_{x} \Pitilde_{\Sigma_B} =  \mathop{\mathbf{E}}_{y \in \Sigma^\star_A} \phi^{\mathcal{H}_A,\mathcal{H}_B}_{yx}
}
assuming as before that $\wp^{\mathcal{H}_A,\mathcal{H}_B}_{x}$ is  a row vector.
\end{lem}
\Proof{
Denoting the probability space induced by $\mathcal{H}_A$ as $(\Sigma^\omega_A, \mathfrak{B},\mu^A)$, and the equivalence class corresponding to state $q \in Q^A_B$ as $\mathcal{E}(q)$, we note:
\mltlne{
\mathop{\mathbf{E}}_{y \in \Sigma^\star_A} \phi^{\mathcal{H}_A,\mathcal{H}_B}_{yx} = \sum_{y \in \Sigma^\star_A} \mu^A(y \Sigma^\omega_A) \phi^{\mathcal{H}_A,\mathcal{H}_B}_{yx} 
\\ = \sum_{q \in Q^A_B} \sum_{yx \in \mathcal{E}(q) } \mu^A(y \Sigma^\omega_A) \phi^{\mathcal{H}_A,\mathcal{H}_B}_{yx} = \sum_{q \in Q^A_B} \left ( \sum_{yx \in \mathcal{E}(q) } \mu^A(y \Sigma^\omega_A) \right )  \pitilde_{\Sigma_B}(q,\cdot)
}
Finally, noting that Definitions~\ref{defprofdist} and \ref{defindXd} imply:
\cgather{
\wp^{\mathcal{H}_A,\mathcal{H}_B}_{x} = \sum_{yx \in \mathcal{E}(q) } \mu^A(y \Sigma^\omega_A)
}
completes the proof.
}
\begin{defn}[$\epsilon$-Synchronization of XPFSA]\label{defXsync}
For stationary ergodic QSPs $\mathcal{H}_A,\mathcal{H}_B$    over   alphabets $\Sigma_A,\Sigma_B$, and a XPFSA $B^A=(Q^A_B,\Sigma_A,\delta',\pitilde_{\Sigma_B})$ from $\mathcal{H}_A$ to $\mathcal{H}_B$, a string $x_0 \in \Sigma^\star_A$ is $\epsilon$-synchronizing with respect to $B^A$, if
\cgather{
\exists q \in Q^A_B, \norm{ \mathop{\mathbf{E}}_{y \in \Sigma^\star_A} \phi^{\mathcal{H}_A,\mathcal{H}_B}_{yx_0} - \pitilde_{\Sigma_B}(q,\cdot)}_\infty \leqq \epsilon
}
\end{defn}
The next result reduces the computation of a $\epsilon$-synchronizing string for a XPFSA to that for a particular PFSA. 
\begin{thm}[$\epsilon$-Synchronization of XPFSA via Projective Composition]\label{thmXsync}
Given stationary ergodic QSPs $\mathcal{H}_A,\mathcal{H}_B$    over   alphabets $\Sigma_A,\Sigma_B$, with $A=(Q,\Sigma_A,\delta,\pitilde)$ being a PFSA encoding $\mathcal{H}_A$, and  $B^A=(Q^A_B,\Sigma_A,\delta',\pitilde_{\Sigma_B})$ being a XPFSA from $\mathcal{H}_A$ to $\mathcal{H}_B$, a string $x_0 \in \Sigma^\star_A$ is $\epsilon$-synchronizing with respect to $B^A$ (in the sense of Definition~\ref{defXsync}), if $x_0$ is  $\epsilon$-synchronizing with respect to the PFSA $A \psync B^A$ (in the sense of Definition~\ref{defepsilonsynchro}).
\end{thm}
\Proof{
We note that Lemma~\ref{lemindXd} implies that for any $x_0 \in \Sigma^\star_A$,
\cgather{
\max_{i=1,\cdots,\vert Q^A_B \vert}\wp^{\mathcal{H}_A,\mathcal{H}_B}_{x_0} \big \vert_i   \geqq 1-\epsilon \notag \\ \Rightarrow \exists q \in Q^A_B, \norm{ \mathop{\mathbf{E}}_{y \in \Sigma^\star_A} \phi^{\mathcal{H}_A,\mathcal{H}_B}_{yx_0} - \pitilde_{\Sigma_B}(q,\cdot)}_\infty \leqq \epsilon
}
Noting that Definition~\ref{defindXd} implies:
\cgather{
\wp^{\mathcal{H}_A,\mathcal{H}_B}_{x_0} = \wp_{x_0}^{A \psync B^A}
}
completes the proof.
}
Thus, to $\epsilon$-synchronize the XPFSA $B^A$, we simply need to find an $\epsilon$-synchronizing string for the PFSA $A \psync B^A$ which is  a problem that we have already solved (See  Lemma~\ref{lemlimderiv} and Theorem~\ref{thmderivheap}). 
However, the definition of the derivative heap (See Definition~\ref{defderivheap}) would need to be suitably generalized (See Definition~\ref{defXderivheap}).

However, before we go into XPFSA inference, we note that the above reduction leads us to the following important corollary, which establishes that $\epsilon$-synchronizing strings exist for any $\epsilon> 0$.
\begin{cor}[To Theorem~\ref{thmXsync}: Existence of $\epsilon$-Synchronizing Strings for XPFSA]\label{corepsexist}
For any $\epsilon > 0$,  stationary ergodic QSPs $\mathcal{H}_A,\mathcal{H}_B$    over   alphabets $\Sigma_A,\Sigma_B$, and a given XPFSA $B^A$ from $\mathcal{H}_A$ to $\mathcal{H}_B$, there exists a string $x_0 \in \Sigma^\star_A$ that $\epsilon$-synchronizes  $B^A$.
\end{cor}
\Proof{
Follows immediately from Theorems~\ref{thmXsync} and \ref{thmepssynchro}.
}
\vspace{4pt}

Before we present our inference algorithm \xalgo, we need an effective approach to compute
cross-derivatives. First we generalize the count function introduced in Definition~\ref{defcount}.

\begin{defn}[Symbolic Cross-Count Function]\label{defcrosscount}
 For  strings $s_A,s_B$ over respective alphabets  $\Sigma_A, \Sigma_B$, the cross-count function $\#^{s_A,s_B}: \Sigma^\star_A \times \Sigma_B \rightarrow \mathbb{N}\cup \{0\}$,  counts the number of times a particular substring occurs in $s_A$, being followed immediately by a particular symbol in string $s_B$. The count is overlapping, $i.e.$, in  strings $s_A=000100,s_B=012212$, we count the number of occurrences of string $00$ in $s_A$, followed immediately by symbol $2$ in $s_B$, as:
\caligns{
& \underline{00}0100  && 0\underline{00}100\\
& 01\underline{2}212 && 012\underline{2}12
}
implying $\#^{s_A,s_B}(00,2) = 2$.
\end{defn}
And, then we define an estimator for cross-derivatives using the cross-count function.
\begin{defn}[Cross-derivative Estimator]\label{defcrossest}
For  strings $s_A,s_B$ over respective alphabets  $\Sigma_A, \Sigma_B$, the cross-derivative estimator $\phi^{s_A,s_B}: \Sigma^\star_A \rightarrow [0,1]^{\vert \Sigma_B \vert}$ is a non-negative vector summing to unity, with entries defined as:
\cgather{
\forall x \in \Sigma^\star_A, \phi^{s_A,s_B}_x)\big \vert_i =  \frac{\#^{s_A,s_B}(x,\sigma_i)}{\displaystyle \sum_{\sigma_i \in \Sigma_B}\#^{s_A,s_B}(x,\sigma_i)}
}
\end{defn}

And, as before (See Theorem~\ref{thmsymderiv}),  we have the following convergence.

\begin{lem}[$\epsilon$-Convergence for Cross-derivatives]\label{lemsymderivX}
For  stationary ergodic QSPs $\mathcal{H}_A,\mathcal{H}_B$, over   $\Sigma_A,\Sigma_B$, producing respective strings $s_A,s_B$, and a given XPFSA $B^A$ from $\mathcal{H}_A$ to $\mathcal{H}_B$, if $x \in \Sigma^\star_A$ is $\epsilon$-synchronizing, then:
\cgather{
\forall \epsilon > 0,  \lim_{\vert s_A \vert, \vert s_B \vert \rightarrow \infty} \norm{ \phi^{s_A,s_B}_x - \pitilde_{\Sigma_B}([x],\cdot) }_\infty \leqq_{a.s.} \epsilon
}
\end{lem}
\Proof{
Since $\phi^{s_A,s_B}_x$ is an empirical distribution for $\phi^{\mathcal{H}_A,\mathcal{H}_B}_x$, the result follows from Glivenko-Cantelli theorem~\cite{Fl70}, using the argument of Theorem~\ref{thmsymderiv}.
}


In close analogy to  PFSA inference described in Section~\ref{sec5}, here we   seek  similar cross-derivatives, and ``merges''  string fragments at which the derivatives turn out to be similar, $i.e.$ define them to reach the same state in the inferred XPFSA. First, we  need to generalize the definition of the derivative heap as follows:

\begin{defn}[Cross-Derivative Heap]\label{defXderivheap}
  For  stationary ergodic QSPs $\mathcal{H}_A,\mathcal{H}_B$, over   $\Sigma_A,\Sigma_B$, producing respective strings $s_A,s_B$,  a cross-derivative heap $\mathcal{D}^{s_A,s_B}: 2^{\Sigma^\star_A} \rightarrow \mathscr{D}(\vert \Sigma_B \vert -1)$ is the set of probability distributions over $\Sigma_B$ calculated for a  subset of strings  $L \subset \Sigma^\star_A$ as:
\cgather{
\mathcal{D}^{s_A,s_B}(L) = \big \{ \phi^{s_A,s_B}_x: x \in L \subset \Sigma^\star_A\big \}
}
\end{defn}

We note that Lemma~\ref{lemlimderiv} and Theorem~\ref{thmderivheap} generalizes immediately:

\begin{lem}[Cross-derivative Heap Covergence]\label{lemXlimderiv}
\begin{enumerate} \item  Define:
\cgather{
\mathcal{D}_\infty \triangleq \lim_{\vert s_A\vert ,\vert s_B \vert \rightarrow \infty }\lim_{L \rightarrow \Sigma^\star_A} \mathcal{D}^{s_A,s_B}(L)
}
If $\mathscr{U}_\infty$ is the convex hull of $\mathcal{D}_\infty$,  $u$ is a vertex of $\mathscr{U}_\infty$ and $\pitilde_{\Sigma_B}$ is the output morph the XPFSA from  $\mathcal{H}_A$ to $\mathcal{H}_B$, then we have:
\cgather{
\exists q \in Q, \textrm{such that } u=\pitilde_{\Sigma_B}(q,\cdot)
}

\item  For  stationary ergodic QSPs $\mathcal{H}_A,\mathcal{H}_B$, over   $\Sigma_A,\Sigma_B$, producing respective strings $s_A,s_B$,  let $\mathcal{D}^{s_A,s_B}(L)$ be computed with $L=\Sigma^{O(log(1/\epsilon))}$. If for   $x_0 \in \Sigma^{O(log(1/\epsilon))}_A$,  $\phi^{s_A,s_B}_{x_0}$  is a vertex of the convex hull of $\mathcal{D}^{s_A,s_B}(L)$, then we have:
\cgather{
Pr(\textrm{$x_0$ is not $\epsilon$-synchronizing}) \leqq e^{-\vert s_A \vert \epsilon p_0}
\label{eq152}}
 where $p_0$ is the probability of encountering  $x_0$ in $s_A$.
\end{enumerate}
\end{lem}
\Proof{
See Lemma~\ref{lemlimderiv} and Theorem~\ref{thmderivheap}.
}

\subsection{Implementation Steps For \xalgo}\label{secXinfer}
We have two steps in \xalgo which infers the strongly connected minimal realization $B^A=(Q^A_B,\Sigma_A,\delta',\pitilde_{\Sigma_B})$:

\begin{enumerate}
\item { \textit{Identification of $\epsilon$-synchronizing string $x_0$:} }
Construct a derivative heap $\mathcal{D}^{s_A,s_B}(L)$ using the observed traces $s_A,s_B$. (Definition~\ref{defXderivheap}), and set $L = \log_{\vert \Sigma_A \vert} 1/\epsilon $. We then identify a vertex of the convex hull for $\mathcal{D}_\infty$, via any standard algorithm for computing the hull~\cite{QHULL}. Choose $x_0$ as the string mapping to this vertex.
\item {\textit{Identification of the transition  function:}} We   generate $\delta'$ as follows: For each state $q$, we associate a string identifier $x^{id}_q \in x_0\Sigma^\star_A$, and a probability distribution $h_q'$ on $\Sigma_B$ (which is an approximation of the  $\Pitilde_{\Sigma_B}$-row corresponding to state $q$). We extend the structure recursively:
\begin{enumerate} 
\item Initialize the set $Q$ as $Q=\{q_0\}$, and set $x^{id}_{q_0}=x_0$, $h_q = \phi^s(x_0)$. 
\item 
\begin{itemize}
\item For each state $q \in Q^A_B$, compute for each symbol $\sigma \in \Sigma_A$, find symbolic derivative $\phi^{s_A,s_B}_{x^{id}_q \sigma}$. 
\item If $\norm{\phi^{s_A,s_B}_{x^{id}_q \sigma} - h_{q' \sigma}}_\infty \leqq \epsilon$ for some $q' \in Q^A_B$, then define $\delta'(q,\sigma) = q'$. 
\item If, on the other hand, no such $q'$ can be found in $Q^A_B$, then add a new state $q'$ to $Q^A_B$, and define  
\calign{
x^{id}_{q'} & =x^{id}_q\sigma\\
h_{q'} & = \phi^{s_A,s_B}_{x^{id}_q \sigma}
}
\item The process terminates when every $q \in Q^A_B$ has  a target state, for each $\sigma \in \Sigma_A$. 
\item Then, if necessary, we ensure strong connectivity using Tarjan's algorithm~\cite{tar146}.
\item The output morph function $\pitilde_{\Sigma_B}$ is given by:
\cgather{
\forall q \in Q^A_B, \pitilde_{\Sigma_B}(q,\cdot) = h_{q}
}
\end{itemize}
\end{enumerate}
\end{enumerate}
\DETAILS{
\subsection{Complexity of \xalgo \& PAC Learnability}
Asymptotic time complexity for \xalgo is essentially identical to that of \algo. We have the following immediate result:
\begin{thm}[Time Complexity]\label{thmXcomplexity}
Assuming the input streams are longer compared to the respective alphabet sizes, the asymptotic runtime complexity of \xalgo is:
\cgather{
\mathscr{T} = O\left (\frac{\vert \Sigma_A \vert ( \vert s_A\vert + \vert s_B \vert )} {\epsilon} \right )
}
\end{thm}
\Proof{
Follows from the argument  in Theorem~\ref{thmcomplexity}, noting that the step corresponding to C1 (See Theorem~\ref{thmcomplexity}) takes $O(1/\epsilon (\vert s_A \vert + \vert s_B\vert))$ time, the step corresponding to C2 takes $O(1/\epsilon \vert \Sigma_B\vert)$ time and the step corresponding to C3 takes $O(\vert Q\vert \vert \Sigma_A\vert (\vert s_A \vert + \vert s_B\vert) )$ time.
}
%

\begin{lem}[Metric For Crossed Probabilistic Automata]\label{lemXmetric}
 For  stationary ergodic QSPs $\mathcal{H}_A,\mathcal{H}_A'$ over alphabets $\Sigma_A$, and $\mathcal{H}_B, \mathcal{H}_B'$   over   $\Sigma_B$, let $G_1,G_2$ be XPFSAs representing the dependencies from  $\mathcal{H}_A$ to $\mathcal{H}_B$ and from $\mathcal{H}_A'$ to $\mathcal{H}_B'$ respectively. If $s_A,s_B,s_A',s_B'$ are streams generated by  $\mathcal{H}_A,  \mathcal{H}_B,\mathcal{H}_A',  \mathcal{H}_B'$ respectively, then:
\cgather{
\Theta_{\Sigma_A,\Sigma_B}(G_1,G_2) = \sup_{x \in \Sigma^\star_A } \left \{  \lim_{\vert s_A \vert, \vert s_B \vert, \vert s_A' \vert, \vert s_B' \vert \rightarrow \infty }\norm{ \phi^{s_A,s_B}_x - \phi^{s_A',s_B'}_x }_\infty \right \} 
}
defines  a metric on the space of crossed probabilistic automata that represent dependencies from processes over $\Sigma_A$ to processes over $\Sigma_B$.
\end{lem}
\Proof{
See Lemma~\ref{lemmetric}.
}
\begin{thm}[PAC-Learnability]\label{thmPACX}
The dependency between two ergodic stationary QSPs $\mathcal{H}_A,\mathcal{H}_B$ over respective alphabets $\Sigma_A,\mathcal{H}_B$ with an ergodic, consistent cross-talk map (Definition~\ref{defdep}) is learnable by \xalgo  in the following sense:

If $G$ denotes the true XPFSA, then $\forall \epsilon, \eta > 0$,  \xalgo learns an estimated XPFSA $G'$ with:
\cgather{
Pr \big (\Theta_{\Sigma_A,\Sigma_B}(G,G') \geqq \epsilon \big ) < \eta \label{eqPACX}
}
and the asymptotic runtime is polynomial in $1/\epsilon, \vert s_A\vert + \vert s_B \vert, 1 /\eta$.
Additionally, to satisfy the above condition, we need:
\cgather{
 \vert s_A\vert + \vert s_B \vert = O\left (\frac{1}{\epsilon} C^{\vert Q \vert} \log \frac{1}{ \eta} \right )
}
where $C < \infty$, and $Q$ is the set of states in the inferred XPFSA.
\end{thm}
\Proof{
On account of Theorem~\ref{thmXsync}, the result in Eq.~\eqref{eqPACX} follows  from the same argument as in Theorem~\ref{thmPAC} (using Eq.~\ref{eq152} instead of Eq.~\ref{eqconfi}). The sample complexity also follows from the same argument, with the modification of including the sum of string lengths arising from 
Theorem~\ref{thmXcomplexity}.
}
}
\section{Generation of Causality Networks}\label{sec7}
The coefficient of causal dependence was introduced in Definition~\ref{defcoeffcd}, to quantify the reduction in uncertainty of the next symbol in the second stream from observations made in the first. It was  clear that this coefficient is asymmetric, in the sense that in general for two ergodic stationary QSPs $\mathcal{H}_A,\mathcal{H}_B$, we have:
\cgather{
\gamma^{\mathcal{H}_A}_{\mathcal{H}_B} \neq \gamma^{\mathcal{H}_B}_{\mathcal{H}_A}
}
Additionally, the  example in Figure~\ref{figEX1} demonstrates that the coefficients  do indeed capture  directional dependence, $i.e.$, the direction of causality flow between two processes. We can extend this idea to a set of interdependent processes; the calculation of the pairwise coefficients would then reveal the possibly intricate flow of causality, leading to what we call the \textit{inferred causality network}.

Consider the set of $n$ ergodic stationary processes
\cgather{ \mathscr{H} = \left \{\mathcal{H}_{i}: i =1,\cdots,n \right  \}}
evolving over respective  alphabets $\Sigma_{i}$, which need not be distinct or have the same cardinalities.

Let the processes possibly  depend on each other  via cross-talk maps that satisfy the properties set forth in Definition~\ref{defdep}. Additionally, assume that the relevant cross-Nerode equivalences have finite indices, implying that there exist XPFSA
that encode the inter-process dependencies.


\begin{notn}[Set of Processes and Inferred Machines]\label{not6} We introduce some notation to denote the relevant inferred machines.

\begin{itemize}
\item $s^i \in \Sigma_i^\star$ denotes the string generated by the process $\mathcal{H}_i $
\item $H^i_j$ with  $i \neq j$ denotes the XPFSA from process $\mathcal{H}_i $ to $\mathcal{H}_j$, where
\cgather{
H^i_j = (Q^i_j,\Sigma_i, \delta^i_j, \pitilde^i_j)
}
\item $H^i$ denotes the PFSA encoding the process $\mathcal{H}_i $ itself, where
\cgather{
H^i = (Q^i,\Sigma_i, \delta^i, \pitilde^i)
}
\item The coefficient of dependence from  $\mathcal{H}_i $ to  $\mathcal{H}_j $ is denoted $\gamma^i_j$
\end{itemize}

\end{notn}

We introduce the stream-run function, which would simplify the computation of the coefficients of dependence in the sequel.
\begin{defn}[Stream-run Function]\label{defstreamrun}
Given a strongly connected labeled graph $G=(Q,\Sigma,\delta)$ be a strongly connected graph with $Q$ as the set of nodes, such that there is a  labeled edge $q_i \xrightarrow{\sigma} q_j$ for $q_i,q_j \in Q$ iff $\delta(q_i,\sigma) = q_j$, and string $s \in \Sigma^\star$, the stream-run function $\rho(G,s)$ is a real-valued vector of length $\vert Q \vert$ with
\cgather{
\forall i, \rho(G,s) \vert_i \in [0, 1], \textrm{and } \sum_i \rho(G,s) \vert_i =1
}
and is computed using Algorithm~\ref{algostreamrun}.
\end{defn}
We need the following technical result, that establishes the connection between the stream-run function and the projected distribution introduced in Definition~\ref{defprofdist}.
\begin{algorithm}[t]\sffamily \fontsize{8}{9}\selectfont
\SetKw{BRK}{break}
\DontPrintSemicolon
\KwIn{Strongly connected labeled graph $G=(Q,\Sigma,\delta)$, string $s \in \Sigma^\star$ }
\KwOut{ $\rho(G,s)$ }
\BlankLine

Initialize zero vector $v$ of length $\vert Q\vert$\;
Choose random node $q_{k^\star} \in Q$\;
$q_{current} \leftarrow q_{k^\star}$\;
$v_{k^\star} \leftarrow 1 $\;
\For{$i \leftarrow 1 \textrm{ to } \vert s \vert$}{
$q_{current} \leftarrow \delta(q_{current},s_i)$\;
\If{$q_{current} == q_k$}
{
$v_{k} \leftarrow v_{k}+1 $\;
}
}
 \tcc*[f]{Normalize vector}\;
\For{$k \leftarrow 1 \textrm{ to } \vert Q \vert$}{

$v_{k} \leftarrow v_{k} / \sum_j v_j $\;
}

\KwRet $\rho(G,s) \leftarrow v$\; 

\captionN{Stream-run Function}\label{algostreamrun}
\end{algorithm}
%
\begin{algorithm}[t]\sffamily \fontsize{8}{9}\selectfont
\SetKw{BRK}{break}
\DontPrintSemicolon
\KwIn{ $\epsilon$, $s^i$, $s^j$ }
\KwOut{ Estimate $\overline{\gamma}^i_j$ for $\gamma^i_j$ }
\BlankLine
\tcp*[h]{Compute XPFSA  }\;
Compute $H^i_j$ \;
\BlankLine
\tcp*[h]{Compute $\phi^{s^j}_\lambda$  }\;
 $\displaystyle r \leftarrow [0 \cdots 0]$\;
\tcp*[h]{Length = $\vert \Sigma_j\vert $}\;
\BlankLine
\For{$k \leftarrow 1 \textrm{ to } \vert s^j \vert $}{
\If{$s^j[k] == \sigma_\ell$}{
$r_\ell \leftarrow r_\ell +1 $\;
}
}
\BlankLine

\tcp*[h]{Compute denominator }\;

$\displaystyle h_0 \leftarrow \sum_{k=1}^{\vert \Sigma_j \vert} r_k \log_2(r_k)$\;

\BlankLine

\tcp*[h]{Compute numerator }\;
$\displaystyle h_1 \leftarrow 0$\;
 $\displaystyle u \leftarrow \rho(H^i_j,s^i)$\;
\For{$k \leftarrow 1 \textrm{ to } \vert Q^i_j \vert $}{
$ \displaystyle h[k] \leftarrow \sum_{\sigma_\ell \in \Sigma_j} \pitilde^i_j(q_k,\sigma_\ell)\log_2 \left ( \pitilde^i_j(q_k,\sigma_\ell)\right )$\;
$ \displaystyle h_1 \leftarrow h_1 + u_k h[k]$\;
}

\KwRet $\overline{\gamma}^i_j \leftarrow \dfrac{h_1}{h_0}$\; 

\captionN{Efficient Computation of the Coefficient of Dependence}\label{algocoeff}
\end{algorithm}

\begin{lem}[Computing  Projected Distribution]\label{lemstreamrunprojdist}
Let $s \in \Sigma^\star$ be generated by an ergodic stationary QSP $\mathcal{H}$ with a finite index Nerode equivalence induced by the underlying probability space $(\Sigma^\omega, \mathfrak{B},\mu)$. Let the minimal PFSA encoding encoding the QSP be $G=(Q,\Sigma,\delta,\pitilde)$. Additionally, let $G'=(Q',\Sigma,\delta')$ be a strongly connected graph with $Q'$ as the set of nodes, such that there is a  labeled edge $q_i \xrightarrow{\sigma} q_j$ for $q_i,q_j \in Q'$ iff $\delta'(q_i,\sigma) = q_j$. 
Then, we have:
\cgather{
 \lim_{\vert s \vert \rightarrow \infty}\rho(G',s)=_{a.s.} \base{G \psync G'}_{G'}
}
\end{lem}
\Proof{
Since $\mathcal{H}$ is ergodic and stationary, and $G \sync G'$ is a non-minimal but strongly connected realization, it follows that $\rho(G \sync G', s)$ converges almost surely to the unique stationary distribution $\wp^\otimes$ on the state space of $G \sync G'$. 
Consider the paths through $G \sync G'$ and through $G'$ for the string $s$ in the course of computing the respective stream-run functions $\rho(G \sync G',s)$ and $\rho(G',s)$.
Noting that the count of state visits $v_{(q,q')}$ for the states $(q,q')$ in $G \sync G'$ relates to the count of   visits $v_{q'}$ for state $q'$ in $G'$ as:
\cgather{
\forall q' \in Q', v_{q'} = \sum_{q \in Q} v_{(q,q')}
}
We conclude that:
\cgather{
\rho(G',s) \xrightarrow[\vert s \vert \rightarrow \infty]{a.s.} u
}
where the vector $u$ satisfies:
\cgather{
u_i = \sum_{q \in Q} \wp^\otimes_{(q,q_i)}
}
Recalling the Definition~\ref{defprofdist}  completes the proof.
}
Based on Lemma~\ref{lemstreamrunprojdist},  Algorithm~\ref{algocoeff} computes the coefficient of dependence  avoiding explicit  computation of the projective composition. Next, we establish correctness and complexity of the algorithm.
\begin{thm}[Error Bound \& Complexity of Algorithm~\ref{algocoeff}]\label{thmalgocoeff}
We have:
\begin{enumerate}
\item Given the parameter $\epsilon$ for XPFSA inference, the absolute error in the estimated coefficient  $\overline{\gamma}^i_j$ (See Algorithm~\ref{algocoeff}) satisfies:
\mltlne{
\forall \epsilon \in ( 0,1/2],  \lim_{{\vert s^i \vert , \vert s^j \vert \rightarrow \infty}} \left \vert \gamma^i_j - \overline{\gamma}^i_j \right \vert \\ \leqq_{a.s.} \frac{1}{\entropy{\vartheta}} \left ( \epsilon \log_2 \frac{\vert \Sigma_j \vert-1}{\epsilon} + (1-\epsilon) \log_2 \frac{1}{1-\epsilon}  \right )  
}
where $\sigma_\ell \in \Sigma_j$ occurs with probability $\vartheta_\ell$  in process  $\mathcal{H}_j$.
\item Assuming  $\vert Q^i_j \vert \ll \frac{1}{\epsilon}$, and $\vert s^i \vert \approx \vert s^j \vert $, the asymptotic run-time complexity of Algorithm~\ref{algocoeff} is $O\left (\tfrac{1}{\epsilon}\vert s^i\vert\vert \Sigma^j \vert\right)$, $i.e.$, the same as for computing only the XPFSA $H^i_j$.
\end{enumerate}
\end{thm}
\Proof{ 
As before, we assume the streams $s^i,s^j$ to be generated by the processes $\mathcal{H}_i$, $\mathcal{H}_j$ respectively.

 Statement 1): The  denominator of $\gamma^i_j$ is given by $\entropy{\wp_\lambda^j\pitilde^j}$ (Lemma~\ref{lemcoeffcd1}), which is the  vector of probabilities with which different symbols appear in $s^j$. It follows  from ergodicity, that $r$ in lines 3-5 in Algorithm~\ref{algocoeff} converges almost surely to the denominator. Lemma~\ref{lemstreamrunprojdist} guarantees that $u$ (lines 8-11) converges almost surely to $\base{H^i \psync H^i_j}_{H^i_j}$. 


Assume that  the error in  infinity norm between the  inferred and actual vectors for any  row of $\Pitilde^i_j$ is bounded above by some $\epsilon \in (0, 1/2]$ almost surely. We refer to this as Assumption A. 

Now, let  $u^0=\base{H^i \psync H^i_j}_{H^i_j}$, and  for all $q_k \in Q^i_j$ the true probability vector corresponding to the inferred $\pitilde^i_j(q_k,\cdot) $  be $e(q_k,\cdot)$.
Also, let:
\cgather{
w^0 =  \begin{pmatrix}
\entropy{e(q_1,\cdot)} \\
\vdots 
\end{pmatrix}, w = \begin{pmatrix}
\entropy{\pitilde^i_j(q_1,\cdot)} \\
\vdots 
\end{pmatrix}
}
Then, we have:
\calign{
\left \vert \gamma^i_j - \overline{\gamma}^i_j \right \vert &=  \frac{\left \vert \left \langle u^0,w^0 
\right \rangle - \langle u, w^0\rangle  +  \langle u, w^0\rangle - \langle u, w\rangle \right \vert }{\entropy{r}} \\
&=  \frac{\left \vert \left \langle u^0-u,w^0 
\right \rangle   +  \langle u, w^0- w\rangle \right \vert }{\entropy{r}} \\
&\leqq  \frac{\left \vert \left \langle u^0-u,w^0 
\right \rangle   \right \vert }{\entropy{r}} +  \frac{\left \vert  \langle u, w^0- w\rangle \right \vert }{\entropy{r}}
}
We note that $u \xrightarrow{a.s.} u^0$, and $\entropy{r} \xrightarrow{a.s.} \entropy{\vartheta}$, and by Assumption A:
\cgather{
\forall q_k \in Q^i_j, \norm{e(q_k,\cdot) - \pitilde^i_j(q_k,\cdot)}_\infty \leqq_{a.s} \epsilon
}
which implies (See \cite{CL14e}, Lemma 7) that if $\epsilon \leqq 1/2$, then we have:
\cgather{
\forall q_k \in Q^i_j, \vert w^0_k - w_k \vert \leqq  \epsilon \log_2 \frac{\vert \Sigma_j \vert-1}{\epsilon} + (1-\epsilon) \log_2 \frac{1}{1-\epsilon} 
}
Hence, we conclude, that given Assumption A, we have:
\cgather{
\lim_{\mathclap{\vert s^i \vert , \vert s^j \vert \rightarrow \infty}} \left \vert \gamma^i_j - \overline{\gamma}^i_j \right \vert \leqq_{a.s.} \frac{1}{\entropy{\vartheta}} \left ( \epsilon \log_2 \frac{\vert \Sigma_j \vert-1}{\epsilon} + (1-\epsilon) \log_2 \frac{1}{1-\epsilon}  \right )
}

Now,  Definition~\ref{lemXmetric} implies that Assumption A is equivalent to:
\cgather{
\Theta_{\Sigma_i,\Sigma_j}(H^i_j, \overline{H^i_j}) \leqq \epsilon
}
where $H^i_j, \overline{H^i_j}$ are respectively the true and estimated XPFSAs for the cross-dependency from the process $\mathcal{H}_i $ to the process $\mathcal{H}_j$.
It follows from Theorem~\ref{thmPACX}, that:
\mltlne{\forall \epsilon \in ( 0,1/2],\\
Pr\left ( \lim_{{\vert s^i \vert , \vert s^j \vert \rightarrow \infty}} \left \vert \gamma^i_j - \overline{\gamma}^i_j \right \vert > \frac{1}{\entropy{\vartheta}} \left ( \epsilon \log_2 \frac{\vert \Sigma_j \vert-1}{\epsilon} + (1-\epsilon) \log_2 \frac{1}{1-\epsilon}  \right )  \right ) \\ \leqq \lim_{{\vert s^i \vert , \vert s^j \vert \rightarrow \infty}} e^{\vert s^i \vert \epsilon p_0} = 0
}
which establishes Statement 1).

Statement 2) follows immediately from Theorem~\ref{thmXcomplexity}. 
}
\begin{rem}
We note that $\frac{1}{\entropy{\vartheta}} > 0 $ implies:
\cgather{
 \lim_{\epsilon \rightarrow 0^+}\epsilon \log_2 \frac{\vert \Sigma_j \vert-1}{\epsilon} + (1-\epsilon) \log_2 \frac{1}{1-\epsilon} = 0
} 
$i.e.$, the bound established in Theorem~\ref{thmalgocoeff} is   small for small values of $\epsilon$. It is important to consider the implication of the factor $1/\entropy{\vartheta}$. In particular, for the bound to be finite, we must assume that the process $\mathcal{H}_j$ does not only produce a single repeated symbol, since in that case $\entropy{\vartheta}=0$. Also,  if the process $\mathcal{H}_j$ has a single symbol occurring with an  overwhelmingly high probability, then $\entropy{\vartheta}$ would be small, which would then imply that  a longer $s^j$ is required. This observation is relevant in applications with relatively rare events; $e.g.$, networks of  spiking neurons, or when attempting to construct causality networks from global seismic data.  
\end{rem}
\DETAILS{
\subsection{Prediction Using Crossed Probabilistic Automata}
Inferred cross-talk between data streams  may be exploited for predicting the future evolution of the processes under consideration. We use the same notation as before. However, in addition to the set of  $n$ ergodic processes $\mathscr{H}=\{ \mathcal{H}^i: i = 1,\cdots, n\}$, we consider an additional process $\mathcal{H}_B$, evolving over the  alphabet $\Sigma_B$ (with the same standard assumptions). We are interested in predicting the future evolution of $\mathcal{H}_B$.
\begin{notn}[Additional Notation]\label{not7}
\begin{itemize}
\item  In accordance  to the naming  scheme described in the previous section (See Notation~\ref{not6}), the XPFSA from $\mathcal{H}_i$ to $\mathcal{H}_B$  is denoted  $H^i_B = (Q^{i}_B,\Sigma_{i}, \delta^{i}_B,\pitilde^{i}_B )$.
\item  $s^B \in \Sigma^\star_B$ is the string observed in the process $\mathcal{H}_B$.
\item In addition to the strings $s^i$ (See Notation~\ref{not6}), we observe relatively short strings $x^i \in \Sigma_{i}^\star, i=1,\cdots,n$ respectively in the $n$ processes $\mathcal{H}_{i}$, which represent the immediate   histories. 

\item In particular, we  use $s_i$ for the stream from process $\mathcal{H}_i$ when we are inferring machines, and use $x^i$ when we need a short history for state localization (explained in the sequel).


\item $\tau^i , i=1,\cdots, n$ denotes the  expected next-symbol distribution in process $\mathcal{H}_B$ computed using the cross-talk or dependency from $\mathcal{H}_i$ to $\mathcal{H}_B$ (assuming that the immediate history observed in the former is $x^i$).
Thus, we have:
\cgather{
\forall \mathcal{H}_i \in \mathscr{H}, \left \{ \begin{array}{l}
\forall k , \tau^i_k \in [0,1] \\
\sum_{k=1}^{\vert \Sigma_B \vert} \tau^i_k = 1
\end{array}\right.
}
\end{itemize}
\end{notn}
\begin{algorithm}[t]\sffamily \fontsize{8}{9}\selectfont
\SetKw{BRK}{break}
\DontPrintSemicolon
\KwIn{ $\epsilon$, $s^i$, $s^B$, $x^i$ }
\KwOut{Predicted next-symbol distribution $\tau^i$ }
\BlankLine
\tcp*[h]{Compute PFSA \& XPFSA  }\;
Compute $H^i=(Q^i,\Sigma_i,\delta^i,\pitilde^i)$ using $s^i, \epsilon$\;
Compute $H^i_B=(Q^i_B,\Sigma_i,\delta^i_B,\pitilde^i_B)$ using $s^i, s^B,\epsilon$ \;
$ G \leftarrow H^i \psync H^i_B$\;
Compute stationary distribution $\wp^G_\lambda$\;
\BlankLine

\ForEach{$ \sigma \in \Sigma_i$ }{
Compute $\Gamma^G_\sigma$\;
}
\BlankLine

$\wp^0 \leftarrow \wp^G_\lambda$\;
\For{$ k \leftarrow 1 \textrm{ to } \vert x^i\vert$ }{
$\displaystyle \wp^0 \leftarrow \frac{\wp^0 \Gamma^G_{x^i_k}}{\norm{\wp^0 \Gamma^G_{x^i_k}}_1}$\;
}
\BlankLine

\KwRet $\tau^i \leftarrow \wp^0 \Pitilde^i_B$\; 

\captionN{Prediction of Next-symbol Distribution From Cross-talk}\label{algopred}
\end{algorithm}

\begin{lem}
Let $\displaystyle G \triangleq H^i \psync H^i_B$, and let  $\wp^G_{x^i}$ be the realized distribution over the states of PFSA  $G=(Q^i_B,\Sigma_i,\delta^i_B,\pitilde')$, given the  occurrence of string $x^i \in \Sigma^\star_i$ beginning with  the stationary distribution $\wp^G_\lambda$. Then:
\cgather{
\tau^i = \wp^G_{x^i} \Pitilde^i_B
}
where $\wp^G_{x^i}$ is assumed to be a row vector.
\end{lem}
\Proof{
We note that Lemma~\ref{lemindXd} tells us:
\cgather{
\tau^i = \mathop{\mathbf{E}}_{y \in \Sigma^\star_{i}} \phi^{\mathcal{H}_i,\mathcal{H}_B}_{yx^i} = \wp^{\mathcal{H}_{i},\mathcal{H}_B}_{x^i} \Pitilde_{B}^{i}
}
The result then follows from Definition~\ref{defcanon} (Canonical Representations)  and Definition~\ref{defindXd} (Induced Cross-Distribution).
}

Algorithm~\ref{algopred} illustrates the pseudo-code for computing $\tau^i$.

\subsubsection{Fusion of Individual Predictions: }
The problem of fusing the predictions $\tau^i$ for the processes in $\mathscr{H}$ to yield the ``best'' prediction $\overline{\tau}$, only admits a heuristic solution (at least with no further information). One approach to carry out this fusion is simply to take the weighted average of the predicted distributions, with the weights chosen to be normalized coefficients of dependence:
\cgather{
\textrm{Prediction Fusion:} \mspace{40mu} \overline{\tau} = \sum_{i=1}^n
 \left (\frac{\gamma^i_B}{\sum_k \gamma^k_B} \right )  \tau^i\mspace{60mu}}

This combination strategy  assigns zero weight to processes for which the corresponding coefficient of dependence is null. 
}
\begin{figure}[t]
\centering
\includegraphics[width=3in]{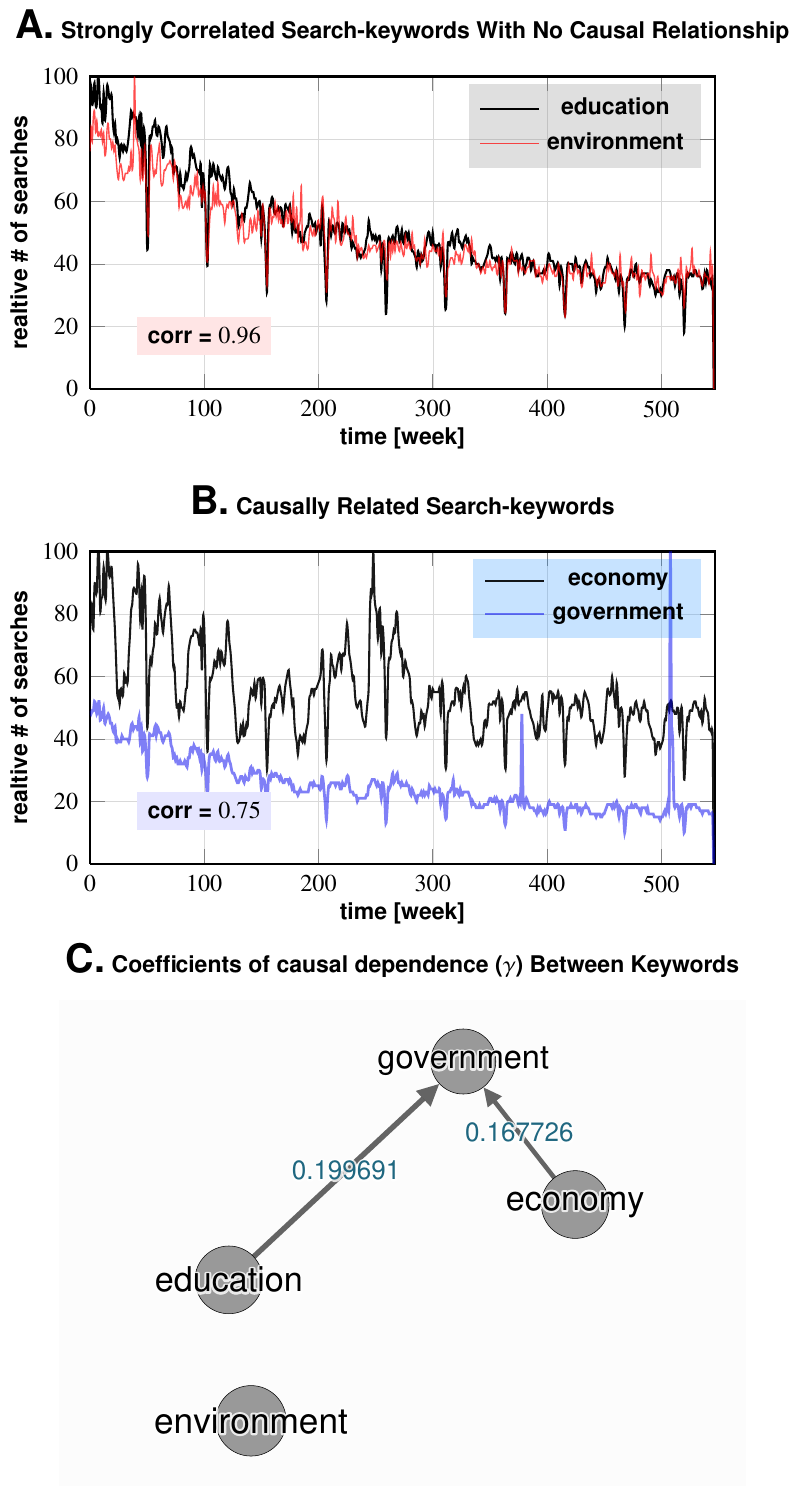}

\captionN{Illustration that high  statistical correlation does not signal a high degree of  causal dependency. Plate A: Strongly positively correlated ($corr=0.96$) search-frequeny data for keywords ``education'', ``environment''. Plate B: Positively correlated  ($corr=0.75$) search-frequeny data for keywords ``economy'', ``government''. Plate C shows that the data in plate A  have little causal dependence in either direction, while those in plate B have a directional causal dependence. The weights on the arcs in plate C are values for the inferred coefficient of causal dependence $\gamma$. As per definition~\ref{defGamma}, ``education'' $\xrightarrow{0.199691}$ ``government'' implies: $1$ bit of information from the search-frequency data  for ``education'' reduces the uncertainty in the immediate future of the data for ``government'' by $0.199691$ bits.}\label{figcaus2}
\end{figure}
\begin{figure*}[!ht]

\includegraphics[width=7.5in]{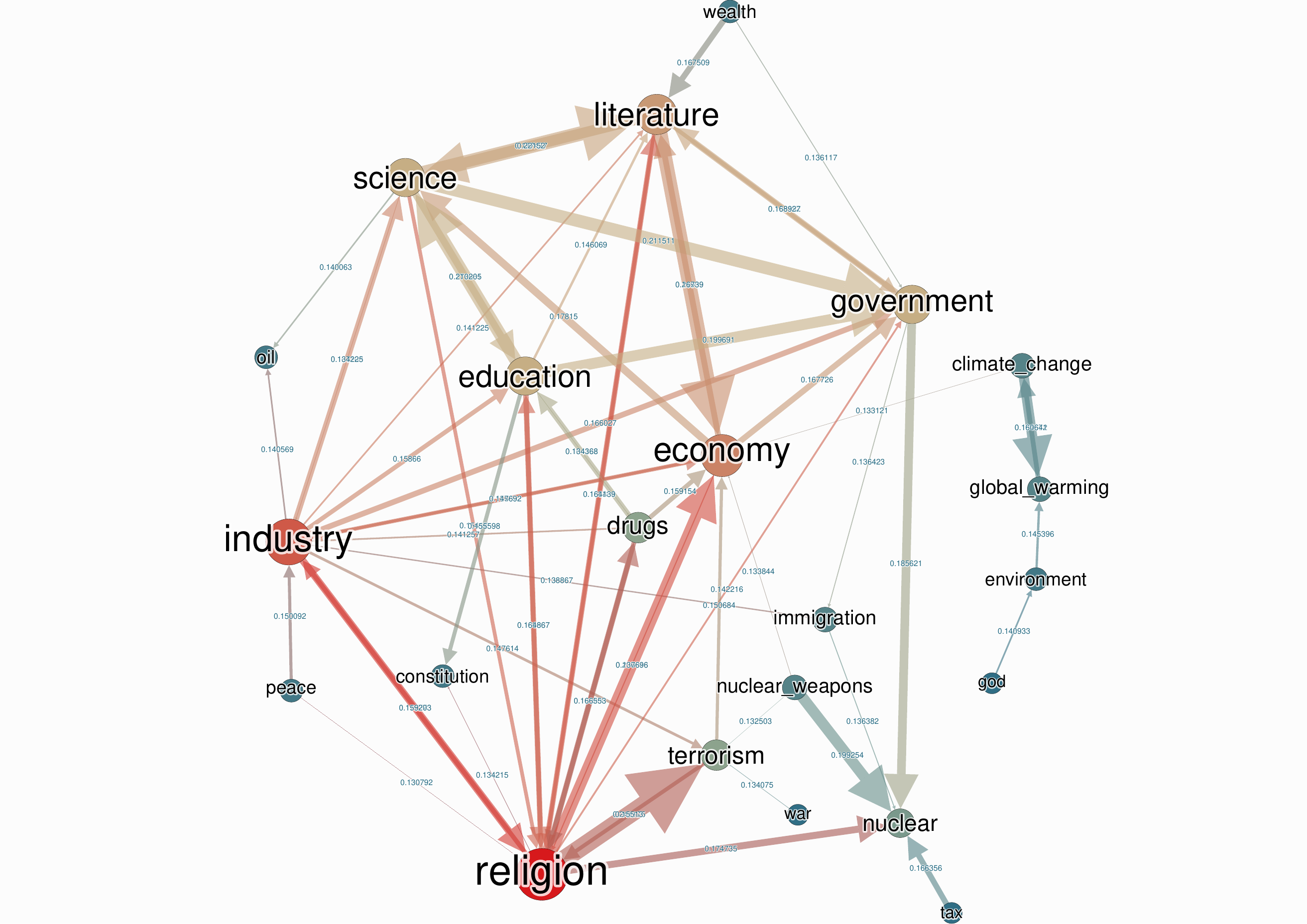}

\captionN{Full causality network computed with weekly search-frequency data corresponding to the keywords tabulated in Table~\ref{tab1} from Google Trends API. The size of the nodes, as well as the degree of ``redness'' is indicative of the weighted degree. The thickness of the arcs are indicative of the coefficient of causal dependence ($\gamma$). ``religion'' seems to have a particularly high degree.  }\label{figtrend}
\end{figure*}

\section{Application To Internet Search Trends}

Google Trends (http://www.google.com/trends/) provides a convenient API to download time series of weekly search frequencies for any given keyword. The time series' are normalized between $[0, 100]$, and corresponding to  a particular keyword, each integer-valued entry of the data series indicates the weekly sum-total of normalized worldwide  queries  submitted to the google website. Data is typically available from the first week of January, 2004. Thus, at the time of writing this paper, each of these search-trend data series are  $\sim 548$ entries long. We selected a small sample of keywords, typically ones that are strongly ``charged'' in a political or social context. The set of keywords are shown in Table~\ref{tab1}.

\begin{table}[t]
\centering
\captionN{Selected Search Keywords In Google Trends}\label{tab1}
\bf \sffamily\scriptsize\begin{tabular}{m{.6in}m{.6in}m{.6in}m{.6in}m{-4pt}}
\raggedright
(1) climate change & 
(2) constitution & 
(3) drugs & 
(4) economy &  \\[2.5ex] \hline 
\raggedright
(5) education &
(6) environment & 
(7) freedom & 
\raggedright
(8) global warming &\\[2ex]\hline  
\raggedright
(9) god & 
(10) government &
(11) guns & 
(12) heathcare &\\[2ex]\hline 
\raggedright
(13) immigration & 
(14) industry & 
(15) literature &
(16) music &\\[2ex]\hline 
\raggedright
(17) nuclear & 
\raggedright
(18) nuclear weapons & 
(19) oil & 
(20) peace &\\[2ex]\hline 
\raggedright
(21) religion & 
(22) science & 
(23) tax & 
(24) terrorism &\\[2ex]\hline 
\raggedright
(25) war &
(26) wealth & & &\\[2ex] 
\end{tabular}

\end{table}

This set of keywords can be easily expanded; however the objective of this example is to illustrate the theory developed in the preceding sections (and not, for example, derive new sociological insights; a potential future topic). The key point to note is that simple correlations do not elucidate any particularly interesting  dependency structure between search frequency data corresponding to different keywords. Indeed most of the data series' are highly correlated, $e.g.$, plate A  in  Figure~\ref{figcaus2} illustrate the data series' for the keywords ``education'', ``environment''; and plate B illustrate the series' for keywords ``economy'', ``government''. As shown, the data sets in plate A have a correlation coefficient of $0.96$, while those in plate B have a correlation coefficient of $0.75$. Thus, both pairs of data series' have  strong positive correlation. 

 It turns out (See plate C in Figure~\ref{figcaus2}) that inspite of having a high positive correlation,  the search-frequency data corresponding to ``education'' has little or no causal dependence on the data for ``environment''.  While, despite having a  lower correlation, search-frequency for ``economy'' and ``government'' seem to be strongly causally related, in one direction. Based on the discussion in Section~\ref{sec1}, this is entirely possible; while the plots in plate A of Figure~\ref{figcaus2} are seemingly very close (which leads to the strong statistical correlation), this has nothing to do with causality - what matters is if one data series carries unique information that can improve prediction of the other.

Note here an interesting consequence from  Granger's notion of causality: Two identical data series necessarily have no causal relationship; if series' $X$ and $Y$ are identical, $i.e.$, $X_t = Y_t,  \forall t$, then neither can improve the prediction of the other. 

The full causality network for the keywords in Table~\ref{tab1} is shown in Figure~\ref{figtrend}. We used a binary quantization to map each integer valued data series to a symbol stream; with symbol ``0'' indicating a drop in the search frequency from the previous week, and a ``1'' indicating identical or increased frequency. The size of the nodes, as well as the degree to which they are colored red, indicate the weighted degree of the graph. It appears that ``religion'' has a particularly high degree.

\section{Conclusion}\label{sec9}
We presented a new non-parametric approach to test for the existence of the degree of causal dependence between ergodic stationary weakly dependent symbol sources.
In addition, the notion of generative cross-models is made precise, and it is shown that crossed probabilistic automata are sufficient to represent a fairly broad class of direction-specific causal dependencies; and that they may be inferred efficiently from data.
Efficient inference of the coefficient of causal dependence, defined here, gives us the ability to investigate the network of causality flows between data sources. It is hoped that the the theoretical development presented here will open the door to understanding hidden mechanisms in diverse data-intensive fields of scientific inquiry.
}
\bibliographystyle{IEEEtran}
\bibliography{BibLib1,eqref,Bibcaus} 
\end{document}